\documentclass{article}

\usepackage{microtype}
\usepackage{graphicx}
\usepackage{subcaption}
\usepackage{booktabs} %

\usepackage{hyperref}

\usepackage[accepted]{icml2026}

\usepackage{amsmath}
\usepackage{amssymb}
\usepackage{mathtools}
\usepackage{amsthm}

\usepackage[capitalize,noabbrev]{cleveref}

\theoremstyle{plain}

\usepackage{times}
\usepackage{url}

\usepackage[utf8]{inputenc} %
\usepackage[T1]{fontenc}    %
\usepackage{amsmath}        %
\usepackage{amsfonts}       %
\usepackage{nicefrac}       %

\usepackage{bm}
\usepackage{multirow}
\usepackage{siunitx}
\usepackage{color}
\usepackage{dsfont}
\usepackage{setspace}
\usepackage{xspace}
\usepackage[normalem]{ulem} %
\usepackage{IEEEtrantools} %
\usepackage{wrapfig}
\usepackage{bigints}

\usepackage{pgfplots} 
\pgfplotsset{compat=newest} 
\pgfplotsset{plot coordinates/math parser=false} 
\usepackage{enumitem}

\newtheorem{assumption}{Assumption}
\newtheorem{lemma}{Lemma}
\newtheorem{theorem}{Theorem}
\newtheorem{corollary}{Corollary}
\newtheorem{proposition}{Proposition}

\usepackage{etoolbox}
\AtEndEnvironment{remark}{\hfill\IEEEQEDclosed}

\newcommand{\ac}{a} %
\newcommand{\st}{s} %

\newcommand{\St}{\mathbb{S}} %
\newcommand{\Ac}{\mathbb{A}} %
\newcommand{\T}{\mathcal{T}} %
\newcommand{\Rew}{\mathcal{R}} %

\newcommand{\ba}{\bm{\ac}}
\newcommand{\bs}{\bm{\st}}

\newcommand{\initstdist}{d_0}

\newcommand{\E}{\mathbb{E}}
\newcommand{\Var}{\var}

\newcommand{\defeq}{\triangleq}

\newcommand{\mse}{{\rm{MSE}}}
\newcommand{\var}{{\rm{Var}}}
\newcommand{\epe}{{\rm{EPE}}}
\newcommand{\supp}{{\rm{supp}}}
\newcommand{\kl}{{\rm{KL}}}

\newcommand{\loss}{\mathcal{L}}

\newcommand{\data}{\mathcal{D}}
\newcommand{\datadist}{\mathcal{D}}

\newcommand{\approxdatadist}{\datadist^{\oplanner}_{m}}

\newcommand{\planner}{p} %
\newcommand{\oplanner}{\planner^\star}
\newcommand{\approxplanner}{\widehat{p}}

\newcommand{\omu}{\mu^\star}
\newcommand{\oxi}{\xi^\star}
\newcommand{\mun}{\widehat{\mu}_{n}}
\newcommand{\xim}{\widehat{\xi}_{m}}

\newcommand{\bcpolclass}{\Theta_\mu}
\newcommand{\idmpolclass}{\Theta_\xi}

\newcommand{\pol}{\pi} %
\newcommand{\opol}{\pol^\star}
\newcommand{\opolidm}{\opol_{\rm{idm}}}
\newcommand{\bcpol}{\pi_\mu}
\newcommand{\idmpol}{\pi_\xi}

\newcommand{\polomu}{\pi_{\omu}}
\newcommand{\poloxi}{\pi_{\oxi}}
\newcommand{\polapproxoxi}{\pi_{\oxi}}

\newcommand{\pred}{f}
\newcommand{\opredbc}{\pred_{\rm{bc}}^\star}
\newcommand{\opredidm}{\pred_{\rm{idm}}^\star}
\newcommand{\predbc}{\pred_\mu}
\newcommand{\predidm}{\pred_\xi}
\newcommand{\predomu}{\pred_{\omu}}
\newcommand{\predoxi}{\pred_{\oxi}}
\newcommand{\predmun}{\pred_{\mun}}
\newcommand{\predxim}{\pred_{\xim}}

\newcommand{\biasmu}{b_{\mun}(\mu)}
\newcommand{\biasxi}{b_{\xim}(\xi)}

\newcommand{\dbiasmu}{b'_{\mun}(\omu)}

\newcommand{\dbbiasmu}{b'_{\mun}(\omu)}
\newcommand{\dbiasxi}{b'_{\xim}(\oxi)}
\newcommand{\dbbiasapproxoxi}{b'_{\xim}(\oxi)}
\newcommand{\biasmusq}{b^2_{\mun}(\omu)}
\newcommand{\biasxisq}{b^2_{\xim}(\oxi)}
\newcommand{\biasfmusq}{b^2_{f_{\mun}}}
\newcommand{\biasfxisq}{b^2_{f_{\xim}}}

\newcommand{\fishermu}{F_{\omu}}
\newcommand{\fisherxi}{F_{\oxi}}

\newcommand{\dmu}{\frac{\partial}{\partial \mu}}
\newcommand{\dxi}{\frac{\partial}{\partial \xi}}

\newcommand{\astar}{A\textsuperscript{$\star$}\xspace}

\newcommand{\revised}[1]{{\color{black} #1}}

\newcommand{\icmlWorkDoneWhileAtMSFT}{\textsuperscript{\textdagger}Work done while at Microsoft\xspace}

\icmltitlerunning{When Does Predictive Inverse Dynamics Outperform Behavior Cloning?}

\begin{document}

\twocolumn[
  \icmltitle{When Does Predictive Inverse Dynamics Outperform Behavior Cloning?}

  \icmlsetsymbol{equal}{*}
  \icmlsetsymbol{work}{\textdagger}

  \begin{icmlauthorlist}
    \icmlauthor{Lukas Sch\"afer}{equal,msft}
    \icmlauthor{Pallavi Choudhury}{equal,msft}
    \icmlauthor{Abdelhak Lemkhenter}{equal,msft}
    \icmlauthor{Chris Lovett}{equal,msft}
    \icmlauthor{Somjit Nath}{work,mcgill}
    \icmlauthor{Luis Fran\c{c}a}{msft}
    \icmlauthor{Matheus Ribeiro Furtado de Mendon\c{c}a}{msft}
    \icmlauthor{Alex Lamb}{work,tsing}
    \icmlauthor{Riashat Islam}{msft}
    \icmlauthor{Siddhartha Sen}{msft}
    \icmlauthor{John Langford}{msft}
    \icmlauthor{Katja Hofmann}{msft}
    \icmlauthor{Sergio Valcarcel Macua}{equal,msft}
  \end{icmlauthorlist}

  \icmlaffiliation{msft}{Microsoft}
  \icmlaffiliation{mcgill}{McGill University, Mila}
  \icmlaffiliation{tsing}{Tsinghua University}

  \icmlcorrespondingauthor{Lukas Sch\"afer}{lukas.schaefer@microsoft.com}
  \icmlcorrespondingauthor{Sergio Valcarcel Macua}{sergiov@microsoft.com}

  \icmlkeywords{Imitation learning, behavior cloning, inverse dynamics models, sample efficiency}

  \vskip 0.3in
]

\printAffiliationsAndNotice{
    \icmlEqualContribution
    \icmlWorkDoneWhileAtMSFT
}

\begin{abstract}
Behavior cloning (BC) is a practical offline imitation learning method, but it often fails when expert demonstrations are limited. Recent works have introduced a class of architectures named predictive inverse dynamics models (PIDMs) that combine a future-state predictor with an inverse dynamics model.
While PIDMs often outperform BC, the reasons behind their benefits remain unclear. In this paper, we provide a theoretical explanation: PIDMs introduce a tradeoff. 
Conditioning the IDM on the predicted future state can significantly reduce variance, but the prediction itself introduces additional bias and variance.
We establish conditions for PIDMs to achieve higher sample efficiency and lower prediction error than BC, with the gap widening when additional data sources are available. 
We validate the theoretical insights empirically in 2D navigation tasks, where BC requires up to five times (three times on average) more demonstrations than PIDM to reach comparable performance.
Results are also illustrated in a complex 3D environment in a modern video game with high-dimensional visual inputs and stochastic transitions, where BC requires over 66\% more samples than PIDM. 
\end{abstract}

\section{Introduction}

\begin{figure*}[t]
    \centering
    \begin{subfigure}{.70\textwidth}
        \centering
        \begin{minipage}{0.48\textwidth}
            \centering
            \includegraphics[width=\linewidth]{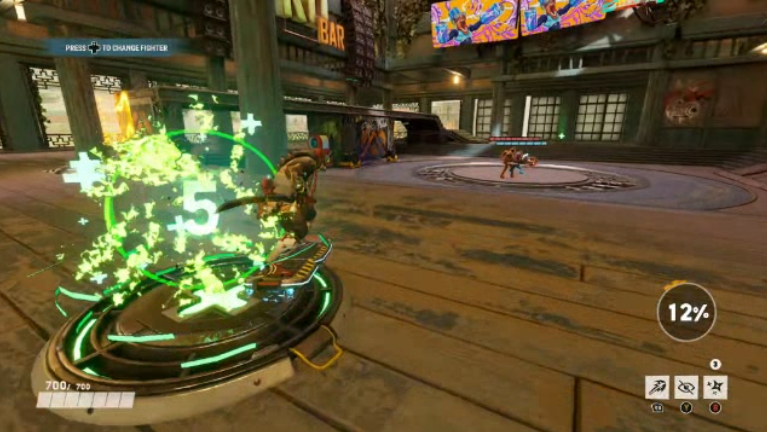} \\
            \small\#2 Grab health marker
        \end{minipage}
        \hspace{.5em}
        \begin{minipage}{0.48\textwidth}
            \centering
            \includegraphics[width=\linewidth]{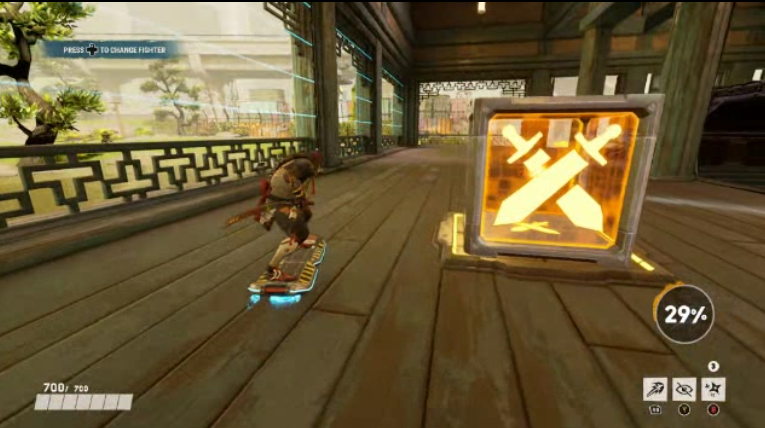} \\
            \small \#7 Avoid obstacle
        \end{minipage}
        \caption{Selected milestones of "Tour" task in 3D world}
    \end{subfigure}
    \hfill
    \begin{subfigure}{.29\linewidth}
        \includegraphics[width=\linewidth]{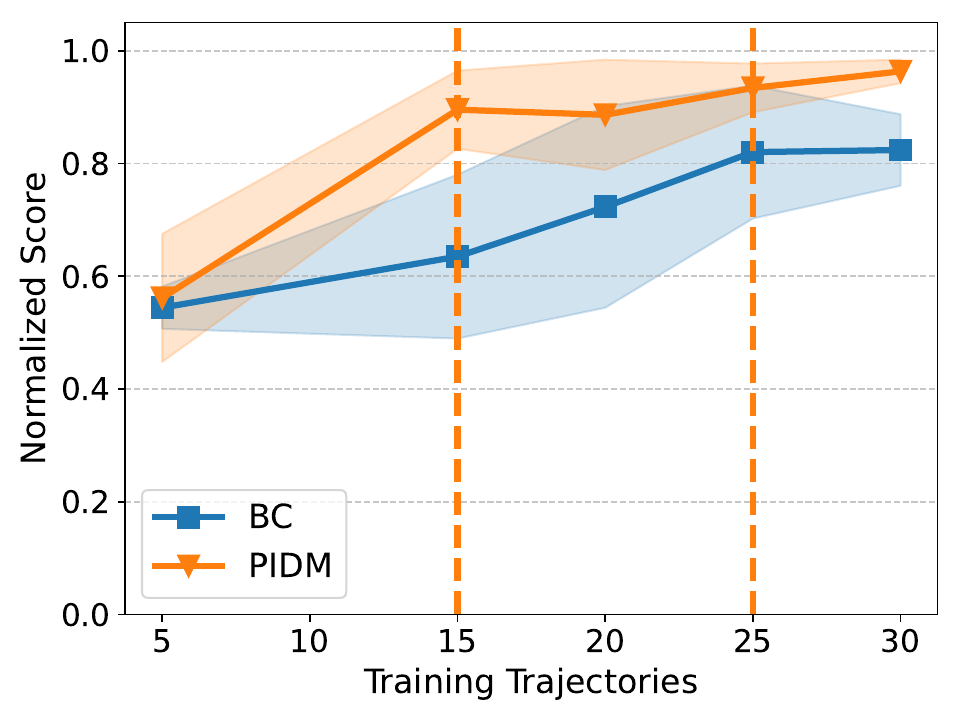}
        \caption{Sample efficiency}
    \end{subfigure}
    \caption{(a) Visualization of selected milestones from the "Tour" task in a 3D video game with stochastic transitions and real-time inference. (b) Sample efficiency curves (mean $\pm$ std) for PIDM and BC, with BC needing 66\% more samples to achieve 80\% success rate.}
    \label{fig:xbox_highlight_result}
\end{figure*}

Offline imitation learning learns closed-loop control policies that replicate expert behavior only from pre-collected data, without access to a reward function or further environment interactions. This paradigm has broad applicability across domains such as robotics~\citep{schaal1999imitation,fang2019survey}, autonomous driving~\citep{pan2020imitation}, and gaming~\citep{pearce2022counter,pearce2023imitating,schafervisual}. 
The most common offline imitation learning approach is behavior cloning (BC), which can learn complex behavior \citep{osa2018algorithmic, pearce2022counter, florence2022implicit} but typically requires many demonstrations per task. 
This reliance on large datasets can be problematic, as collecting expert demonstrations at scale is often costly, time-consuming, or even infeasible.

\begin{figure*}[t]
  \centering
  \begin{subfigure}{0.14\textwidth}
    \centering
    \includegraphics[width=\linewidth,clip,trim=0.9cm 0.2cm 1.4cm 0cm]{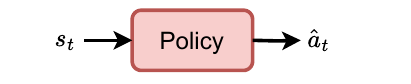}
    \caption{BC}
    \label{fig:bc_architecture}
  \end{subfigure}
  \hfill
  \begin{subfigure}{0.28\textwidth}
    \centering
    \includegraphics[width=\linewidth,clip,trim=1cm 0.15cm 1.5cm 0cm]{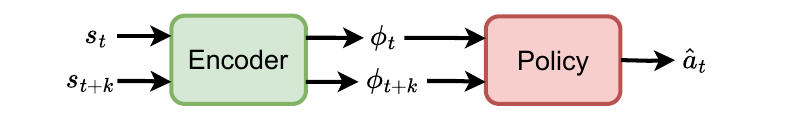}
    \caption{Multi-step IDM}
    \label{fig:midm_architecture}
  \end{subfigure}
  \hfill
  \begin{subfigure}{0.28\textwidth}
    \centering
    \includegraphics[width=\linewidth,clip,trim=0.9cm 0.1cm 1.3cm 0cm]{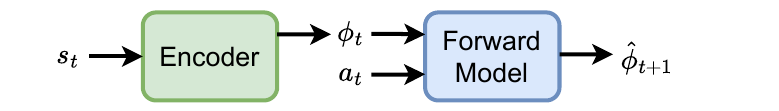}
    \caption{Forward model}
    \label{fig:forward_model_architecture}
  \end{subfigure}
  \hfill
  \begin{subfigure}{0.28\textwidth}
    \centering
    \includegraphics[width=\linewidth,clip,trim=0.9cm 0.2cm 1.5cm 0cm]{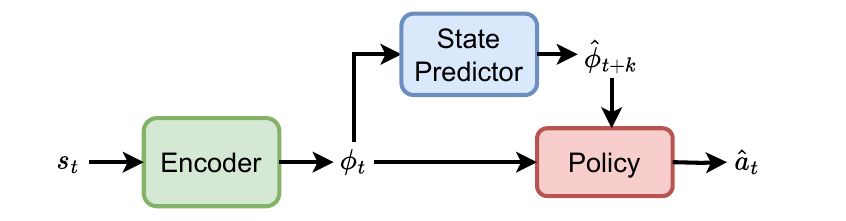}
    \caption{PIDM}
    \label{fig:pidm_architecture}
  \end{subfigure}
  \caption{\textbf{(a)} BC learns a policy conditioned on the current state. 
  \textbf{(b)} IDM learns a policy conditioned on the current state and a future state $k$ steps ahead. 
  \textbf{(c)} Forward models predict a future state (representation) given a state and action. 
  Note (b) IDM and (c) forward models can serve as auxiliary objectives to learn an encoder that provides effective state representations. 
  \textbf{(d)} PIDM represents an alternative to BC that consists of a state predictor (akin to an action-free forward model) that predicts future-state representations, and an IDM policy. 
  }
  \label{fig:architectures}
\end{figure*}

Recent work has introduced predictive inverse dynamics models (PIDMs) as a promising alternative to behavior cloning (BC)~\citep{du2023learning,
tian2025predictive,
tot2025adapting,
xie2025latent,pai2026mimic}. 
A PIDM combines two components: a state predictor that predicts plausible future states, and an inverse dynamics model (IDM) that infers the actions needed to reach them (see \Cref{fig:pidm_architecture}). 
By augmenting limited expert demonstrations with additional data sources, such as action-free demonstrations or non-expert data, PIDMs have shown strong empirical performance~\citep{xie2025latent}. 
Interestingly, prior work also shows that even without additional data, PIDMs can outperform BC.

However, the underlying reasons for the efficiency of PIDMs remain unclear. 
Is there something intrinsic to the modular architecture that enables this advantage? Under what conditions can we expect such gains to consistently emerge?

In this work, we analyze PIDMs and show theoretically why decomposing the decision-making problem into a state predictor and an IDM can improve performance over BC.
First, we show that the prediction error of the optimal PIDM estimator is always less than or equal to that of BC, resulting in a  principled performance gap in favor of PIDM. This gap is characterized by the expected conditional variance of actions given future states.
We then extend the analysis to non-optimal estimators and reveal a key tradeoff in PIDM. Conditioning on a future state reduces total variance by eliminating the conditional variance identified above; however, predicting the future state introduces a covariate shift that can reduce the effective performance gap.
We conclude by deriving conditions under which PIDM is at least as sample-efficient as BC, achieving comparable or lower training-time mean squared error with fewer expert demonstrations.

Second, we provide empirical evidence that the predicted performance gains apply to more general conditions than those studied in the theory, including the small-data regime. We perform experiments on a toy benchmark of four 2D navigation tasks,\footnote{Code available at \url{https://github.com/microsoft/understanding_pidm_for_imitation_learning}.} 
using a dataset of human demonstrations, and observe that BC requires up to $5\times$ more demonstrations than PIDM.  
To further understand how the theory manifests in practice, we compute the conditional action variance and bias of the state predictor, and empirically validate the predicted correlations from our theoretical analysis.
These experiments in a state-based environment additionally isolate the efficiency gains due to the predicted error gap from the representational benefits of IDM shown in previous work~\citep{lamb2023guaranteed,koul2023pclast,levine2024multistep,islam2022agent}.

After building intuition as to \textit{why} the PIDM decomposition is effective, we extend our investigation to complex tasks that require imitating complex human demonstrations, from image inputs, in a 3D world with stochastic transitions, in real-time. In this real-world setting, sample efficiency is critical since obtaining human demonstrations is costly, and real-time requirements introduce additional constraints on the solution.
Even with a simple state predictor, we continue to observe substantial efficiency gains: BC requires 66\% more samples than PIDM, demonstrating that the predicted performance gap is relevant for real-world applications.

\section{Related Work}
\label{sec:related_work}

\textbf{Inverse dynamics models. }
Inverse dynamics models (IDMs) predict the action that initiates a sequence leading from the current state to a future state $k$ steps ahead. 
The multi-step inverse dynamic loss has been used to train encoders on high-dimensional observations, since it can filter out exogenous factors and learn rich state representations that generalise across tasks~\citep{mhammedi2023musik, efroni2022ppe, lamb2023guaranteed,koul2023pclast}.
In this work, we focus on architectures in which states are encoded into a latent space (\Cref{fig:midm_architecture}), and both the state predictor and the IDM policy operate on these latent representations.

\textbf{PIDMs. }
Prior works instantiate PIDM with varying choices of state space, training data, and optimization strategy.
\citet{du2023learning} trained a diffusion model to predict future images directly in pixel space.
\citet{xie2025latent} instead operated in a compact latent space, exploiting action-free demonstrations for the state predictor and diverse action-labeled data for the IDM.
\citet{tian2025predictive} trained both components jointly end-to-end. \citet{tot2025adapting} conditioned on representations from pretrained vision and world models.
\citet{pai2026mimic} used a fine-tuned video model as the state predictor, generating frames that visualize task completion from language conditioning.
Collectively, these works provide empirical evidence that PIDM can outperform BC. However, they do not explain when or why such gains arise. Our work addresses this question through a theoretical and empirical analysis of the PIDM decomposition.

\textbf{Forward models. }
Forward models (\Cref{fig:forward_model_architecture}), also referred to as world models, can serve as auxiliary objectives to improve learned representations~\citep{levine2024multistep} or facilitate planning via reinforcement learning~\citep{thrun1990planning,hafner2025mastering} and model-predictive control~\citep{zhou2024dino,bar2025navigation}.
The PIDM state predictors differ from forward models in two key ways. First, it is conditioned only on the current state, not on actions. Second, it predicts future states $k \geq 1$ steps ahead rather than the immediate next state.

\textbf{BC and trajectory modeling. }
Recent analysis \citep{foster2024behavior} argues that many practical implementations of BC, which rely on a log-loss, are implicitly modeling the whole state-action sequence. 
Our work complements such analysis by providing evidence that explicitly modeling part of the trajectory, the future state in the PIDM case, can improve sample efficiency and generalization.

\section{Preliminaries}

\textbf{Problem setting. } We consider an MDP $(\St, \Ac, \T, \Rew, \initstdist)$ with 
state space $\St$, 
action space $\Ac$, 
transition model $\T$,
reward function $\Rew$,
and initial state distribution $\initstdist$. Let $\datadist$ denote the data distribution resulting from following an unknown expert policy $\opol$. 
Let $\pol_\mu$ denote a policy parameterized by $\mu$. 
We consider the offline imitation learning setting in which the goal is to learn $\mu$ such that $\pol_\mu \approx \opol$ given a dataset of state--action trajectories drawn from $\datadist$, and without access to the reward. 
We study two architectures: BC and PIDM.

\textbf{BC} treats this setting as supervised learning problem, training a policy $\bcpol$ with parameter $\mu$ to imitate actions in the dataset given the most recent state (\Cref{fig:bc_architecture}). Given a dissimilarity measure $\ell$, it minimizes the following loss:
\begin{equation}
  \label{eq:bc_loss}
  \loss_\text{bc}(\bcpol) 
= 
    \E_{
        (\bs_t, \ba_t) \sim \datadist, \widehat{\ba}_t \sim \bcpol(\cdot \mid \bs_t)
    } 
    \left[
        \ell(\widehat{\ba}_t, \ba_t)
    \right].
\end{equation}

\textbf{PIDM} consists of two main models (\Cref{fig:pidm_architecture}): a state predictor $\planner$, which predicts future states for some horizon $k$, and an IDM $\pol_\xi$ with parameter $\xi$, which predicts the next action needed to get from the current state to a future state. They can be trained with the following losses:
\begin{align}
    \loss_\text{sp}(\approxplanner) 
&= 
    \E_{
        \substack{
            (\bs_t, \bs_{t+k}) \sim \datadist
            ,\\
            \widehat{\bs}_{t+k} \sim \approxplanner(\cdot \mid \bs_t)
        }
    } 
    \left[
        \ell(\widehat{\bs}_{t+k}, \bs_{t+k})
    \right] 
,
\label{eq:state_predictor_loss}
\\
  \loss_\text{idm}(\idmpol) 
&= 
    \E_{
        \substack{
                (\bs_t, \ba_t) 
            \sim 
                \datadist
            , 
                \bs_{t+k}
            \sim
                \planner(\cdot \mid \bs_t)
            ,\\
                \widehat{\ba}_{t} 
            \sim 
                \idmpol(\cdot \mid \bs_t, \bs_{t+k})
        }
    } 
    \left[
        \ell(\widehat{\ba}_{t}, \ba_t)
    \right]
.
\label{eq:pidm_loss}
\end{align}

PIDM offers several design choices: the state predictor and IDM models can be trained jointly or independently, operate directly on observations or on a shared latent space (see \Cref{fig:pidm_architecture}), and use the same or different data sources. 
In addition, the IDM can be trained using future states drawn either from $\data$ or from the learned state predictor.

 \textbf{Notation. } We distinguish between a \emph{policy} $\pol_\mu$ and its \emph{action predictor} $f_\mu$. The BC and IDM policies define conditional \emph{distributions} over actions denoted by $\bcpol: \St \to P(\Ac)$ and $\idmpol: \St \times \St \to P(\Ac)$ with parameters $\mu\in\bcpolclass$ and $\xi\in\idmpolclass$, respectively, where $P(\Ac)$ denotes a distribution over actions. In contrast, the corresponding predictors define a deterministic mapping induced by that distribution (e.g., the conditional mean), denoted by $\predbc: \St \to \Ac$ and $\predidm: \St \times \St \to \Ac$, respectively.
\section{Theoretical Analysis}
\label{sec:theoretical-analysis}

Under the PIDM architecture, the expert state-conditioned policy can be expressed in terms of an expert IDM policy, $\opolidm$, together with an explicit model of the future state:
\begin{equation}
\label{eq:decomposition}
    \opol(a_t \mid s_t)
=
    \int_{\St}
        \oplanner(s_{t+k} \mid s_t)
        \opolidm(a_t \mid s_t, s_{t+k})
        \, ds_{t+k},
\end{equation}
where $\oplanner$ denotes the true future-state distribution.

Intuitively, this decomposition can simplify the learning of a policy whenever conditioning the IDM on a future state provides useful information for identifying the correct action. 
In this section, we compare BC and PIDM in terms of generalization, measured through test-time prediction error (\Cref{ssec:epe-gap}),
and sample efficiency, measured through the training-time mean squared error (\Cref{ssec:sample-efficiency}).
All proofs are deferred to \Cref{app:proofs}.

For clarity, we present the results for the scalar case: 
$
    \mu \in \bcpolclass \subseteq \mathbb{R}
$, 
$
    \xi \in \idmpolclass \subseteq \mathbb{R}
$, but the extension to vector parameters is straightforward. 
We also focus on single-point predictors under the mean-squared loss (see discussion on extension to other losses in \Cref{app:cross-entropy-loss}).

\subsection{Generalization Gap}
\label{ssec:epe-gap}
Let 
$
    d
$
be the marginal distribution over the current state.
The prediction error for any BC and IDM population estimators is given by:
\begin{align}
    \epe(\pred) 
&
\defeq
    \E_{\bs_t \sim d, \ba_t \sim \opol(\cdot|\bs_t)}
    \left[ 
        \left( 
            \ba_t - \pred(\bs_t)
        \right)^2 
    \right]
,
\\
    \epe(\pred; p) 
&
\defeq
    \E_{
        \substack{
            \bs_t \sim d, \bs_{t+k} \sim p(\cdot|\bs_t),
        \\
            \ba_t \sim \opolidm(\cdot|\bs_t, \bs_{t+k})
        }
    }
    \left[ 
        \left( 
        \ba_t
        -
        \pred(\bs_t, \bs_{t+k})
        \right)^2 
    \right]
,
\end{align}
where $\epe(\cdot; p)$ highlights that the expected predicted error for IDM is computed using future states sampled from $p$.

Define the optimal BC and IDM predictors that minimize their respective expected prediction error:
\begin{align}
    \opredbc(\bs_t) 
&
\defeq 
    \E_{\ba_t \sim \opol( \cdot \mid \bs_t)}[\ba_t ]
,
\\
    \opredidm(\bs_t, \bs_{t+k}) 
&
\defeq 
    \E_{\ba_t \sim \opolidm( \cdot \mid \bs_t, \bs_{t+k})}[ \ba_t ]
.
\end{align}

Define the predicted error gap for the optimal predictors under the true future-state distribution $\oplanner$:
\begin{equation}
    \Delta_{\oplanner}
\defeq
    \epe(\opredbc)
    -
    \epe(\opredidm; \oplanner).
\label{eq:epe-gap}
\end{equation}
Our first result quantifies the prediction error gap between the optimal estimators in terms of the uncertainty in $\ba_t$ arising from uncertainty in $\bs_{t+k}$.
\begin{theorem}
\label{theorem:mse-gap}
For optimal predictors $\opredbc$ and $\opredidm$: 
\begin{equation}
    \Delta_{\oplanner}
=
    \E_{\bs_t \sim d}
    \left[
        \var_{\bs_{t+k} \sim \oplanner(\cdot| \bs_t)}
        \left(
            \opredidm(\bs_t, \bs_{t+k})
        \right)
    \right]
\ge
    0
.
\end{equation}
\end{theorem}

\Cref{theorem:mse-gap} shows that conditioning the optimal predictor additionally on $\bs_{t+k}$ sampled from the true future-state distribution $\oplanner$ can only decrease the expected prediction error of $\ba_t$.

Next, we study how the prediction error gap changes for non-optimal estimators. 
We consider the setting where the IDM policy is learned using future states sampled from the true future-state distribution,
$s_{t+k} \sim \oplanner(\cdot \mid s_t)$, 
but at test-time this distribution is not available. Therefore, we instead condition the IDM policy on future states generated by an approximate state predictor:
$s_{t+k} \sim \approxplanner(\cdot \mid s_t)$.
As a result, the policy is evaluated on a different input distribution than the one used during training, inducing a covariate shift in the IDM inputs~\citep{sugiyama2012machine}.

When only a finite dataset is available, this covariate shift
($\oplanner \neq \approxplanner$)
introduces additional estimation error.
Consequently, the predictor induced by PIDM has two main sources of error: 
(i) its intrinsic bias and variance (e.g., due to finite-sample and/or model misspecification), and 
(ii) the additional  bias and variance caused by the covariate shift induced by the approximate state predictor.

These sources of estimation error can reduce the predicted error gap, as formalized in the following corollary.
We require that the support of $\approxplanner$ must be contained in 
that of $\oplanner$.
\begin{assumption}%
\label{ass:support}
$
    \supp(\approxplanner(\cdot | \bs_t))
    \subseteq
    \supp(\oplanner(\cdot | \bs_t))
$,
$
    \forall \bs_t \in \St
$.
\end{assumption}

Provided \Cref{ass:support} holds, \Cref{proposition:same-optimal-predictor} in \Cref{app:proposition:same-optimal-predictor} shows that the optimal IDM predictor $\opredidm$ is independent of the  test-time future-state distribution. The assumption also ensures that the following density ratio that quantifies the covariate shift is well-defined:
\begin{equation}
    w(s_t,s_{t+k})
\defeq
    \frac{
        \approxplanner(s_{t+k}\mid s_t)
    }{
        \oplanner(s_{t+k}\mid s_t)
    }
.
\end{equation}
For simplicity, we also assume i.i.d. sampling  
(see \Cref{app:non-iid} for a discussion of the non-i.i.d.\ case).
\begin{assumption}[i.i.d.] 
Samples are drawn i.i.d.\ from the data distributions.
\label{ass:iid}
\end{assumption}
Let $\datadist_n$ and $\approxdatadist$ denote the training datasets 
for BC and IDM with $n$ and $m$ samples, respectively:
\begin{align*}
    \datadist_n
&\defeq
   \left \lbrace
            \left(
                s_t, a_t
            \right)
    \stackrel{\text{i.i.d.}}{\sim}
        \datadist 
    \right \rbrace^n
,
\\
    \approxdatadist
& \defeq
   \left \lbrace
            \left(
                s_t, a_t, s_{t+k}
            \right)
    \mid
        (s_t, a_t) 
    \stackrel{\text{i.i.d.}}{\sim}
        \datadist 
    ,
        s_{t+k} 
    \sim 
        \oplanner(\cdot \mid s_{t})
    \right \rbrace^m
\end{align*}
When we consider the datasets as random variables themselves (so the estimators derived from them are also random variables),
we write $\E_{\datadist_n}$, $\var_{\datadist_n}$,  $\E_{\approxdatadist}$, and $\var_{\approxdatadist}$
for the expectations and variance over dataset realizations.
\begin{corollary}
\label{corollary:gap-general-estimator}
Under Assumptions \ref{ass:support}--\ref{ass:iid}, let $\predmun$ and $\predxim$ denote the BC and IDM predictors learned from $\datadist_n$ and $\approxdatadist$, respectively.
Define the variance, bias, and irreducible-noise gaps under covariate shift:
\begin{align}
    \delta
&
\defeq
    \E_{\bs_t \sim d}
    \Big[
        \var_{\datadist_n}
        \left(
            \predmun
            \left(
                \bs_t
            \right)
        \right)
        -
        \E_{\bs_{t+k}\sim \oplanner(\cdot | \bs_t)}
        \Big[
\notag\\
&
\qquad\qquad
            w(\bs_t, \bs_{t+k})
            \var_{\approxdatadist}
            \left(
                \predxim(\bs_t, \bs_{t+k})
            \right)
        \Big]
    \Big]
,
\\
    \beta
&
\defeq
    \:
    \biasfmusq
    -
    \biasfxisq
,
\\
    \gamma
&\defeq
    \E_{
        \substack{
            \bs_t \sim d, \\
            \bs_{t+k} \sim \oplanner(\cdot | \bs_t)}
        }
    \Big[
        \left(
            1 - w(\bs_t,\bs_{t+k})
        \right)
        \var(\ba_t \mid \bs_t, \bs_{t+k})
    \Big]
\end{align}
with bias terms defined as:
\begin{align}
    \biasfmusq
&
\defeq
\:
    \E_{\bs_t \sim d}
    \left[
        \left(
            \E_{\datadist_n}\left[ \predmun(\bs_t) \right]
            -
            \opredbc(\bs_t)
        \right)^2
    \right],
\\
    \biasfxisq
&
\defeq
\:
        \E_{
            \substack{
                \bs_t \sim d,\\
                \bs_{t+k} \sim \oplanner(\cdot|\bs_t)
            }
        }
        \bigg[
            w(\bs_t, \bs_{t+k})
\notag\\
&
            \Big(
                \E_{\approxdatadist}\left[ \predxim(\bs_t, \bs_{t+k}) \right]
                -
                \opredidm(\bs_t, \bs_{t+k})
            \Big)^2
        \bigg]
.
\end{align}
Then, the predicted error gap is given by:
\begin{align}
    \widehat{\Delta}_{\approxplanner}
\defeq
    \epe
    \left(
        \predmun
    \right)
    -
    \epe
    \left(
        \predxim
        ;
        \approxplanner
    \right)
=
    \Delta_{\oplanner}
    +
    \delta
    +
    \beta
    +
    \gamma
.
\label{eq:gap-general-estimator}
\end{align}
\end{corollary}

A positive error gap $\widehat{\Delta}_{\approxplanner}$ implies the PIDM estimator offers lower prediction error than BC.
\Cref{corollary:gap-general-estimator} shows how PIDM introduces a  tradeoff:
$\Delta_{\oplanner}$ represents the (nonnegative) variance reduction of the optimal IDM predictor due to knowing the future state, and contributes to potential gains of PIDM over BC. 

On the other hand, $\beta$ and $\delta$ represent the differences in bias and variance of the BC and PIDM estimators. 
In most cases, we expect $\beta < 0$ and $\delta < 0$, due to additional bias and variance induced by the covariate shift arising from an approximate state predictor (assuming both BC and PIDM estimators have comparable intrinsic bias and variance as required for a fair comparison) and, thus, they might reduce the potential gains of PIDM.
The covariate shift might also reduce the gains of PIDM when $\gamma < 0$.

Interestingly, note that the covariate shift can also benefit PIDM ($\gamma > 0$) when $\approxplanner$ places more mass than $p^\star$ (such that $0<w<1$) on future states with lower irreducible variance. Similarly, $\approxplanner$ might also reduce the estimator bias ($\beta > 0$) and variance ($\delta >0$) by putting more mass on future states with lower estimator bias and/or variance.

\Cref{corollary:gap-general-estimator} further motivates the use of additional data sources. Action-free demonstrations of the same task can be used to train a more accurate state predictor, reducing the magnitude of the covariate shift $\gamma$ and, consequently, the additional bias and variance arising from $\approxplanner$, so that $\delta$ and $\beta$ approach zero. 
Also, when $k=1$, expert demonstrations from different tasks (or even non-expert demonstrations) collected in the same environment can reduce the variance of $\predxim$, thereby making $\delta > 0$. \citet{xie2025latent} empirically reported performance gains of PIDM from leveraging some of these additional data sources. \Cref{corollary:gap-general-estimator} provides a theoretical explanation for these gains by showing how extra data can directly reduce some of the terms governing the predicted-error gap.

\subsection{Sample Efficiency Gain}
\label{ssec:sample-efficiency}
In this subsection, we analyze sample efficiency through the training-time mean squared error (MSE) of the estimated parameters.
We assume correct specification to anchor the Fisher information to the data generating process.
We also assume standard regularity conditions under which the Cramér-Rao lower bound holds (see, e.g., \citet[Theorem 11.10.1]{cover2005elements}).
\begin{assumption}[Expressivity]
\label{ass:policy_class_expressivity}
The BC and IDM policy classes are correctly-specified:
\begin{align}
    \exists \omu \in \bcpolclass
    \; | \;
&
    \polomu = \opol
.
\quad
    \exists \oxi \in \idmpolclass
    \; | \;
    \poloxi
=
    \opolidm
.
\end{align}
\end{assumption}
Let the bias of each policy parameter and its derivative evaluated at the optimal parameter be given by:
\begin{align}
    \biasmu
&
\defeq
    \E
    \left[
        \mun
    \right]
    -
    \mu
,
\;
    \dbiasmu
\defeq
    \frac{
        \partial 
    }{
        \partial \mu
    }
    \biasmu
\biggr\rvert_{\mu=\omu}
,
\\
    \biasxi
&\defeq
    \E
    \left[
        \xim
    \right]
    -
    \xi
,
\;
    \dbbiasapproxoxi
\defeq
    \frac{
        \partial 
    }{
        \partial \xi
    }
    \biasxi
\biggr\rvert_{\xi=\oxi}
,
\end{align}
where the expectation is taken over datasets generated by policies $\pol_\mu$ and $\pol_\xi$, respectively. Define the Fisher information evaluated at the optimal parameter:
\begin{align}
&
    \fishermu
\defeq
    \E_{\substack{
        \bs_t \sim d,\\
        \ba_t \sim \pol_{\omu}(\cdot | \bs_t)
    }}
    \left[
        \left(
            \frac{\partial \: \ln \pol_\mu (\ba_t | \bs_t)}{\partial \mu}
            \Big\rvert_{\mu=\omu}
        \right)^2
    \right]
,
\\
&\fisherxi
\defeq
\notag
\\
&
    \E_{\substack{
            \bs_t \sim d,
            \bs_{t+k} \sim \oplanner(\cdot | \bs_t),\\
            \ba_t \sim \polapproxoxi(\cdot | \bs_t, \bs_{t+k})
    }}
    \left[
        \left(
            \frac{\partial \: \ln \idmpol (\ba_t | \bs_t, \bs_{t+k})}{\partial \xi}
            \Big\rvert_{\xi=\oxi}
        \right)^2
    \right].
\end{align}

\begin{theorem}
\label{theorem:sample-efficiency-gain}
Let Assumptions~\ref{ass:iid}--\ref{ass:policy_class_expressivity} hold.
Let $\mun$ and $\xim$ denote the estimated parameters for the BC and IDM predictors learned from $\datadist_n$ and $\approxdatadist$, respectively, where $n$ and $m$ denote the number of training samples required to achieve a target MSE level $\varepsilon$.
Assume that both estimators asymptotically attain the biased Cramér-Rao lower bound.
Then, for sufficiently large $n$ and $m$, we have:
\begin{align}
    \eta
&\defeq
    \frac{n}{m}
\approx
    \frac{\fisherxi}{\fishermu}
        \frac{
            \left(
                \varepsilon 
                - 
                \biasxisq
            \right)
            \left(
                1
                +
                \dbbiasmu
            \right)^2
        }{
            \left(
                \varepsilon 
                - 
                \biasmusq
            \right)            
            \left(
                1
                +
                \dbiasxi
            \right)^2
        }
.
\label{eq:sample-efficiency-gain}
\end{align}
\end{theorem}
Although \eqref{eq:sample-efficiency-gain} may look complex at first, 
it can be simplified in a straightforward and intuitive way. 
We seek conditions that ensure $\eta \ge 1$, which implies that PIDM is at least as sample efficient as BC.
We assume that $\bcpol$ can be expressed as the marginal of the IDM policy $\idmpol$ for any $\xi$ in a neighborhood of $\oxi$.
We also assume regularity conditions ensuring applicability of dominated convergence to $\idmpol$.
Then, the following lemma shows the Fisher ratio is at least one.
\begin{assumption}[Local shared parameterization]
\label{ass:local-shared-param}
There exists $\rho > 0$ and a continuously differentiable map $\Psi: U \to \bcpolclass$ on $U \defeq (\oxi - \rho, \oxi + \rho) \subset\idmpolclass$, with $\Psi(\oxi) = \omu$ and $\Psi'(\oxi) = 1$, such that for all $\xi \in U$ and $\bs_t\in\St$:
\begin{equation}
    \pi_{\Psi(\xi)}(\cdot \mid \bs_t)
=
    \int_\St
        \oplanner(s_{t+k} \mid \bs_t)\,
        \pi_\xi(\cdot \mid \bs_t, s_{t+k})\,
        ds_{t+k}.
\end{equation}
\end{assumption}
\begin{assumption}[Regularity]
\label{ass:regularity-fisher}
The mapping $\xi \mapsto \idmpol(a_t \mid s_t, s_{t+k})$ is differentiable and dominated by an integrable function uniformly in a neighborhood of $\xi$, so that differentiation and integration can be interchanged:
\begin{equation}
    \dxi 
    \E_{\oplanner}
    \left[
        \idmpol(a_t \mid s_t, s_{t+k})
    \right]
=
    \E_{\oplanner}
    \left[
        \dxi 
        \idmpol(a_t \mid s_t, s_{t+k})
    \right]
.
\end{equation}
\end{assumption}
\begin{lemma}
\label{lemma:fisher-ratio-greater-than-one}
Under Assumptions \ref{ass:policy_class_expressivity}--\ref{ass:regularity-fisher}:
$
    \fisherxi
\ge 
    \fishermu
$.
\end{lemma}

At training time, the IDM parameter $\xim$ is learned using future states sampled from the true distribution $\oplanner$. 
Consequently, no covariate shift is introduced during training, and any bias arises solely from intrinsic estimation error (e.g., due to finite data). 
For a fair comparison, we assume that the intrinsic bias terms of BC and IDM are equal, so that they cancel in \eqref{eq:sample-efficiency-gain}.
Under this assumption, the following result shows that PIDM is at least as sample efficient as BC.
We use $\gtrsim$ to denote greater than or approximately equal to.
\begin{corollary}
\label{corollary:sample-efficient-similar-bias}
Under the conditions of \Cref{theorem:sample-efficiency-gain} and Assumptions 
\ref{ass:local-shared-param}--\ref{ass:regularity-fisher}, 
if
$\biasmusq = \biasxisq$ and $\dbiasmu=\dbiasxi$, then:
$
\eta \gtrsim 1
$.
\end{corollary}

Although these conditions for sample efficiency gains have been derived for large enough samples, 
\Cref{sec:experiments} provides empirical evidence that efficiency gains hold even in the small-data regime.

\begin{figure*}[t]
  \centering
  \begin{subfigure}{0.2\textwidth}
    \centering
    \includegraphics[width=\textwidth]{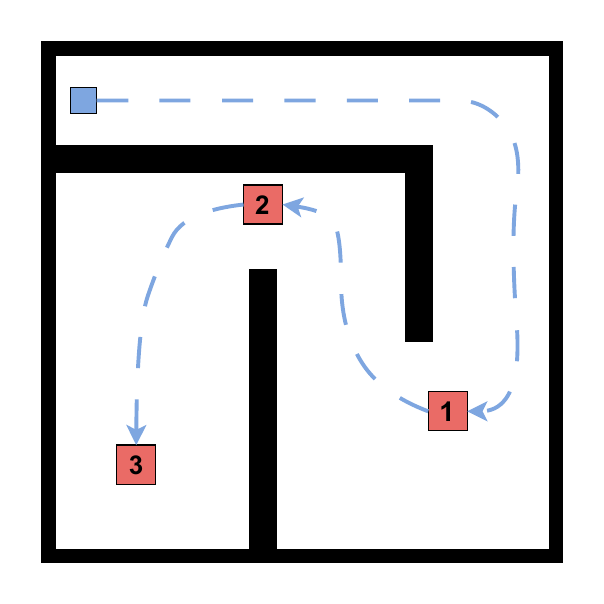}
    \caption{2D Navigation task}
    \label{fig:toy_env}
  \end{subfigure}
  \hfill
  \begin{subfigure}{0.19\textwidth}
    \centering
    \includegraphics[width=\textwidth]{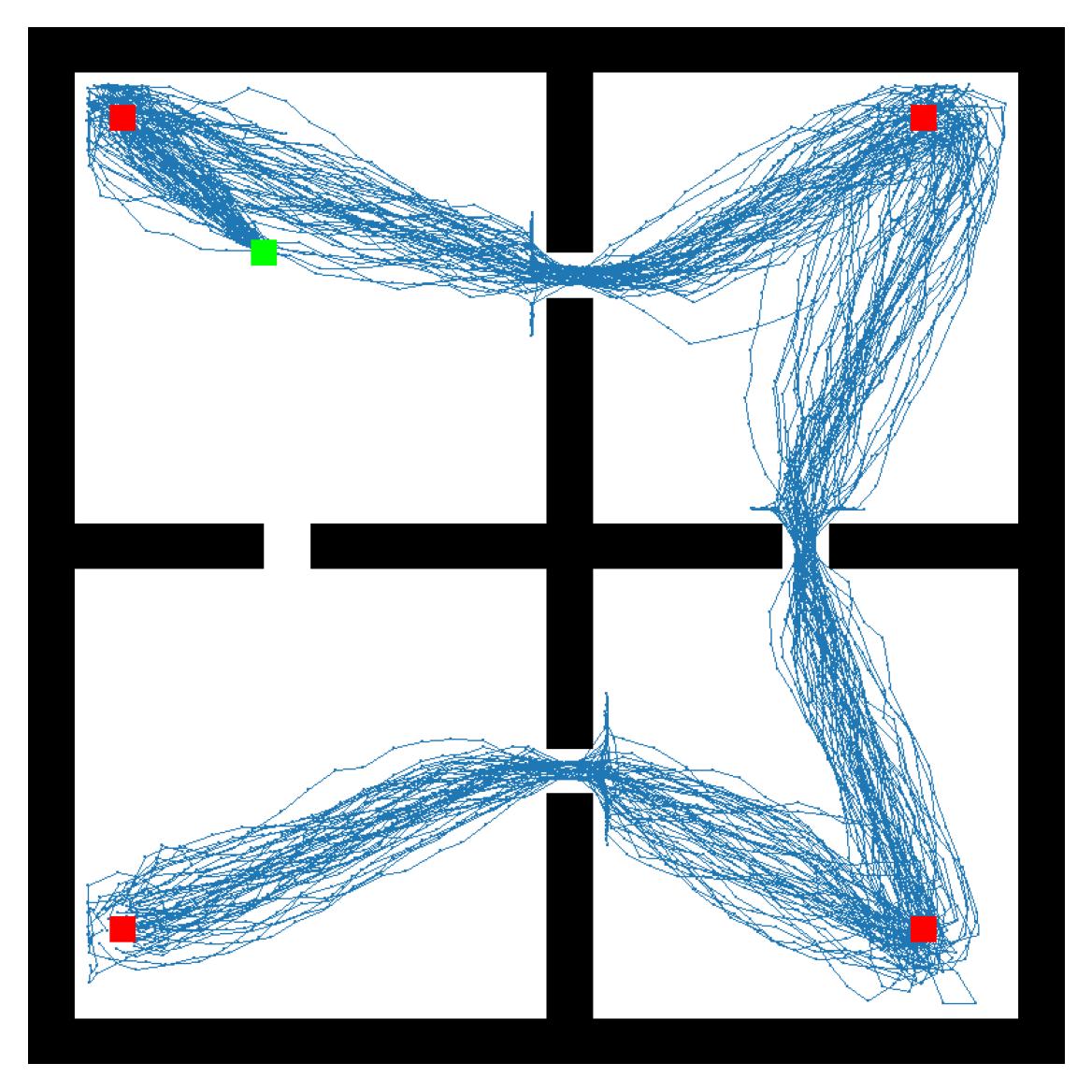}
    \caption{Four room}
    \label{fig:toy_four_room_dataset}
  \end{subfigure}
  \hfill
  \begin{subfigure}{0.19\textwidth}
    \centering
    \includegraphics[width=\textwidth]{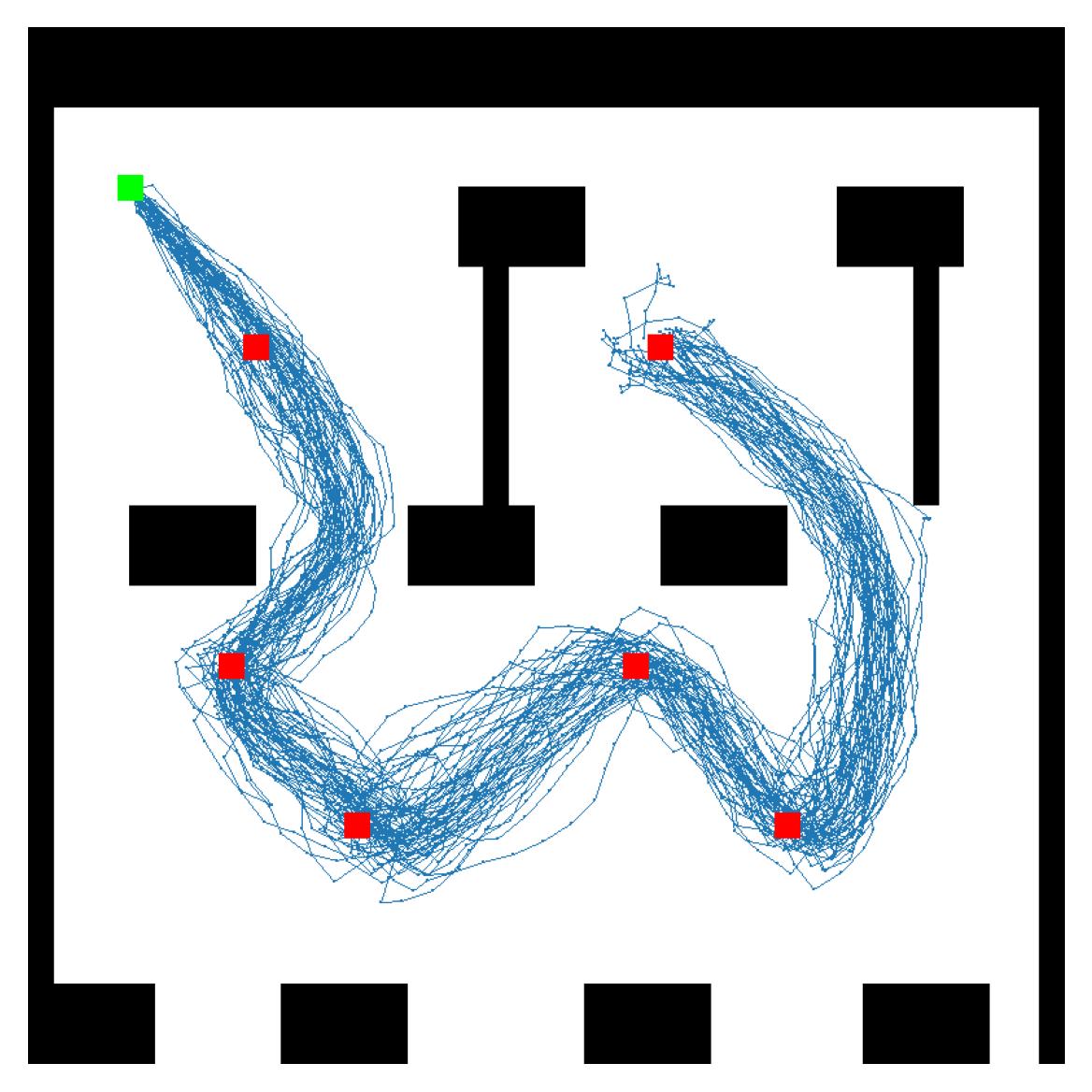}
    \caption{Zigzag}
    \label{fig:toy_zigzag_dataset}
  \end{subfigure}
  \hfill
  \begin{subfigure}{0.19\textwidth}
    \centering
    \includegraphics[width=\textwidth]{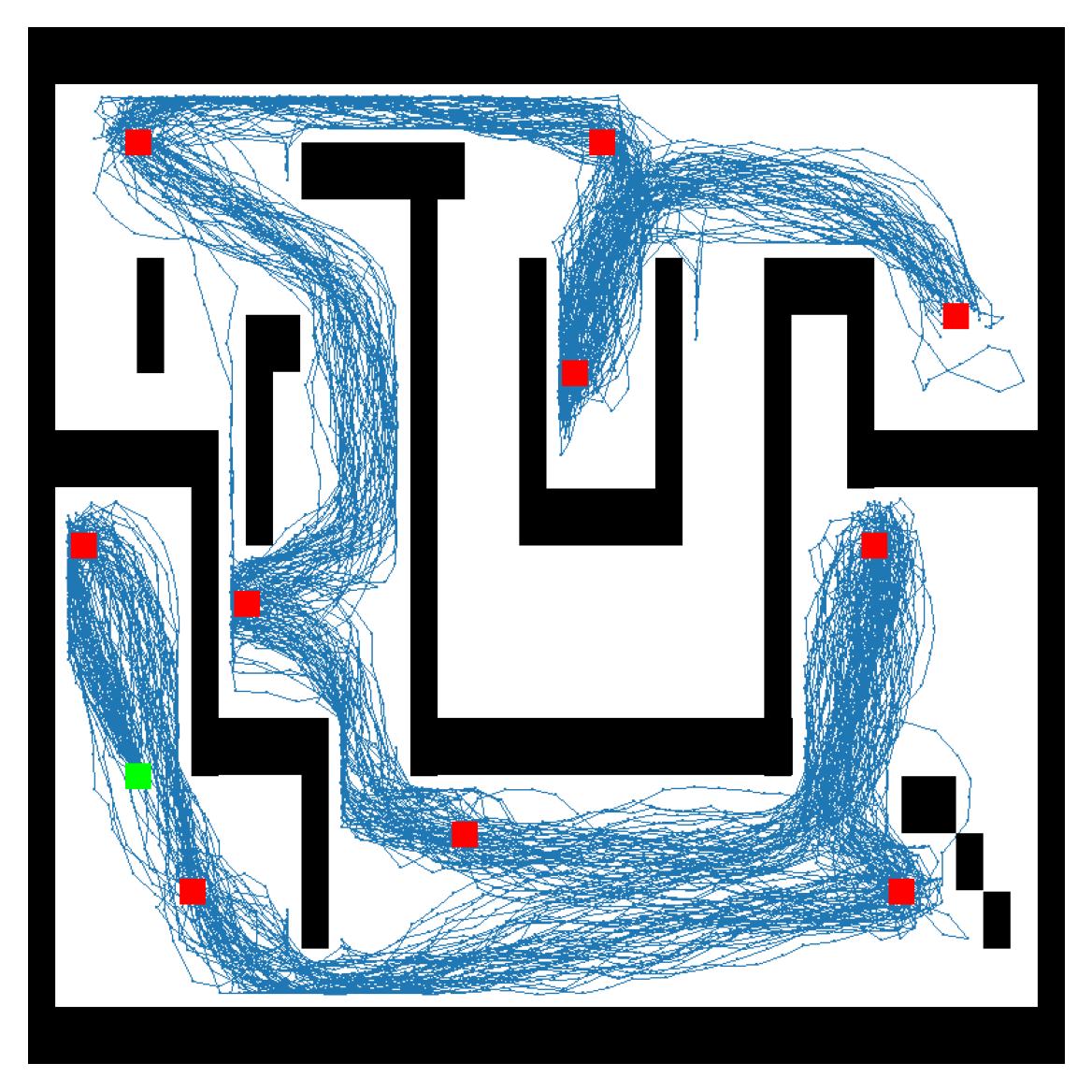}
    \caption{Maze}
    \label{fig:toy_maze_dataset}
  \end{subfigure}
\hfill
  \begin{subfigure}{0.19\textwidth}
    \centering
    \includegraphics[width=\textwidth]{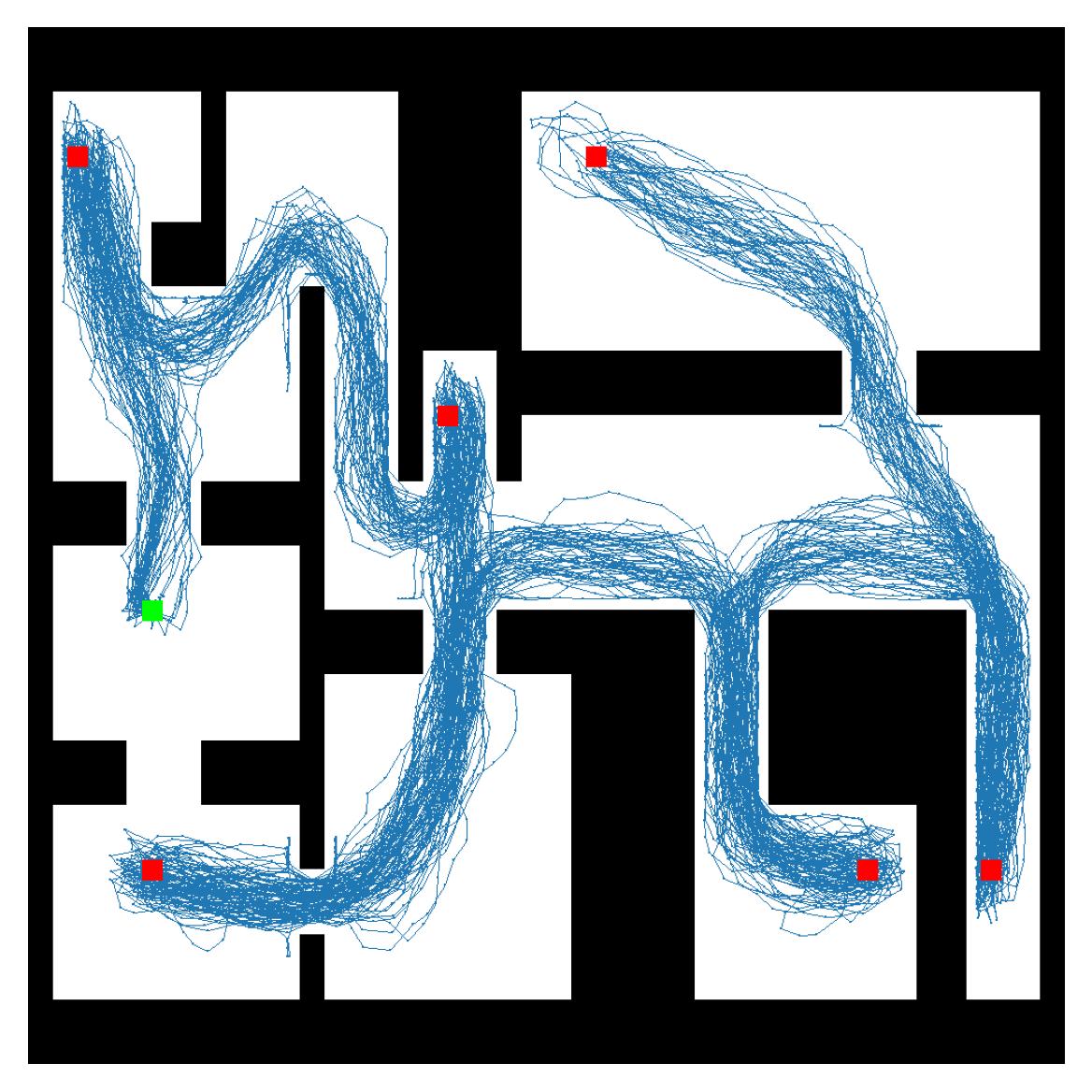}
    \caption{Multiroom}
    \label{fig:toy_multiroom_dataset}
  \end{subfigure}
  \caption{Visualization of 2D navigation environment. (a) Tasks require the agent (blue box) to navigate to reach the goals (red boxes) in a particular order. (b) - (e) Visualizations of all four tasks and the traces of the 50 expert demonstrations within each dataset.}
  \label{fig:toy}
\end{figure*}

\section{Experiments}
\label{sec:experiments}
To better understand how our theoretical insights manifest in practice, we perform experiments in a 2D navigation environment, where we can easily analyze the properties of datasets and policies. We then conduct experiments in a 3D world that require precise execution of a complex task from images to illustrate our findings under real-world conditions.

\subsection{Environments and Algorithm Details}
\label{sec:env_and_algorithm_details}

\textbf{2D navigation environment.}
We consider four tasks in the 2D navigation environment shown in \Cref{fig:toy} where the agent must reach a sequence of goals. Tasks are fully observable, with low-dimensional states containing the agent's $(x,y)$ position, the positions of all goals and whether each goal has already been reached. This simplified setting isolates efficiency gains from PIDM's action decomposition from other gains due to improved representations reported in prior work~\citep{lamb2023guaranteed,koul2023pclast,levine2024multistep}. The agent selects actions in $[-1,1]^2$, and transitions are stochastic due to Gaussian noise $\mathcal{N}(0,0.2)$ added to the actions. For each task, we collect a dataset of 50 demonstrations by a human completing the task, naturally exhibiting action variability, as illustrated in \Cref{fig:toy} (see \Cref{app:toy_details} for more details). 
To study the effect of action variability on BC and PIDM, we additionally evaluate both methods on datasets collected using a deterministic \astar  planner (see \Cref{app:evaluation-deterministic-target-policy}). 

\textbf{3D world.}
To evaluate under realistic conditions, we construct a dataset of human demonstrations in the Dojo practice level of the modern 3D video game \emph{Bleeding Edge}, developed by Ninja Theory. Observations are third-person video frames that are encoded using a pretrained vision model before being passed to BC or PIDM. We use the pretrained ViT-B/16 Theia encoder~\citep{shang2024theia}. The player controls character and camera movement through continuous actions in $[-1,1]^4$. The camera orientation directly affects the agent’s movement, introducing a complex coupling of perception and navigation. State transitions occur asynchronously at 30 FPS, requiring real-time inference, and exhibit random latency and visual artifacts due to deployment on a remote cloud server. We consider the \emph{Tour} task that evaluates navigation accuracy, obstacle avoidance, and interaction with points of interest across $\sim$36-second with 11 milestones to complete (see \Cref{app:xbox_details} for details).

\textbf{Model architecture.}
In the fully observable 2D navigation environment, we use MLPs for the BC policy network and for the PIDM encoder and policy networks.
We use $k=1$ for PIDM. The 3D environment is partially observable, so we use the vision embeddings from three frames spanning the previous second to approximate a single state.
BC conditions on the representations of the current state, whereas the PIDM policy additionally conditions on the representations of a predicted future state and a one-hot encoding of $k$. Following prior work on multi-step IDMs~\citep{lamb2023guaranteed,koul2023pclast}, PIDM is trained with $k \in \{1,6,11,16,21,26\}$ and evaluated with $k=1$ (see \Cref{app:toy_sensitivity_k} for discussion on choice of $k$). For both BC and PIDM, action logits are passed through $\tanh$ activations to enforce the $[-1,1]$ action range.

\textbf{State predictor.}
In the 2D navigation environment, we use a deterministic instance-based learning model~\citep{Keogh2010} as a deterministic state predictor:
\begin{equation}
    \label{eq:instance_state_predictor}
    \planner(s_t) 
= 
    \st_{\tau^\star + k}^{i^\star} 
\; 
    \text{with} 
\; 
    \left(
        \tau^\star, i^\star
    \right)
\defeq
    \arg\min_{\tau, i} \lvert\lvert s_t - \st_\tau^i \rvert\rvert^2
,
\end{equation}
where $s^i_\tau$ denotes the state at time step $\tau$ of demonstration $i$. The predictor retrieves the nearest state in the training set using Euclidean distance and returns the state $k$ steps ahead in the same demonstration. Nearest-neighbour retrieval is restricted to states with the same current goal. In the 3D environment, the state predictor predicts $s^i_{t+k}$ conditioned on time step $t$ and a randomly sampled training demonstration $i$. Despite their simplicity, these state predictors can provide efficiency gains when paired with an IDM.

\textbf{Training details.}
BC and PIDM are trained on batches of \num{4096} samples, consisting of either $(s_t,a_t)$ or $(s_t,a_t,s_{t+k})$ tuples drawn from $\data$. All models are optimised for \num{100000} steps using Adam. Full details on architectures and hyperparameters are provided in \Cref{app:toy_details,app:xbox_details}.

\begin{figure*}[t]
    \centering
    \begin{subfigure}{.245\linewidth}
        \centering
        \includegraphics[width=\linewidth]{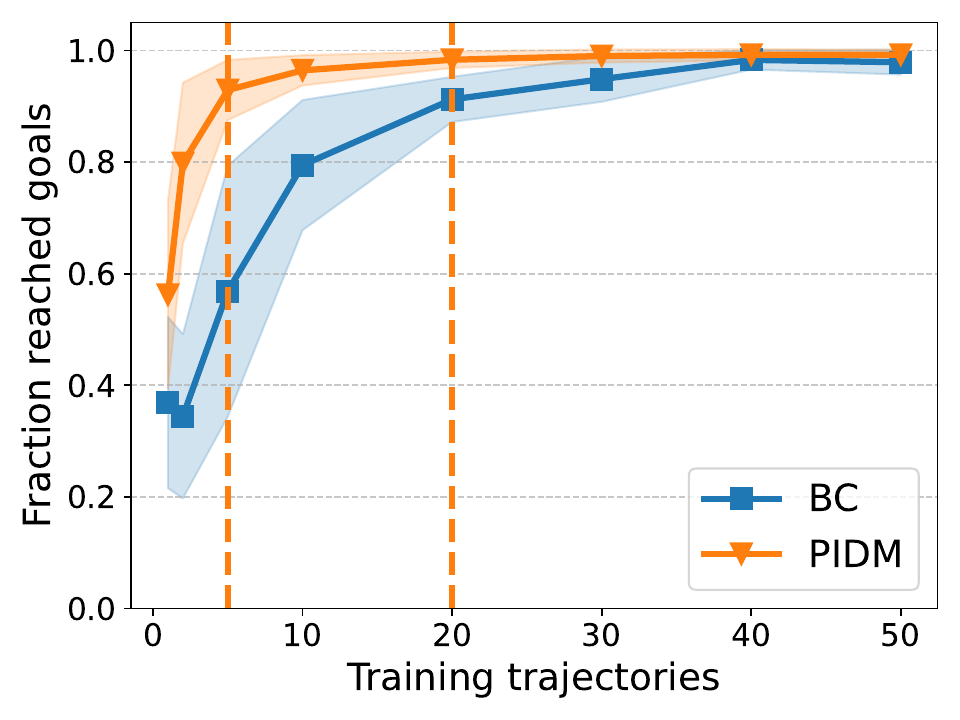}
        \caption{Four room}
        \label{fig:toy_four_room_human_sample_efficiency}
    \end{subfigure}
    \hfill
    \begin{subfigure}{.245\linewidth}
        \centering
        \includegraphics[width=\linewidth]{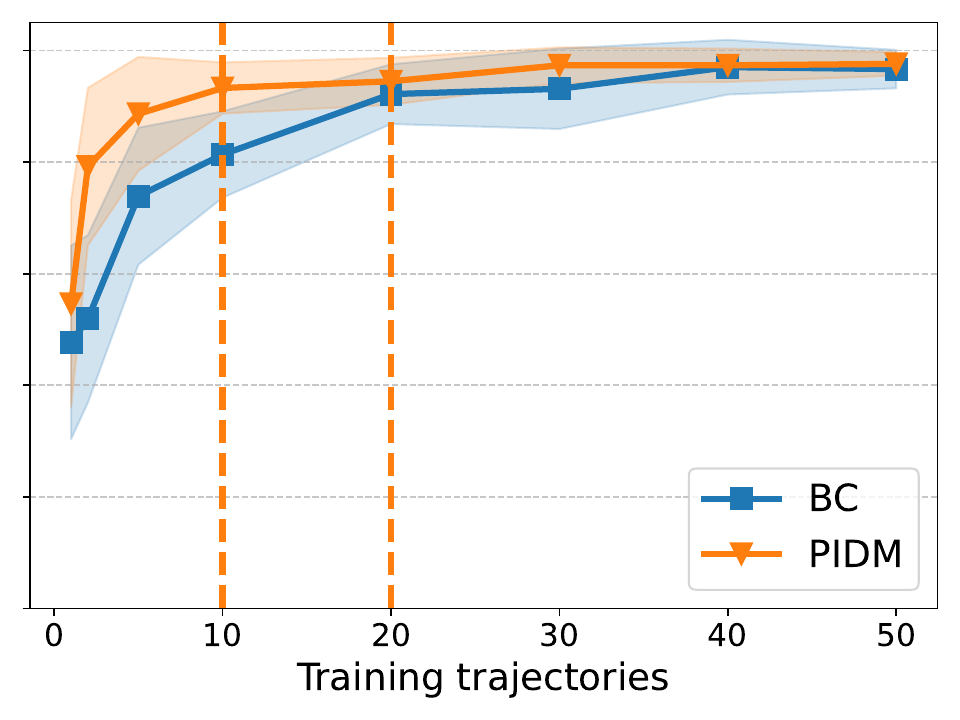}
        \caption{Zigzag}
        \label{fig:toy_zigzag_human_sample_efficiency}
    \end{subfigure}
    \hfill
    \begin{subfigure}{.245\linewidth}
        \centering
        \includegraphics[width=\linewidth]{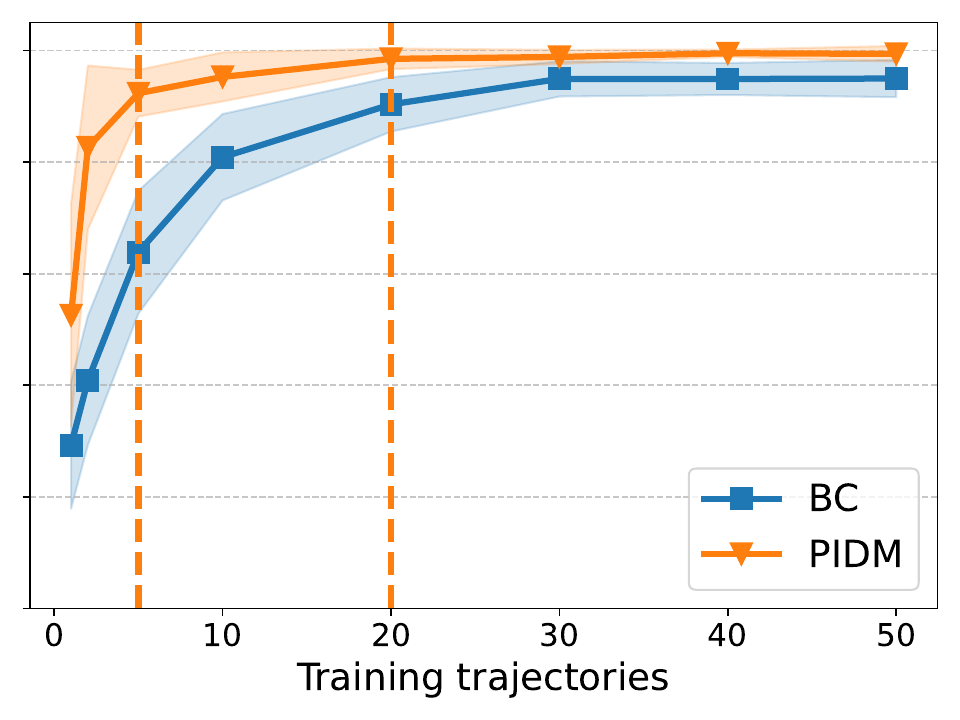}
        \caption{Maze}
        \label{fig:toy_maze_human_sample_efficiency}
    \end{subfigure}
    \hfill
    \begin{subfigure}{.245\linewidth}
        \centering
        \includegraphics[width=\linewidth]{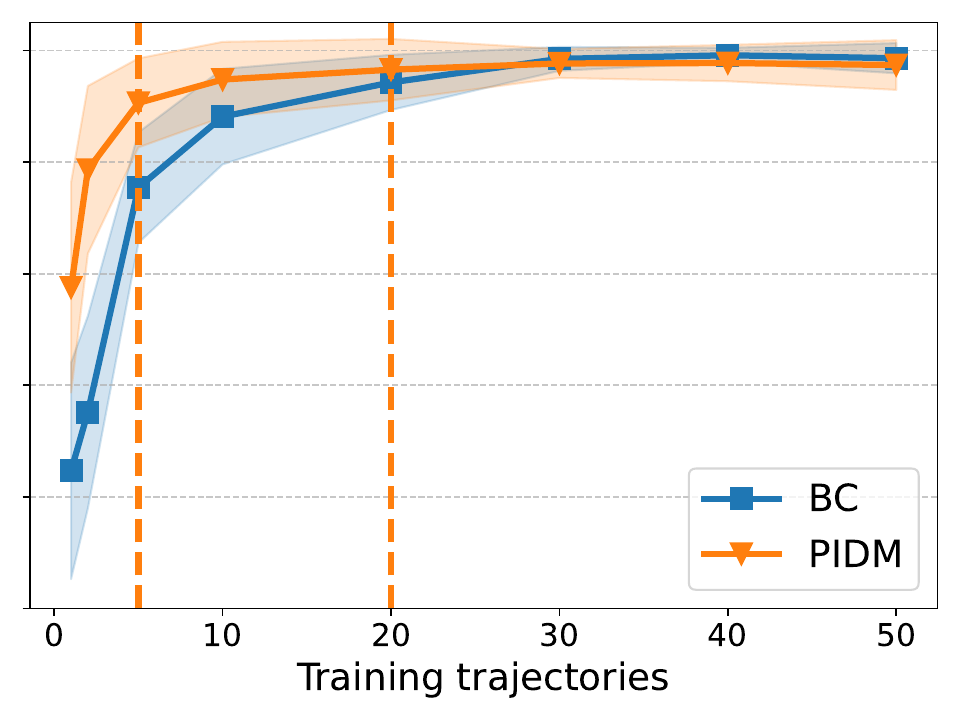}
        \caption{Multiroom}
        \label{fig:toy_multiroom_human_sample_efficiency}
    \end{subfigure}
    \caption{Performance per number of training demonstrations for BC and PIDM in four tasks trained on human datasets. Lines and shading correspond to the average and standard deviation across 20 seeds. We further visualize the number of samples required by PIDM and BC to reach 90\% of the highest achievable performance with vertical dotted lines.}
    \label{fig:toy_human_sample_efficiency}
\end{figure*}

\begin{table}[t]
    \centering
    \caption{Maximum reached goal ratio and sample efficiency ratios of PIDM over BC for 2D navigation tasks and average across tasks.}
    \label{tab:toy_efficiency}
    \resizebox{\linewidth}{!}{
        \begin{tabular}{l S[table-format=1.2] S[table-format=1.2] S[table-format=1.2] S[table-format=1.2] S[table-format=1.2]}
            \toprule
            Task & {Four room} & {Zigzag} & {Maze} & {Multiroom} & {Average} \\
            \midrule
            $\max$ BC $\uparrow$ & 0.98 & 0.97 & 0.95 & 0.99 & {--}\\
            $\max$ PIDM $\uparrow$ & 0.99 & 0.98 & 0.99 & 0.98 & {--}\\
            \midrule
            {$\eta_{\rm{PIDM}}(80\%)$ $\uparrow$} & 4.0 & 2.0 & 5.0 & 2.0 & 3.25\\
            {$\eta_{\rm{PIDM}}(90\%)$ $\uparrow$} & 4.0 & 2.0 & 4.0 & 4.0 & 3.5\\
            {$\eta_{\rm{PIDM}}(95\%)$ $\uparrow$} & 4.0 & 1.33 & 5.0 & 2.0 & 3.08 \\
            \bottomrule
        \end{tabular}
    }
\end{table}

\subsection{Sample Efficiency Gains for 2D Navigation}
\label{sec:toy_efficiency}

While our theory analyzes test-time generalization and training-time sample-efficiency gains separately, the experiments evaluate their combined effect through test-time sample efficiency, measured by task performance as a function of dataset size. For a fixed dataset size, test-time performance depends on both the quality of the learned estimator, determined by sample efficiency, and the generalization gap predicted by the theory, including the tradeoff induced by covariate shift. Therefore, performance-versus-dataset-size curves provide a single empirical measure of the overall benefit predicted by the theory.

We train BC and PIDM on datasets containing varying numbers of demonstrations $(1,2,5,10,20,30,40,50)$. Performance is measured by the fraction of goals reached in the correct order. For each task and dataset size, we train BC and PIDM for 20 random seeds and evaluate four checkpoints (\num{5000}, \num{10000}, \num{50000}, and \num{100000} optimization steps) using 50 rollouts each. We report aggregate results across seeds using the best checkpoint for each task and dataset size (see \Cref{app:best_last_checkpoint}). We summarize efficiency gains using the efficiency ratio
\begin{equation}
    \eta_\text{PIDM}(c)
=
    \frac{
        n(\text{BC}, c)
    }{
        n(\text{PIDM}, c)
    }
,
    \label{eq:efficiency_ratio}
\end{equation}
where $n(A,c)$ denotes the number of demonstrations required for the average performance curve of algorithm $A$ to reach a fraction $c$ of the task's maximum performance.
Thus, $\eta_\text{PIDM}(c)$ measures how many times more samples BC requires than PIDM to achieve that performance level.

\Cref{fig:toy_human_sample_efficiency} shows performance across varying numbers of demonstrations, while \Cref{tab:toy_efficiency} summarizes the corresponding efficiency ratios. Consistent with the analysis in \Cref{sec:theoretical-analysis}, PIDM exhibits substantial gains over BC, requiring up to $5 \times$ fewer demonstrations and roughly $3\times$ fewer on average across tasks. Using the less diverse datasets collected by a deterministic \astar planner further increases PIDM's advantage, as the reduced state-prediction error lowers PIDM prediction error (discussed further in \Cref{sec:toy_state_predictor_bias_analysis}). To verify that these gains arise from the PIDM decomposition rather than the retrieval-based state predictor, we also show that PIDM significantly outperforms a retrieval-based BC variant in \Cref{app:retrieval_bc_comparison}.

\begin{figure*}[t]
    \centering
    \begin{minipage}{0.528\linewidth}
        \begin{subfigure}{0.38\linewidth}
            \centering
            \includegraphics[width=\linewidth,trim=0 0 34.2cm 1.05cm,clip]{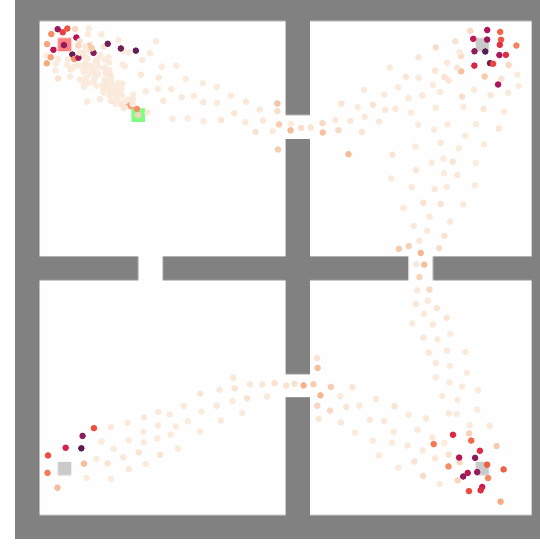}
            \caption{Four room}
            \label{fig:four_room_variance_delta_per_state_main}
        \end{subfigure}
        \hfill
        \begin{subfigure}{0.545\linewidth}
            \centering
            \includegraphics[width=\linewidth,trim=30.2cm 0 0 1.05cm,clip]{images/toy_env/delta_variance_plots/variance_delta_scatter.pdf}
            \caption{Multiroom}
            \label{fig:multiroom_variance_delta_per_state_main}
        \end{subfigure}
        \captionof{figure}{State-wise EPE gaps $\Delta(s)$ (\Cref{eq:delta_per_state}) for Four room and Multiroom datasets. We observe large gaps in states surrounding the goals where human actions are more diverse.}
        \label{fig:toy_datasets_variance_delta_per_state_main}
    \end{minipage}
    \hfill
    \begin{minipage}{0.45\linewidth}
        \begin{subfigure}{0.455\linewidth}
            \centering
            \includegraphics[width=\linewidth]{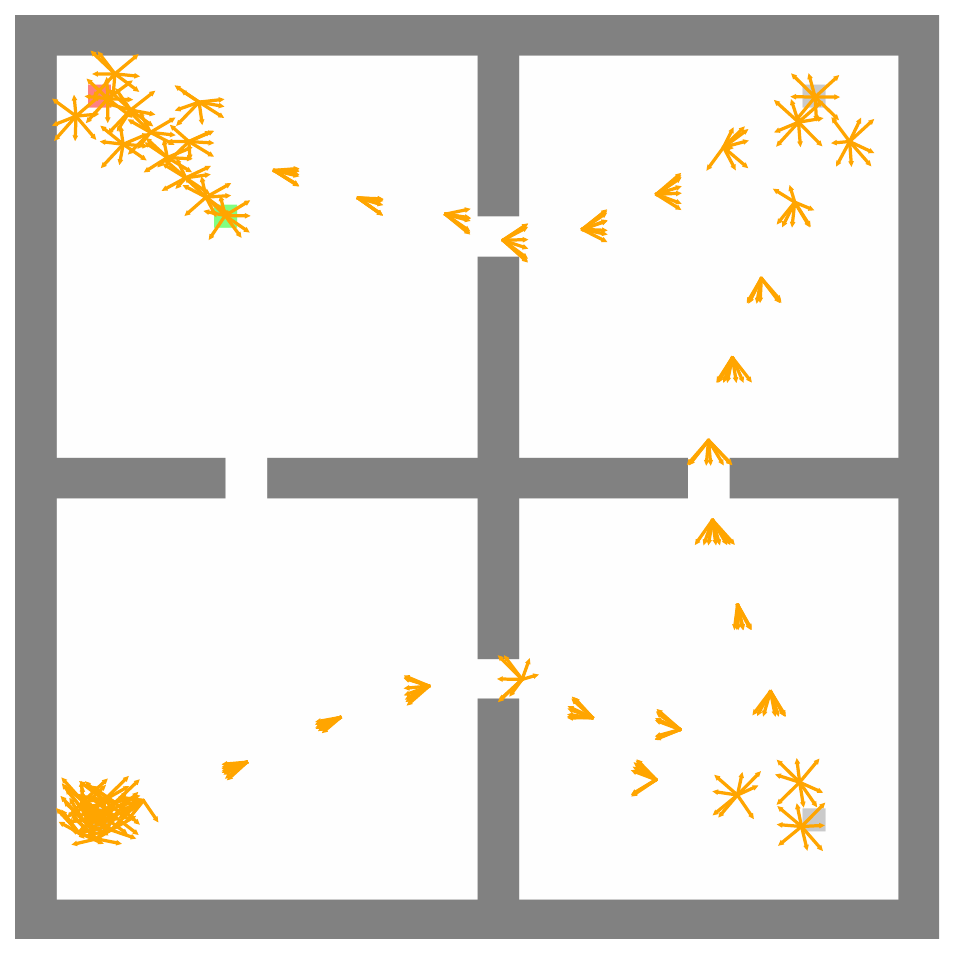}
            \caption{Four room}
            \label{fig:four_room_idm_policy_visualization_main}
        \end{subfigure}
        \hfill
        \begin{subfigure}{0.455\linewidth}
            \centering
            \includegraphics[width=\linewidth]{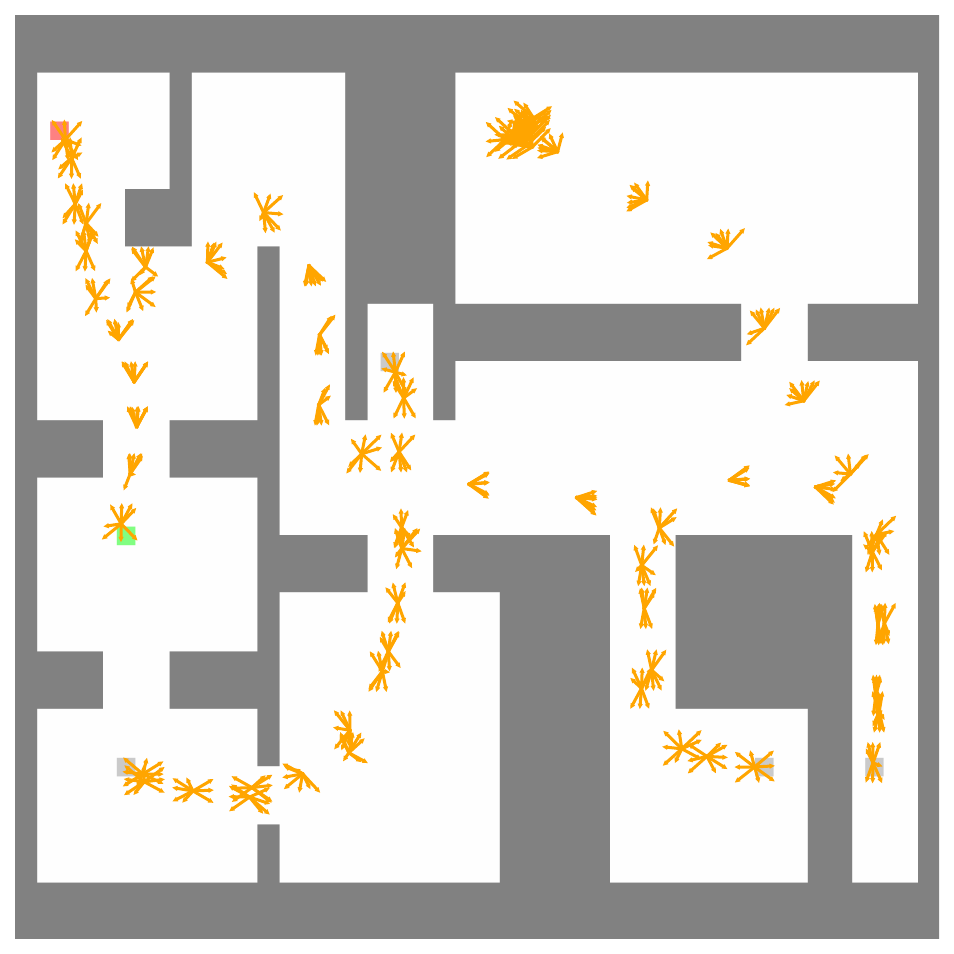}
            \caption{Multiroom}
            \label{fig:multiroom_idm_policy_visualization_main}
        \end{subfigure}
        \captionof{figure}{IDM policies for future states in each cardinal and diagonal direction in Four room and Multiroom. The IDM policy only attends to future states in states with large $\Delta(s)$.}
        \label{fig:idm_policy_visualization_main}
    \end{minipage}
\end{figure*}

\begin{figure*}[t]
    \centering
    \begin{subfigure}{0.2\textwidth}
        \centering
        \includegraphics[width=\textwidth]{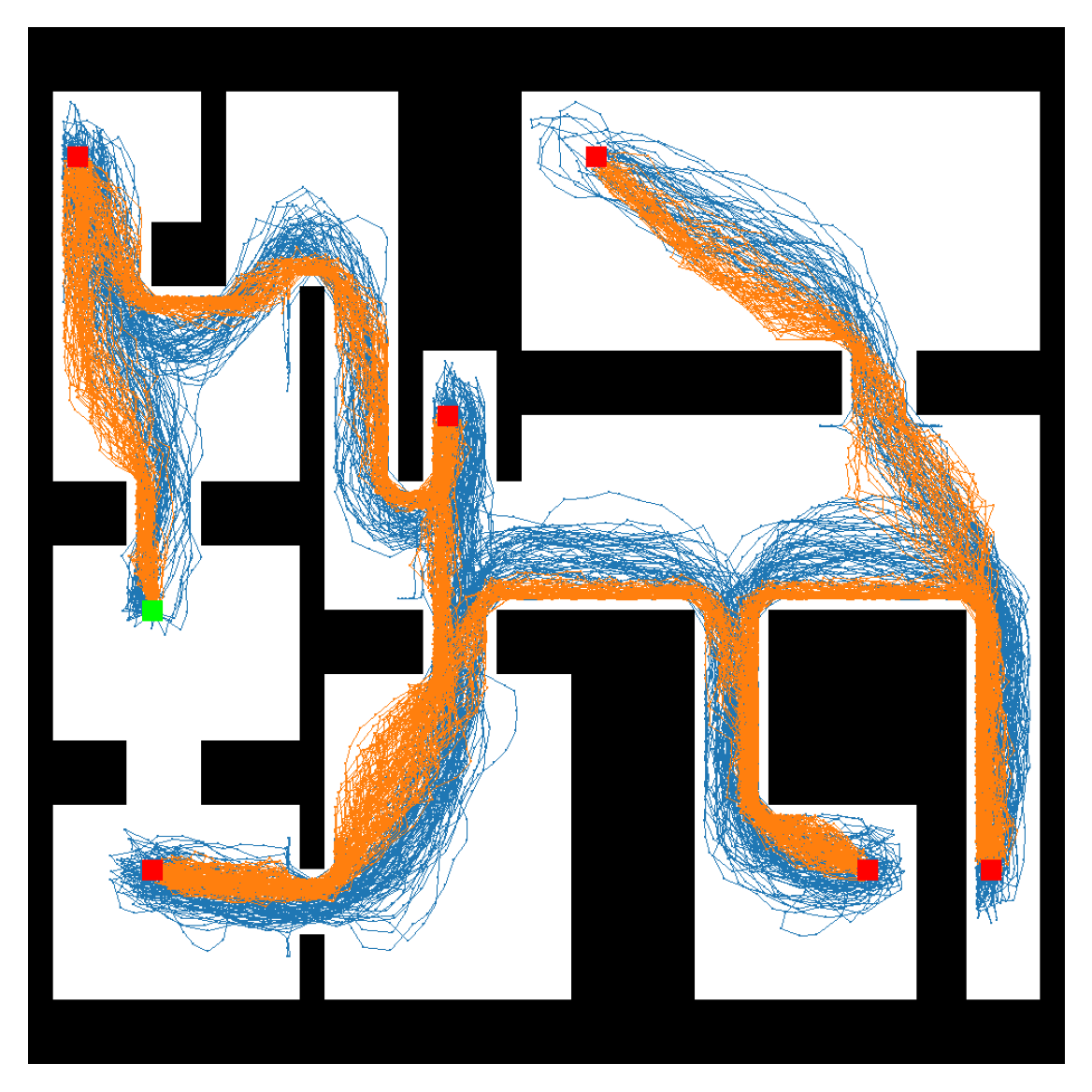}
        \caption{Dataset}
        \label{fig:toy_multiroom_human_planner_dataset}
    \end{subfigure}
    \hfill
    \begin{subfigure}{.26\linewidth}
        \centering
        \includegraphics[width=\linewidth]{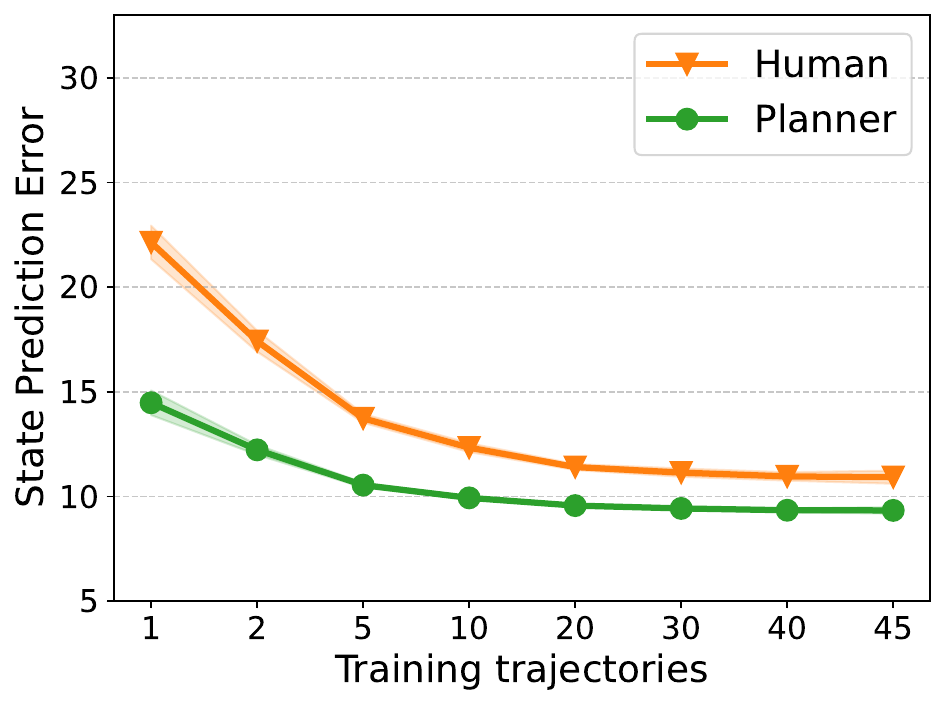}
        \caption{State prediction error}
        \label{fig:toy_pidm_human_planner_state_prediction_error_multiroom}
    \end{subfigure}
    \hfill
    \begin{subfigure}{.26\linewidth}
        \centering
        \includegraphics[width=\linewidth]{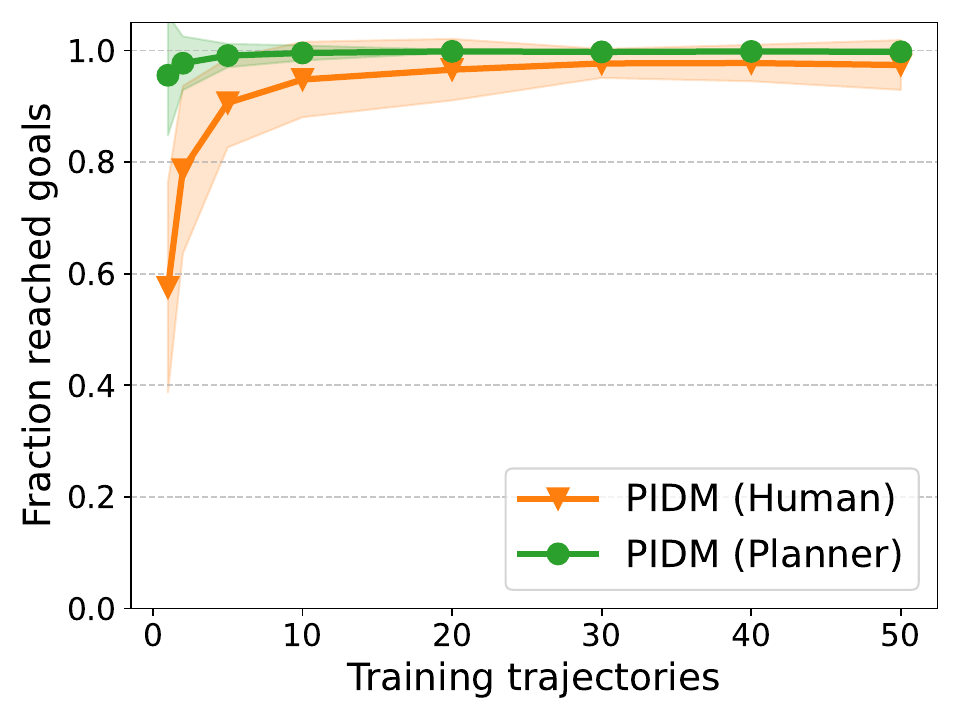}
        \caption{Sample efficiency}
        \label{fig:toy_multiroom_pidm_human_planner_sample_efficiency}
    \end{subfigure}
    \hfill
    \begin{subfigure}{.26\linewidth}
        \centering
        \includegraphics[width=\linewidth]{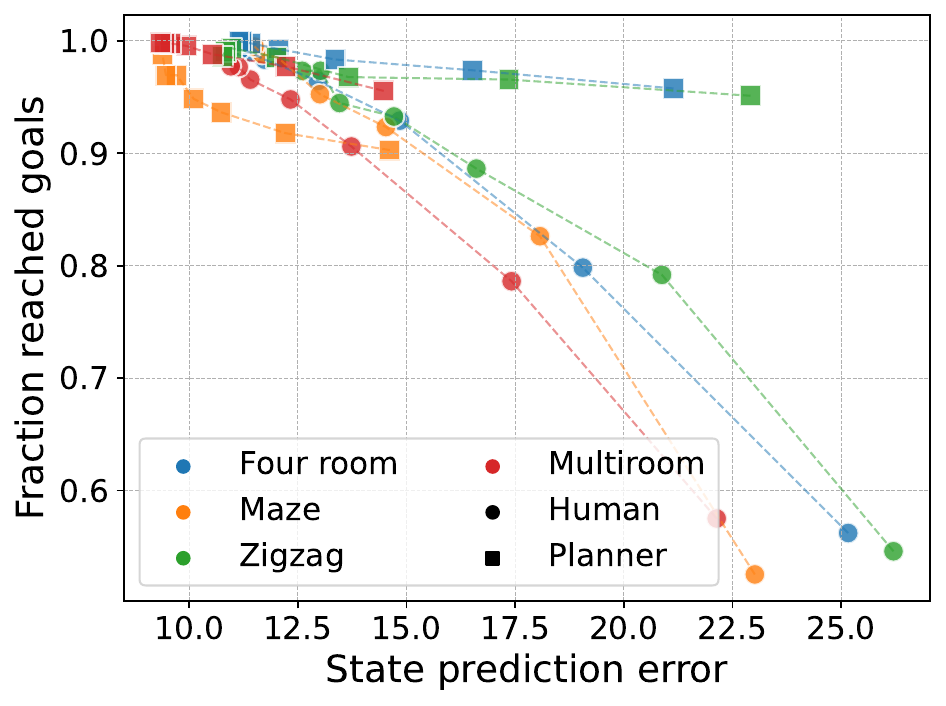}
        \caption{Correlation}
        \label{fig:toy_pidm_human_planner_state_prediction_error_rollout_correlation}
    \end{subfigure}
    \caption{Impact of state predictor bias on PIDM sample efficiency. (a) Visualization of human (blue) and \astar planner (orange) datasets. (b) State prediction error of the instance-based state predictor for both datasets. (c) Sample efficiency curve for PIDM trained on both datasets. (d) Correlation between state prediction error and rollout performance for PIDM on all datasets and tasks.}
\end{figure*}

\subsection{Future Conditioning for Variance Reduction}%
\label{sec:toy_future_conditioning_analysis}
Our theoretical insights of \Cref{theorem:mse-gap} and \Cref{corollary:gap-general-estimator} indicate that states in which future states reduce the uncertainty over actions, as given by $\Delta_{\oplanner}$, are key to realize performance gains of PIDM over BC. 
Where do these states manifest in practice, and what effect do these states have on the learned IDM policy?

To answer these questions, we qualitatively analyse the learned IDM policies for each task and estimate the EPE gap at each state, defined as:
\begin{equation}
    \Delta(\st) 
\defeq
    \var_{\bs_{t+k} \sim \oplanner(\cdot| \st)}
    \left(
        \opredidm(\st, \bs_{t+k})
    \right),
\label{eq:delta_per_state}
\end{equation}
such that $\Delta_{\oplanner} = \E_{\bs \sim d}[\Delta(\bs)]$. 
We emphasize that $\Delta(s)$ is a property of the data distribution and therefore depends only on the task and data collection policy.

To approximate $\Delta(s)$ for continuous states in our 2D environment, we discretize the state space within each human dataset using $K$-means clustering, and estimate the variance of predicted actions over the distribution of future-state clusters, conditioned on each current-state cluster.
\Cref{fig:toy_datasets_variance_delta_per_state_main} visualizes the resulting $\Delta(s)$ estimates for Four room and Multiroom using $K=500$ clusters and $k=1$.

Across tasks, states with large $\Delta(s)$ are concentrated near goal locations that the player has to navigate towards. Near a goal, small deviations in actions become increasingly important, causing the human player to choose different actions in response to different future outcomes and thereby increasing action variability. By \Cref{theorem:mse-gap} and \Cref{corollary:gap-general-estimator}, states with large $\Delta(s)$ correspond to larger potential gains for PIDM over BC, as future conditioning reduces uncertainty over actions. These results suggest that PIDM is particularly effective in goal-adjacent states, where accurate action prediction is most critical for successful task completion.

To understand the impact of these states on the learned IDM policies, we obtain representative states through $K$-means clustering (using $K=75$ clusters for Maze and Multiroom and $K=50$ for Four room and Zigzag) and visualize IDM policies trained on 50 demonstrations. For each cluster centroid, we compute eight future states reachable within $k=1$ step in each cardinal and diagonal direction. We then condition the IDM policy on these future states and visualize the predicted actions as arrows in \Cref{fig:idm_policy_visualization_main}. 

For centroids near goal locations, where $\Delta(s)$ is large (as seen in \Cref{fig:toy_datasets_variance_delta_per_state_main}), the IDM policy exhibits the expected behavior: the predicted actions point towards the queried future states. 
In contrast, for states where $\Delta(s)$ is small, the IDM policy behaves similarly to a BC policy, with predicted actions largely independent of the queried future state. 

This result shows that the IDM policy learns to attend to future states only when they provide useful information for reducing uncertainty over action predictions (i.e. where $\Delta(s)$ is large), corresponding precisely to the states where our theory predicts performance gains for PIDM. 
Conversely, when future information is uninformative (i.e. states where $\Delta(s)$ is small), the IDM policy largely ignores it and therefore remains unaffected by any errors introduced by an approximate state predictor. 

Additional visualizations of EPE gaps and IDM policies for all tasks are provided in \Cref{app:toy_additional_analysis}.

\subsection{Covariate Shift Induced by the State Predictor}
\label{sec:toy_state_predictor_bias_analysis}
\Cref{corollary:gap-general-estimator} further shows that the covariate shift induced by an approximate state predictor is an important factor that can reduce the potential gains of PIDM over BC. How does this affect the performance and sample efficiency of PIDM in practice?

To analyze the impact of the covariate shift, we collect additional datasets with an \astar planner that deterministically selects the optimal action towards the next goal. 
This data collection policy produces less diverse state distributions, as illustrated in \Cref{fig:toy_multiroom_human_planner_dataset} for the Multiroom task. 
We hypothesize that the instance-based state predictor will be more accurate on these less-diverse datasets because nearest neighbor retrieval is more likely to find states close to the queried state. 
\Cref{fig:toy_pidm_human_planner_state_prediction_error_multiroom} confirms this hypothesis, showing that the state-prediction error is substantially lower across all dataset sizes when using demonstrations collected with the \astar planner instead of human demonstrations.

Furthermore, the reduction in state prediction error also improves the sample efficiency of PIDM. As shown in \Cref{fig:toy_multiroom_pidm_human_planner_sample_efficiency}, PIDM trained on the \astar dataset requires $5\times$ fewer demonstrations to reach the 90\% performance threshold than PIDM trained on human demonstrations. 
Finally, \Cref{fig:toy_pidm_human_planner_state_prediction_error_rollout_correlation} shows a clear relationship between state prediction error and evaluation performance: 
lower prediction error corresponds to higher performance, in line with our theoretical analysis.

We refer to \Cref{app:evaluation-deterministic-target-policy} for additional details on the \astar dataset and results across all tasks.

\subsection{Sample Efficiency Gains in a 3D World}

So far, we have demonstrated the theoretically predicted efficiency gains of PIDM over BC and developed intuition for how these models operate in a simplified 2D navigation environment. We now investigate whether, in practice, even a highly simplified state predictor can provide sufficiently useful future conditioning for PIDM to improve sample efficiency over BC. 
To this end, we study the more complex \emph{Tour} task using a simple state predictor that receives the time step as input (see \Cref{sec:env_and_algorithm_details}).
In this task, the agent navigates from image observations in the 3D world of a modern video game that requires real-time inference. 
Success is defined by achieving 11 milestones (see also \Cref{app:tour-env-details}).

To compare agent performance on this task, we train BC and PIDM using 5, 15, 20, 25, and 30 human demonstrations, with 5 random seeds per setting. The final checkpoint of each training run is evaluated over 10 rollouts, measuring the fraction of milestones reached. We report the mean and standard deviation across seed-wise average rollout performances.
\Cref{fig:xbox_highlight_result} shows that PIDM achieves an average success rate of 95\% at the end of training and reaches 87\% success with only 15 demonstrations. In contrast, BC requires 25 demonstrations to achieve an 81\% success rate. Consequently,
$
    \eta_\text{PIDM}(80\%) 
=
    1.66
$,
indicating that BC requires 1.66 times more demonstrations than PIDM to reach a success rate of at least 80\%.
We perform Welch's $t$-test to compare PIDM and BC at each dataset size and find the gains of PIDM to be statistically significant ($p\ll0.001$) for all settings except 5 demonstrations. 

These results demonstrate that PIDM can substantially improve sample efficiency over BC, even in the small-data regime, under complex real-world conditions, and with a highly simplified state predictor. 
Furthermore, we expect these gains to increase when more informative state predictors and/or additional data sources, which can be used to improve either the state-predictor or the IDM policy, become available.

We note that the PIDM state predictor in this experiment is conditioned on the time step $t$ which is not available to either the IDM or BC policies. In preliminary experiments, LSTM-based IDM and BC policies underperformed compared to feedforward policies. This suggests that access to time-step information is not critical for policy performance. Consequently, we use the simpler feedforward architecture for both IDM and BC policies.

\section{Conclusion}

This work analyzes the performance advantages of PIDM as an alternative to BC for offline imitation learning.
Through theoretical analysis and empirical experiments, we shed light onto the advantages of PIDM observed in prior studies: by conditioning on future states, especially in regions of high uncertainty, PIDM increases sample efficiency and reduces action prediction error, but the benefits can be reduced by the covariate shift of an approximate state predictor.
Moreover, we establish conditions under which PIDM is guaranteed to outperform BC. 
Finally, we formally motivate the use of additional data sources when available. 
Empirical results across navigation tasks in 2D and 3D environments confirmed sample efficiency gains, with BC requiring up to $5\times$ more demonstrations than PIDM to achieve comparable performance. 
Interestingly, qualitative analysis showed that learned PIDM policies attend to future states only when they provide informative context for reducing prediction variance. 
Altogether, this work provides a principled explanation for PIDM's effectiveness and 
offers insights that pave the way for more efficient imitation learning methods that leverage state prediction and future conditioning.

For clarity and tractability, we make several simplifying assumptions, including squared loss and point estimators, i.i.d. sampling, shared parameterization, and sufficiently large sample sizes. Despite these simplifications, our theoretical findings are consistent with both our experiments and prior empirical work~\cite{du2023learning,tian2025predictive,xie2025latent,pai2026mimic,li2026causal,ye2026worldactionmodelszeroshot}, which report PIDM outperforming BC even with correlated data, across a range of policy classes, including diffusion and transformer models, and across diverse data regimes.
These results suggest that the performance tradeoff underlying the efficiency gains is a feature of the PIDM architecture rather than an artifact of our simplifying assumptions. \Cref{app:extensions} sketches how the analysis may extend to broader settings, leaving a formal treatment of these extensions to future work.

\section*{Acknowledgements}
We thank Dave Bignell, Sarah Parisot, Tim Pearce, and Raluca Stevenson for their insightful discussions and feedback throughout the project. We further thank Tarun Gupta, Shu Ishida, and Marko Tot for their contributions to early versions of this project.

\section*{Impact Statement}

This paper presents new insights into offline imitation learning approaches through theoretical and empirical investigation. The goal of this analysis is to shed light onto previously unexplained findings with the goal of advancing the general fields of machine learning and imitation learning. There are many potential societal consequences of our work, none which we feel must be specifically highlighted here.

\bibliography{references}
\bibliographystyle{icml2026}

\clearpage
\appendix
\onecolumn

\crefalias{section}{appendix}
\crefalias{subsection}{appendix}

\section{Proofs}
\label{app:proofs}
\setcounter{theorem}{0} 
\setcounter{corollary}{0} 
\setcounter{lemma}{0}

\subsection{Notation}
To make notation lighter, we define these shortcuts:
\begin{align}
    \E_{d} [\cdot]
&\defeq
    \E_{\bs_t \sim d} [\cdot]
\\
    \E_{\oplanner} [\cdot \mid \bs_t]
&\defeq
    \E_{\bs_{t+k} \sim \oplanner(\cdot | \bs_t)} [\cdot]
\\
    \E_{\approxplanner} [\cdot \mid \bs_t]
&\defeq
    \E_{\bs_{t+k} \sim \approxplanner(\cdot | \bs_t)} [\cdot]
\\
    \E_{\opol} [\cdot \mid \bs_t]
&\defeq
    \E_{\ba_t \sim \opol(\cdot \mid \bs_t)} [\cdot]
\\
    \E_{\opolidm} [\cdot \mid \bs_t, \bs_{t+k}]
&\defeq
    \E_{\ba_t \sim \opolidm(\cdot | \bs_t, \bs_{t+k})} [\cdot]
,
\end{align}
and the same for the variance terms.
Moreover, we extend the same notation to expectations over multiple variables, like:
\begin{equation}
    \E_{d, \oplanner}
    [\cdot]
\defeq
    \E_{d}
    [
        \E_{\oplanner}
        [\cdot \mid \bs_t]
    ]
=
    \E_{\bs_t \sim d, \bs_{t+k} \sim \oplanner(\cdot | \bs_t)}
    [\cdot]
.
\end{equation}
Finally, we skip the policy distribution in the irreducible noise terms:
\begin{align}
    \var(\ba_t \mid \bs_t)
&\defeq
    \var_{\opol}(\ba_t \mid \bs_t)
\defeq
    \E_{\opol}
        \left[ 
                \left(
                    \ba_t 
                    - 
                    \E_{\opol}[\ba_t \mid \bs_t]
                \right)^2 
            \big\rvert
                \bs_t
        \right] 
,
\\
    \var(\ba_t \mid \bs_t, \bs_{t+k})
&\defeq
    \var_{\opolidm}(\ba_t \mid \bs_t, \bs_{t+k})
\defeq
        \E_{\opolidm}
        \left[ 
                \left(
                    \ba_t 
                    - 
                    \E_{\opolidm}[\ba_t \mid \bs_t, \bs_{t+k}]
                \right)^2 
            \big\rvert
                \bs_t, \bs_{t+k}
        \right] 
.
\end{align}

\subsection{Proof of \Cref{theorem:mse-gap}}

\begin{theorem}
For optimal predictors $\opredbc$ and $\opredidm$: 
\begin{equation}
    \Delta_{\oplanner}
=
    \E_{\bs_t \sim d}
    \left[
        \var_{\bs_{t+k} \sim \oplanner(\cdot| \bs_t)}
        \left(
            \opredidm(\bs_t, \bs_{t+k})
        \right)
    \right]
\ge
    0
.
\label{eq:gap-optimal-as-variance-term-bis}
\end{equation}
\end{theorem}

\begin{IEEEproof}
Recall that the prediction error (w.r.t. to the expert actions) for each of the optimal predictors is given by:
\begin{align}
    \epe(\opredbc) 
&
\defeq
    \E_{d, \opol}
    \left[ 
            \left( 
                \ba_t - \opredbc(\bs_t)
            \right)^2
        \mid
            \bs_t
    \right]
,
\\
    \epe(\opredidm; \oplanner) 
&
\defeq
    \E_{d, \oplanner, \opolidm}
    \left[ 
            \left( 
                \ba_t
                -
                \opredidm(\bs_t, \bs_{t+k})
            \right)^2 
        \mid
            \bs_t, \bs_{t+k}
    \right]
.
\end{align}
where $\epe(\cdot; p)$ highlights that the expected predicted error for IDM is computed using future states sampled from $p$.
All random variables follow the same distributions throughout the remainder of the proof: 
\begin{align}
    \bs_t 
&\sim   
    d
, \\
    \ba_t 
&\sim 
    \opol(\cdot \mid \bs_t)
\quad \text{or} \quad
    \ba_t 
\sim 
    \opolidm(\cdot \mid \bs_t, \bs_{t+k})
, \\
    \bs_{t+k} 
&\sim 
    \oplanner(\cdot \mid \bs_t)
.
\end{align}

We can rewrite the EPE by using iterated expectation and replacing the definitions of optimal estimators:
\begin{align}
    \epe(\opredbc) 
&= 
    \E_{d}
        \left[ 
            \E_{\opol}
            \left[ 
                    \left(
                        \ba_t 
                        - 
                        \E_{\opol} \left[ \ba_t \mid \bs_t \right]
                    \right)^2 
                \mid
                    \bs_t
            \right] 
        \right]
\notag\\
&= 
    \E_{d}\left[ \var(\ba_t \mid \bs_t) \right]    
\label{eq:mse-bc}
\\
    \epe(\opredidm ; \oplanner) 
&= 
    \E_{d, \oplanner}
    \left[ 
        \E_{\opolidm}
        \left[ 
                \left(
                    \ba_t 
                    - 
                    \E_{\opolidm}[\ba_t \mid \bs_t, \bs_{t+k}]
                \right)^2 
            \mid
                \bs_t, \bs_{t+k}
        \right] 
    \right]
,
\notag\\
&= 
    \E_{d, \oplanner}
    \left[
            \var(\ba_t \mid \bs_t, \bs_{t+k})
    \right] 
.
\label{eq:mse-pidm}
\end{align}
We want to connect \eqref{eq:mse-bc} with \eqref{eq:mse-pidm}. 
First, we apply the law of total variance to $\var(\ba_t \mid \bs_t)$:
\begin{equation}
    \var(\ba_t \mid \bs_t) 
= 
    \E_{\oplanner} 
    \left[ 
        \var(\ba_t \mid \bs_t, \bs_{t+k}) 
        \mid
            \bs_t
    \right]
    + 
    \var_{\oplanner}
    \left(
            \E_{\opolidm}[\ba_t \mid \bs_t, \bs_{t+k}]
        \mid
            \bs_t
    \right)
.
\end{equation}
Second, we take the expectation over $\bs_t \sim d$:
\begin{equation}
    \E_{d}[\var(\ba_t \mid \bs_t)]
= 
    \E_{d}
    \left[ 
        \E_{\oplanner} 
        \left[ 
                \var(\ba_t \mid \bs_t, \bs_{t+k}) 
            \mid
                \bs_t
        \right]
    \right]
  + 
  \E_{d}
  \left[ 
        \var_{\oplanner}
        \left(
            \E_{\opolidm}[\ba_t \mid \bs_t, \bs_{t+k}]
        \right)
    \mid
        \bs_t
  \right]
.
\end{equation}
Third, we simplify the first term of the r.h.s.:
\begin{equation}
    \E_{d}
    \left[
        \E_{\oplanner}
        \left[
            \var(\ba_t \mid \bs_t , \bs_{t+k})
            \mid
                \bs_t
        \right]
    \right]
=
    \E_{d, \oplanner}
    \left[ 
            \var(\ba_t \mid \bs_t, \bs_{t+k})
        \mid
            \bs_t
    \right]
.
\end{equation}
Finally, we have:
\begin{equation}
    \E_{d}[\var(\ba_t \mid \bs_t)]
= 
    \E_{d, \oplanner}
    \left[
        \var(\ba_t \mid \bs_t, \bs_{t+k})
        \mid
            \bs_t
    \right]
    + 
    \E_{d}
    \left[ 
        \var_{\oplanner}
        \left(
            \E_{\opolidm}[\ba_t \mid \bs_t, \bs_{t+k}]
        \right)
        \mid
            \bs_t
    \right]
.
\end{equation}
Now, we can easily compute the performance gap between the MSEs of both estimators:
\begin{align}
    \epe(\opredbc) 
    - 
    \epe(\opredidm; \oplanner)
&= 
    \E_{d}[\var(\ba_t \mid \bs_t)] 
    - 
    \E_{d, \oplanner}[\var(\ba_t \mid \bs_t, \bs_{t+k})]
\notag\\
&= 
    \E_{d}
    \left[ 
        \var_{\oplanner}
        \left(
            \E_{\opolidm}
            \left[
                \ba_t \mid \bs_t , \bs_{t+k}
            \right]
        \right) 
        \mid
            \bs_t
    \right]
\notag\\
&= 
    \E_{d}
    \left[ 
        \var_{\oplanner}
        \left(
            \opredidm
            \left(
                \bs_t , \bs_{t+k}
            \right)
        \right) 
        \mid
            \bs_t
    \right]
\notag\\
&= 
    \E_{\bs_t \sim d}
    \left[
        \var_{\bs_{t+k} \sim \oplanner(\cdot| \bs_t)}
        \left(
            \opredidm(\bs_t, \bs_{t+k})
        \right)
    \right]
.
\label{eq:Delta-from-irreducible-variances}
\end{align}
\end{IEEEproof}

\subsection{Proof of \Cref{proposition:same-optimal-predictor}}
\label{app:proposition:same-optimal-predictor}

\begin{proposition}
\label{proposition:same-optimal-predictor}

Let $\mathcal P$ denote a family of state-predictor distributions over future states conditioned on the current state.
For any two future-state distributions
$
\oplanner, \approxplanner \in \mathcal P
$,
define the population losses of the IDM predictor under the same IDM policy $\opolidm$:
\begin{align}
    \loss_{\oplanner}(f)
&\defeq
    \E_{d, \oplanner, \opolidm}
\left[
    \left(
        \ba_t
        -
        f(\bs_t,\bs_{t+k})
    \right)^2
\right],
\\
\loss_{\approxplanner}(f)
&\defeq
\E_{d, \approxplanner, \opolidm}
\left[
    \left(
        \ba_t
        -
        f(\bs_t,\bs_{t+k})
    \right)^2
\right].
\end{align}
Let the optimal predictors be given by:
\begin{align}
\pred^\star_{\oplanner}
&\defeq
\arg\min_f \loss_{\oplanner}(f),
\\
\pred^\star_{\approxplanner}
&\defeq
\arg\min_f \loss_{\approxplanner}(f).
\end{align}
Assume that the support of $\approxplanner$ is contained in that of $\oplanner$:
\begin{equation}
    \supp(\approxplanner(\cdot \mid s_t))
    \subseteq
    \supp(\oplanner(\cdot \mid s_t)),
    \;
    \forall s_t \in \St.
\end{equation}
Then:
\begin{align}
    \pred^\star_{\oplanner}(s_t,s_{t+k})
=
    \pred^\star_{\approxplanner}(s_t,s_{t+k})
=
    \E_{\opolidm}[\ba_t \mid s_t, s_{t+k}]
\;
    \text{ a.s. } 
    \forall s_t \in \St,
    \forall s_{t+k} \in \supp(\approxplanner(\cdot \mid s_t))
.
\end{align}

\end{proposition}
\begin{IEEEproof}
This is a standard result, we include the proof for completeness.
Let's expand the conditional expectation over the policy:
\begin{align}
    \E_{\opolidm}
    \left[
        \left(
            \ba_t
            -
            f(\bs_t,\bs_{t+k})
        \right)^2
        \mid
            \bs_t, \bs_{t+k}
    \right]
=
    \Var(
        \ba_t
        \mid
        \bs_t,\bs_{t+k}
    )
    +
    \left(
        f(\bs_t,\bs_{t+k})
        -
        \E_{\opolidm}
        \left[
            \ba_t
            \mid
            \bs_t,\bs_{t+k}
        \right]
    \right)^2
.
\end{align}

The first term on the right side is the irreducible conditional variance, which is independent of $f$.
Therefore:
\begin{align}
    \pred^\star_{\oplanner}
&=
    \arg\min_f
    \E_{d, \oplanner}
    \left[
        \left(
            f(\bs_t,\bs_{t+k})
            -
            \E_{\opolidm}[
                \ba_t
                \mid
                \bs_t,\bs_{t+k}
            ]
        \right)^2
    \right]
,
\\
    \pred^\star_{\approxplanner}
&=
    \arg\min_f
    \E_{d, \approxplanner}
    \left[
        \left(
            f(\bs_t,\bs_{t+k})
            -
            \E_{\opolidm}[
                \ba_t
                \mid
                \bs_t,\bs_{t+k}
            ]
        \right)^2
    \right]
.
\end{align}

Both objectives are minimized pointwise by
\begin{equation}
    f^\star(s_t, s_{t+k})
=
    \E_{\opolidm}
    \left[
        \ba_t
        \mid
        s_t, s_{t+k}
    \right]
,
\end{equation}
and this is the unique solution when the weight $\approxplanner(s_{t+k} \mid s_t)$ is positive only when 
$\oplanner(s_{t+k} \mid s_t) > 0$.
\end{IEEEproof}

\subsection{Proof of \Cref{corollary:gap-general-estimator}}

\begin{corollary}
\label{corollary:gap-general-estimator-bis}
Under Assumptions \ref{ass:support}--\ref{ass:iid}, let $\predmun$ and $\predxim$ denote the BC and IDM predictors learned from $\datadist_n$ and $\approxdatadist$, respectively.
Define the variance, bias, and irreducible-noise gaps under covariate shift:
\begin{align}
    \delta
&
\defeq
    \E_{\bs_t \sim d}
    \Big[
        \var_{\datadist_n}
        \left(
            \predmun
            \left(
                \bs_t
            \right)
        \right)
        -
        \E_{\bs_{t+k}\sim \oplanner(\cdot | \bs_t)}
        \left[
            w(\bs_t, \bs_{t+k})
            \var_{\approxdatadist}
            \left(
                \predxim
                \left(
                    \bs_t, \bs_{t+k}
                \right)
            \right)
        \right]
    \Big]
,
\label{eq:estimators-var-gap-bis}
\\
    \beta
&
\defeq
    \:
    \biasfmusq
    -
    \biasfxisq
,
\label{eq:estimators-bias-gap-bis}
\\
    \gamma
&\defeq
    \E_{
        \substack{
            \bs_t \sim d, \\
            \bs_{t+k} \sim \oplanner(\cdot | \bs_t)}
        }
    \left[
        \left(
            1 - w(\bs_t,\bs_{t+k})
        \right)
        \var(\ba_t \mid \bs_t, \bs_{t+k})
    \right]
\label{eq:covariate-shift-bis}
\end{align}
with bias terms defined as:
\begin{align}
    \biasfmusq
&
\defeq
\:
    \E_{\bs_t \sim d}
    \left[
        \left(
            \E_{\datadist_n}\left[ \predmun(\bs_t) \right]
            -
            \opredbc(\bs_t)
        \right)^2
    \right],
\\
    \biasfxisq
&
\defeq
\:
        \E_{
            \substack{
                \bs_t \sim d,\\
                \bs_{t+k} \sim \oplanner(\cdot|\bs_t)
            }
        }
        \bigg[
            w(\bs_t, \bs_{t+k})
            \Big(
                \E_{\approxdatadist}\left[ \predxim(\bs_t, \bs_{t+k}) \right]
                -
                \opredidm(\bs_t, \bs_{t+k})
            \Big)^2
        \bigg]
.
\end{align}
Then, the predicted error gap is given by:
\begin{equation}
    \widehat{\Delta}_{\approxplanner}
\defeq
    \epe
    \left(
        \predmun
    \right)
    -
    \epe
    \left(
        \predxim
        ;
        \approxplanner
    \right)
=
    \Delta_{\oplanner}
    +
    \delta
    +
    \beta
    +
    \gamma
.
\label{eq:gap-general-estimator-bis}
\end{equation}
\end{corollary}

\begin{IEEEproof}
All random variables follow the same distributions, given by the test-time data distribution, throughout the remainder of the proof: 
\begin{align}
    \bs_t 
&\sim   
    d
, \\
    \ba_t 
&\sim 
    \opol(\cdot|\bs_t)
\quad \text{or} \quad
    \ba_t 
\sim 
    \opolidm(\cdot|\bs_t, \bs_{t+k})
, \\
    \bs_{t+k} 
&\sim 
    \approxplanner(\cdot|\bs_t)
.
\end{align}
Moreover, $\datadist_n$ and $\approxdatadist$ follow the distribution of datasets induced by $\datadist$ and $\datadist$ with $\oplanner$, respectively.

The EPE can be expressed as the sum of the irreducible variance and the estimator's own variance and bias. We do the derivation here for completeness:
\begin{align}
    \epe(\predmun) 
&=
    \E_{d, \opol, \datadist_n}
    \left[ 
            \left( 
                \ba_t - \predmun(\bs_t)
            \right)^2 
        \mid 
            \bs_t
    \right]
\notag\\
&=
    \E_{d, \opol, \datadist_n}
    \left[ 
            \left( 
                \ba_t - \predmun(\bs_t) + \opredbc(\bs_t)  - \opredbc(\bs_t)
            \right)^2 
        \mid 
            \bs_t
    \right]
\notag\\
&=
    \E_{d, \opol, \datadist_n}
    \left[ 
            \left( 
                \left(
                    \ba_t - \opredbc(\bs_t) 
                \right)
                + 
                \left(
                    \opredbc(\bs_t)  - \predmun(\bs_t)
                \right)
            \right)^2 
        \mid 
            \bs_t
    \right]
\notag\\
&=
    \E_{d, \opol}
    \left[ 
            \left(
                \ba_t - \opredbc(\bs_t) 
            \right)^2
        \mid 
            \bs_t
    \right]
    + 
    \E_{d, \datadist_n}
    \left[ 
            \left(
                \opredbc(\bs_t)  - \predmun(\bs_t)
            \right)^2 
        \mid 
            \bs_t
    \right]
\notag\\
&\quad
    +
    2
    \E_{d, \opol, \datadist_n}
    \left[ 
            \left(
                \ba_t - \opredbc(\bs_t) 
            \right)
            \left(
                \opredbc(\bs_t)  - \predmun(\bs_t)
            \right)
        \mid 
            \bs_t
    \right]
.
\label{eq:cross-term}
\end{align}
The first term is the expected conditional variance:
\begin{align}
    \E_{d, \opol}
    \left[ 
            \left(
                \ba_t - \opredbc(\bs_t) 
            \right)^2
        \mid
            \bs_t
    \right]
&=
    \E_{d}
    \left[ 
        \E_{\opol}
        \left[
                \left(
                    \ba_t
                    -
                    \E[\ba_t \mid \bs_t]
                \right)^2
            \mid
                \bs_t
        \right]
    \right]
\notag\\
&=
    \E_{d}
    \left[ 
        \var(\ba_t \mid \bs_t) 
    \right]
.
\end{align}
The cross-term vanishes
(recall that 
$
    \opredbc(s_t) 
\defeq 
    \E_{\opol}
    \left[
        \ba_t \mid s_t
    \right]
$):
\begin{align}
    \E_{d, \opol, \datadist_n}
    \left[ 
            \left(
                \ba_t - \opredbc(\bs_t) 
            \right)
            \left(
                \opredbc(\bs_t)  - \predmun(\bs_t)
            \right)
        \mid
            \bs_t
    \right]
&=
    \E_{d, \datadist_n}
    \left[ 
        \E_{\opol}
        \left[
                \ba_t - \opredbc(\bs_t) 
            \mid
                \bs_t
        \right]
        \left(
            \opredbc(\bs_t)  - \predmun(\bs_t)
        \right)
    \right]
\notag\\
&=
    \E_{d, \datadist_n}
    \left[ 
        \left(
            \E[\ba_t \mid \bs_t]
             - \opredbc(\bs_t) 
        \right)
        \left(
            \opredbc(\bs_t)  - \predmun(\bs_t)
        \right)
    \right]
\notag\\
&=
    0
.
\end{align}
The second term decomposes in the expected variance and expected bias terms:
\begin{align}
    \E_{d, \datadist_n}
    &
    \left[
        \left(
            \opredbc(\bs_t)  - \predmun(\bs_t)
        \right)^2 
    \right]
\notag\\
&=
    \E_{d}
    \left[
        \E_{\datadist_n}
        \left[
            \Big(
                \opredbc(\bs_t)  - \predmun(\bs_t)
                + \E_{\datadist_n}\left[ \predmun(\bs_t) \right]
                - \E_{\datadist_n}\left[ \predmun(\bs_t) \right]
            \Big)^2 
        \right]
    \right]
\notag\\
&=
    \E_{d}
    \left[
        \E_{\datadist_n}
        \left[
            \Big(
                \Big(
                \E_{\datadist_n}\left[ \predmun(\bs_t) \right]
                - 
                \predmun(\bs_t)
                \Big)
                +
                \Big(
                \opredbc(\bs_t) 
                - 
                \E_{\datadist_n}\left[ \predmun(\bs_t) \right]
                \Big)
            \Big)^2 
        \right]
    \right]
\notag\\
&=
    \E_{d}
    \left[
        \E_{\datadist_n}
        \left[
            \Big(
                \E_{\datadist_n}\left[ \predmun(\bs_t) \right]
                - 
                \predmun(\bs_t)
            \Big)^2 
        \right]
    \right]
    +
    \E_{d}
    \left[
        \E_{\datadist_n}
        \left[
            \Big(
                \opredbc(\bs_t) 
                - 
                \E_{\datadist_n}\left[ \predmun(\bs_t) \right]
            \Big)^2 
        \right]
    \right]
\notag\\
&\quad
    +
    2
    \E_{d}
    \Big[
        \E_{\datadist_n}
        \Big[
            \Big(
                \E_{\datadist_n}\left[ \predmun(\bs_t) \right]
                - 
                \predmun(\bs_t)
            \Big)
            \Big(
                \opredbc(\bs_t) 
                - 
                \E_{\datadist_n}\left[ \predmun(\bs_t) \right]
            \Big)
        \Big]
    \Big]
\notag\\
&=
    \E_{d}
    \left[
        \var_{\datadist_n}
        \left(
            \predmun
            \left(
                \bs_t
            \right)
        \right)
    \right]
    +
    \E_{d}
    \left[
        \Big(
            \opredbc(\bs_t) 
            - 
            \E_{\datadist_n}\left[ \predmun(\bs_t) \right]
        \Big)^2 
    \right]
,
\end{align}
where the cross-term also vanished since its rightmost term is independent of the distribution over datasets:
\begin{align}
    \E_{\datadist_n}
&
    \Big[
        \Big(
            \E_{\datadist_n}\left[ \predmun(\bs_t) \right]
            - 
            \predmun(\bs_t)
        \Big)
        \Big(
            \opredbc(\bs_t) 
            - 
            \E_{\datadist_n}\left[ \predmun(\bs_t) \right]
        \Big)
    \Big]
\notag\\
&
=
    \E_{\datadist_n}
    \Big[
        \Big(
            \E_{\datadist_n}\left[ \predmun(\bs_t) \right]
            - 
            \predmun(\bs_t)
        \Big)
    \Big]    
    \Big(
        \opredbc(\bs_t) 
        - 
        \E_{\datadist_n}\left[ \predmun(\bs_t) \right]
    \Big)
\notag\\
&
=
    \Big(
        \E_{\datadist_n}\left[ \predmun(\bs_t) \right]
            - 
        \E_{\datadist_n}
        \left[
            \predmun(\bs_t)
        \right]    
    \Big)
    \Big(
        \opredbc(\bs_t) 
        - 
        \E_{\datadist_n}\left[ \predmun(\bs_t) \right]
    \Big)
\notag\\
&
=
    0
\end{align}

Putting the terms together, we have:
\begin{align}
    \epe(\predmun) 
&=
    \E_{d}
    \left[ 
        \var(\ba_t \mid \bs_t) 
    \right]
    +
    \E_{d}
    \left[
        \var_{\datadist_n}
        \left(
            \predmun
            \left(
                \bs_t
            \right)
        \right)
    \right]
    +
    \E_{d}
    \left[
        \left(
            \opredbc(\bs_t)
            -
            \E_{\datadist_n}\left[ \predmun(\bs_t) \right]
        \right)^2
    \right]    
.
\label{eq:mse-hatmu}
\end{align}

Now, we expand the expected-predicted-error for the IDM estimator, which is given by:
\begin{align}
    \epe
    \left(
        \predxim; \approxplanner 
    \right) 
&
=
    \E_{
        d, 
        \approxplanner,
        \opolidm,
        \approxdatadist
    }
    \left[ 
            \left( 
                \ba_t
                -
                \predxim(\bs_t, \bs_{t+k})
            \right)^2 
        \mid
            \bs_t, \bs_{t+k}
    \right]
,
\end{align}
where the covariate shift arises between the future-state distribution $\oplanner$ used to generate the training dataset $\approxdatadist$ and the approximate future-state distribution $\approxplanner$ used at test-time to evaluate the predicted error.
Applying the same derivation to $\predxim$ as for $\predmun$ yields:
\begin{align}
    \epe
    \left(
        \predxim; \approxplanner
    \right) 
&=
    \E_{d, \approxplanner}
    \left[
        \var(\ba_t \mid \bs_t, \bs_{t+k})
    \right] 
    +
    \E_{d, \approxplanner}
    \left[
        \var_{\approxdatadist}
        \left(
            \predxim
            \left(
                \bs_t, \bs_{t+k}
            \right)
        \right)
    \right]
\notag\\
&\quad
    +
    \E_{d, \approxplanner}
    \left[
        \left(
            \opredidm(\bs_t, \bs_{t+k})
            -
            \E_{\approxdatadist}
            \left[ 
                \predxim(\bs_t, \bs_{t+k}) 
            \right]
        \right)^2
    \right]
.
\label{eq:mse-hatxi}
\end{align}
To compare \eqref{eq:mse-hatxi} with \eqref{eq:mse-hatmu} and make the effect of the covariate shift explicit, we use the density ratio,
which is well-defined under \Cref{ass:support}:
$
w(s_t,s_{t+k}) < \infty
$ for all $s_t \in \St$ and all $s_{t+k} \in \supp(\approxplanner)$.
Hence, we can change the distribution in the expected irreducible noise:
\begin{align}
    \E_{d, \approxplanner}
    \left[
        \var(\ba_t \mid \bs_t, \bs_{t+k})
    \right] 
=&
    \E_{d, \oplanner}
    \left[
        w(\bs_t,\bs_{t+k})
        \var(\ba_t \mid \bs_t, \bs_{t+k})
    \right] 
.
\end{align}
Doing the same change of expectation for the bias and variance terms in \eqref{eq:mse-hatxi}, we have:
\begin{align}
    \epe
    \left(
        \predxim; \approxplanner
    \right) 
=&
    \E_{d, \oplanner}
    \left[
        w(\bs_t,\bs_{t+k})
        \var(\ba_t \mid \bs_t, \bs_{t+k})
    \right] 
\notag\\
&
    +
    \E_{d, \oplanner}
    \left[
        w(\bs_t,\bs_{t+k})
        \var_{\approxdatadist}
        \left(
            \predxim
            \left(
                \bs_t, \bs_{t+k}
            \right)
        \right)
    \right]
\notag\\
&
    +
    \E_{d, \oplanner}
    \left[
        w(\bs_t,\bs_{t+k})
        \left(
            \opredidm(\bs_t, \bs_{t+k})
            -
            \E_{\approxdatadist}
            \left[ 
                \predxim(\bs_t, \bs_{t+k}) 
            \right]
        \right)^2
    \right]
.
\label{eq:mse-hatxi-importance-weighted}
\end{align}

Using the definition of $\gamma$ from \eqref{eq:covariate-shift-bis} and subtracting \eqref{eq:mse-hatxi-importance-weighted} from \eqref{eq:mse-hatmu} and grouping terms concludes the proof.
\end{IEEEproof}

\subsection{Proof of \Cref{theorem:sample-efficiency-gain}}
\label{app:proof-theorem-sample-efficiency-gain}
We start by recalling the notation needed for the subsequent proof of \Cref{theorem:sample-efficiency-gain}. 
We first define the bias of the BC policy parameter, and its derivative evaluated at $\omu$:
\begin{align}
    \biasmu
&
\defeq
    \E
    \left[
        \mun
    \right]
    -
    \mu
,
\\
    \dbiasmu
&
\defeq
    \frac{
        \partial 
    }{
        \partial \mu
    }
    \biasmu
\biggr\rvert_{\mu=\omu}
.
\end{align}
where the expectation is taken over datasets generated by some policy $\pol_\mu$
(note that for the expert policy, $\opol$, we made the distribution over datasets explicit with $\E_{\datadist_n}[\cdot]$).

Next, we analogously define these terms for the IDM policy parameter, with its derivative evaluated at $\oxi$:
\begin{align}
    \biasxi
&=
    \E
    \left[
        \xim
    \right]
    -
    \xi
,
\\
    \dbbiasapproxoxi
&\defeq
    \frac{
        \partial 
    }{
        \partial \xi
    }
    \biasxi
\biggr\rvert_{\xi=\oxi}
.
\end{align}

Lastly, under correct model specification, the Fisher information evaluated at the optimal parameter measures how much information the available data carry about the true parameter:
\begin{align}
    \fishermu
&\defeq
    \E_{\substack{
        \bs_t \sim d(\cdot),\\
        \ba_t \sim \pol_{\omu}(\cdot\mid \bs_t)
    }}
    \left[
        \left(
            \dmu
            \ln \pi_{\mu} (\ba_t \mid \bs_t)
            \biggr\rvert_{\mu=\omu}
        \right)^2
    \right]
,
\\
\fisherxi
&\defeq
    \E_{\substack{
            \bs_t \sim d(\cdot),\\
            \bs_{t+k} \sim \oplanner(\cdot \mid \bs_t),\\
            \ba_t \sim \polapproxoxi(\cdot \mid \bs_t, \bs_{t+k})
    }}
    \left[
        \left(
            \dxi
            \ln \idmpol (\ba_t \mid \bs_t, \bs_{t+k})
            \biggr\rvert_{\xi=\oxi}
        \right)^2
    \right].
\end{align}

\begin{theorem}
Let Assumptions~\ref{ass:iid}--\ref{ass:policy_class_expressivity} hold.
Let $\mun$ and $\xim$ denote the estimated parameters for the BC and IDM predictors learned from $\datadist_n$ and $\approxdatadist$, respectively, where $n$ and $m$ denote the number of training samples required to achieve a target MSE level $\varepsilon$.
Assume that both estimators asymptotically attain the biased Cramér-Rao lower bound.
Then, for sufficiently large $n$ and $m$, we have:
\begin{align}
    \eta
&\defeq
    \frac{n}{m}
\approx
    \frac{\fisherxi}{\fishermu}
        \frac{
            \left(
                \varepsilon 
                - 
                \biasxisq
            \right)
            \left(
                1
                +
                \dbbiasmu
            \right)^2
        }{
            \left(
                \varepsilon 
                - 
                \biasmusq
            \right)            
            \left(
                1
                +
                \dbiasxi
            \right)^2
        }
.
\label{eq:sample-efficiency-gain-bis}
\end{align}
\end{theorem}

\begin{IEEEproof}
Define the MSE for the BC and IDM parameters:
\begin{align}
    \mse(\mun)
&\defeq
    \E_{\datadist_n}
    \left[
        \left(
            \mun - \omu
        \right)^2
    \right]
,
\\
    \mse(\xim)
&\defeq
    \E_{\approxdatadist}
    \left[
        \left(
            \xim - \oxi
        \right)^2
    \right]
.
\end{align}

The Cramér-Rao Lower Bound (CRLB) gives a lower bound on how much the statistic $\mun$ can fluctuate around the true parameter~\citep[Eq. (11.290)]{cover2005elements}:
\begin{equation}
    \var_{\datadist_n}(\mun)
\ge
     \frac{
        \left(
            1
            +
            \dbiasmu
        \right)^2
     }{
        n\fishermu
     }
.
\label{eq:general-cramer-rao}
\end{equation}
Since the theorem assumes the estimator asymptotically attains the lower bound,
we can approximate \eqref{eq:general-cramer-rao} with equality for large enough number of samples.
Hence, using the standard bias-variance decomposition of the MSE, we have:
\begin{align}
    \mse(\mun)
&=
    \var_{\datadist_n}(\mun)
    +
    \biasmusq
\approx
    \frac{
        \left(
            1
            +
            \dbbiasmu
        \right)^2
    }{n\fishermu}
    +
    \biasmusq
.
\end{align}
Since $\mse\left(\mun\right) = \varepsilon$, we can solve for $n$:
\begin{equation}
    n
\approx
    \frac{
        \left(
            1
            +
            \dbbiasmu
        \right)^2
    }{
        \left(
            \varepsilon 
            - 
            \biasmusq
        \right)
        \fishermu
    }
.
\end{equation}

Now, we follow the same reasoning for $\mse\left(\xim \right)$:
\begin{align}
    \mse(\xim)
&=
    \var_{\approxdatadist}(\xim)
    +
    \biasxisq
\approx
    \frac{
        \left(
            1
            +
            \dbiasxi
        \right)^2
    }{m \fisherxi}
    +
    \biasxisq
.
\end{align}
Similarly, since $\mse\left(\xim\right) = \varepsilon$, 
we can solve for $m$: 
\begin{equation}
    m
\approx
    \frac{
        \left(
            1
            +
            \dbiasxi
        \right)^2
    }{
        \left(
            \varepsilon 
            - 
            \biasxisq
        \right)
        \fisherxi
    }
.
\end{equation}

Using the definition of the sample efficiency ratio, we have:
\begin{align}
    \eta
&\defeq
    \frac{n}{m}
\approx
    \frac{\fisherxi}{\fishermu}
        \frac{
            \left(
                \varepsilon 
                - 
                \biasxisq
            \right)
            \left(
                1
                +
                \dbbiasmu
            \right)^2
        }{
            \left(
                \varepsilon 
                - 
                \biasmusq
            \right)            
            \left(
                1
                +
                \dbiasxi
            \right)^2
        }
.
\label{eq:unsimplified-ratio}
\end{align}
\end{IEEEproof}

\subsection{Proof of \Cref{lemma:fisher-ratio-greater-than-one}}
\label{app:lemma:fisher-ratio-greater-than-one}

\begin{lemma}
\label{lemma:fisher-ratio-greater-than-one-bis}
Under Assumptions \ref{ass:policy_class_expressivity}--\ref{ass:regularity-fisher}:
$
    \fisherxi
\ge 
    \fishermu
$.
\end{lemma}
\begin{IEEEproof}
By \Cref{ass:local-shared-param}, in a neighborhood of $\oxi$ the BC policy is the marginal of the IDM policy under $\oplanner$, with parameter map $\Psi$ and $\Psi(\oxi) = \omu$:
\begin{equation}
\label{eq:marginal-common-param-approx}
    \pi_{\Psi(\xi)}(a_t \mid s_t)
=
    \int_{\St}
        \oplanner(s_{t+k} \mid s_t)\,
        \idmpol(a_t \mid s_t, s_{t+k})\,
        ds_{t+k}
.
\end{equation}
Recall the Fisher information for the BC and IDM predictors are given by:
\begin{align}
    \fishermu
&\defeq
    \E_{\substack{
        \bs_t \sim d(\cdot),\\
        \ba_t \sim \pol_{\omu}(\cdot\mid \bs_t)
    }}
    \left[
        \left(
            \dmu
            \ln \pi_{\mu} (\ba_t \mid \bs_t)
            \biggr\rvert_{\mu=\omu}
        \right)^2
    \right]
    \label{eq:lemma1_fisher_bc}
\\
\fisherxi
&\defeq
    \E_{\substack{
            \bs_t \sim d(\cdot),\\
            \bs_{t+k} \sim \oplanner(\cdot \mid \bs_t),\\
            \ba_t \sim \polapproxoxi(\cdot \mid \bs_t, \bs_{t+k})
    }}
    \left[
        \left(
            \dxi
            \ln \idmpol (\ba_t \mid \bs_t, \bs_{t+k})
            \biggr\rvert_{\xi=\oxi}
        \right)^2
    \right].
    \label{eq:lemma1_fisher_idm}
\end{align}
By the chain rule, since $\pi_{\Psi(\xi)}$ depends on $\xi$ only through $\Psi$ and $\Psi(\oxi)=\omu$, we have
\begin{equation}
    \dxi \ln \pi_{\Psi(\xi)}(\ba_t\mid\bs_t)\biggr\rvert_{\xi=\oxi}
=
    \Psi'(\oxi)\,
    \dmu \ln \pi_\mu(\ba_t\mid\bs_t)\biggr\rvert_{\mu=\omu}
,
\end{equation}
such that:
\begin{equation}
    \E_{\substack{
        \bs_t \sim d(\cdot),\\
        \ba_t \sim \pol_{\omu}(\cdot\mid \bs_t)
    }}
    \left[
        \left(
            \dxi
            \ln \pol_{\Psi(\xi)} (\ba_t \mid \bs_t)
            \biggr\rvert_{\xi=\oxi}
        \right)^2
    \right]
=
    \big(\Psi'(\oxi)\big)^2 \fishermu
,
\end{equation}
from which follows:
\begin{equation}
    \fishermu
=
    \frac{1}{\big(\Psi'(\oxi)\big)^2} \E_{\substack{
        \bs_t \sim d(\cdot),\\
        \ba_t \sim \pol_{\omu}(\cdot\mid \bs_t)
    }}
    \left[
        \left(
            \dxi
            \ln \pol_{\Psi(\xi)} (\ba_t \mid \bs_t)
            \biggr\rvert_{\xi=\oxi}
        \right)^2
    \right]
.
\label{eq:bc-score-identification}
\end{equation}

Since $\oplanner$ is independent of $\xi$, we expand the score of the marginal policy, which holds for all $\xi \in U$ (where $U$ is the neighborhood defined by \Cref{ass:local-shared-param}):
\begin{align}
    \dxi
    \ln \pi_{\Psi(\xi)} (a_t \mid s_t)
&=
    \frac{
        \dxi
        \pi_{\Psi(\xi)} (a_t \mid s_t)
    }{
        \pi_{\Psi(\xi)} (a_t \mid s_t)
    }
\notag\\
&=
    \frac{
        \dxi \int_{\St}
            \oplanner(s_{t+k} \mid s_t)
            \idmpol(a_t \mid s_t, s_{t+k})
            ds_{t+k}
    }{
        \pi_{\Psi(\xi)} (a_t \mid s_t)
    }
\notag\\
&=
    \frac{
        \int_{\St}
            \oplanner(s_{t+k} \mid s_t)
            \left(
                \dxi
                \idmpol(a_t \mid s_t, s_{t+k})
            \right)
            ds_{t+k}
    }{
        \pi_{\Psi(\xi)} (a_t \mid s_t)
    }
\notag\\
&=
    \frac{
        \int_{\St}
            \oplanner(s_{t+k} \mid s_t)
            \idmpol (a_t \mid s_t, s_{t+k})
            \left(
                \dxi
                \ln \idmpol (a_t \mid s_t, s_{t+k})
            \right)
            ds_{t+k}
    }{
        \pi_{\Psi(\xi)} (a_t \mid s_t)
    }
\notag\\
&=
    \int_{\St}
        P_{\xi}(s_{t+k} \mid s_t, a_t)
        \left(
            \dxi
            \ln \idmpol (a_t \mid s_t, s_{t+k})
        \right)
        ds_{t+k}
\notag\\
&=
    \E_{s_{t+k} \sim P_{\xi}(\cdot \mid s_t, a_t)}
    \left[
        \dxi
        \ln \idmpol (a_t \mid s_t, s_{t+k})
    \right]
,
\end{align}
where we interchanged derivative and integral (\Cref{ass:regularity-fisher}), used the identity $\dxi \idmpol = \idmpol\, \dxi \ln \idmpol$, and defined the Bayes posterior
\begin{equation}
    P_{\xi}(s_{t+k} \mid s_t, a_t)
\defeq
    \frac{
        \oplanner(s_{t+k} \mid s_t)\,
        \pi_\xi(a_t \mid s_t, s_{t+k})
    }{
        \pi_{\Psi(\xi)}(a_t \mid s_t)
    }
.\label{eq:phat_lemma1_def}
\end{equation}
By Jensen's inequality applied to the square:
\begin{align}
\label{eq:jensen-approx}
    \left(
        \dxi
        \ln \pi_{\Psi(\xi)} (a_t \mid s_t)
    \right)^2
&=
    \left(
        \E_{\bs_{t+k} \sim P_{\xi}(\cdot \mid s_t, a_t)}
        \left[
            \dxi
            \ln \idmpol (a_t \mid s_t, \bs_{t+k})
        \right]
    \right)^2
\notag\\
&\le
    \E_{\bs_{t+k} \sim P_{\xi}(\cdot \mid s_t, a_t)}
    \left[
        \left(
            \dxi
            \ln \idmpol (a_t \mid s_t, \bs_{t+k})
        \right)^2
    \right]
.
\end{align}
The identities above hold for all $\xi$ in the neighborhood; we now instantiate them at $\xi = \oxi$, where $\pi_{\Psi(\oxi)} = \pol_{\omu}$ by $\Psi(\oxi)=\omu$, so that $P_{\oxi}$ from \eqref{eq:phat_lemma1_def} has now denominator $\pol_{\omu}(a_t \mid s_t)$. Since \eqref{eq:jensen-approx} holds pointwise for all $s_t \in \St$ and all $a_t \in \Ac$, taking expectations preserves the inequality. Using the identification \eqref{eq:bc-score-identification} to express $\fishermu$ via the $\xi$-derivative, we obtain:
\begin{align}
    \fishermu
&\overset{\eqref{eq:bc-score-identification}}{=}
    \frac{1}{\big(\Psi'(\oxi)\big)^2} \E_{\substack{
        \bs_t \sim d(\cdot),\\
        \ba_t \sim \pol_{\omu}(\cdot\mid \bs_t)
    }}
    \left[
        \left(
            \dxi
            \ln \pol_{\Psi(\xi)} (\ba_t \mid \bs_t)
            \biggr\rvert_{\xi=\oxi}
        \right)^2
    \right]
\notag\\
&=
    \frac{1}{\big(\Psi'(\oxi)\big)^2}
    \int_{\St \times \Ac}
        d(s_t)\,
        \pol_{\omu}(a_t \mid s_t)
        \left(
            \dxi
            \ln \pi_{\Psi(\xi)} (a_t \mid s_t)
            \big\rvert_{\xi=\oxi}
        \right)^2
        da_t\, ds_t
\notag\\
&\overset{\eqref{eq:jensen-approx}}{\le}
    \frac{1}{\big(\Psi'(\oxi)\big)^2}
    \int_{\St \times \Ac \times \St}
        d(s_t)\,
        \pol_{\omu}(a_t \mid s_t)\,
        P_{\oxi}(s_{t+k} \mid s_t, a_t)
        \left(
            \dxi
            \ln \idmpol (a_t \mid s_t, s_{t+k})
            \big\rvert_{\xi=\oxi}
        \right)^2
        ds_{t+k}\, da_t\, ds_t
\notag\\
&\overset{\eqref{eq:phat_lemma1_def}}{=}
    \frac{1}{\big(\Psi'(\oxi)\big)^2}
    \int_{\St \times \Ac \times \St}
        d(s_t)\,
        \oplanner(s_{t+k} \mid s_t)\,
        \pi_{\oxi}(a_t \mid s_t, s_{t+k})
        \left(
            \dxi
            \ln \idmpol (a_t \mid s_t, s_{t+k})
            \big\rvert_{\xi=\oxi}
        \right)^2
        ds_{t+k}\, da_t\, ds_t
\notag\\
&\overset{\eqref{eq:lemma1_fisher_idm}}{=}
    \frac{1}{\big(\Psi'(\oxi)\big)^2}
    \fisherxi
\notag\\
&= \fisherxi
,
\label{eq:fisher-inequality-expanded-approx}
\end{align}
where the last equality follows from $\Psi'(\oxi) = 1$ (\Cref{ass:local-shared-param}).
\end{IEEEproof}

\subsection{Proof of \Cref{corollary:sample-efficient-similar-bias}}
\begin{corollary}
Under the conditions of \Cref{theorem:sample-efficiency-gain} and
Assumptions 
\ref{ass:local-shared-param}--\ref{ass:regularity-fisher}, 
if
$\biasmusq = \biasxisq$ and $\dbiasmu=\dbiasxi$, then:
$
\eta \gtrsim 1
$.
\end{corollary}
\begin{IEEEproof}
Under similar bias and bias derivative for $\mun$ and $\xim$, \Cref{eq:sample-efficiency-gain} simplifies to:    
\begin{equation}
    \eta
\approx
    \frac{\fisherxi}{\fishermu}
.
\end{equation}
Using \Cref{lemma:fisher-ratio-greater-than-one} concludes the proof.
\end{IEEEproof}

\section{Extensions}
\label{app:extensions}

\subsection{Cross Entropy Loss and Other Losses}
\label{app:cross-entropy-loss}
Redefine
$
    \predomu(a_t \mid s_t)
$
and
$
    \predoxi(a_t \mid s_t,s_{t+k})
$
so they denote the Bayes-optimal predictors under the cross-entropy loss.
Define the corresponding predicted error gap: 
\begin{equation}
\Delta_{\rm{CE}} \defeq \epe(\predomu) - \epe(\predoxi)
\end{equation}
We have a result analogous to \Cref{theorem:mse-gap} but for the cross-entropy loss. 
\begin{theorem}
$
    \Delta_{\rm{CE}}
= 
    \E_{d, \oplanner} 
    \left[
        \kl
        \left(
            \pi_\xi(\ba_t \mid \bs_t, \bs_{t+k} )
            \|
            \pi_\mu (\ba_t \mid \bs_t)
        \right)
    \right]
\ge 
    0
$.
\end{theorem}
\begin{IEEEproof}
Note the minimal risk is the conditional entropy: 
\begin{align}
    \epe(\predomu) 
&=
    H(\ba_t \mid \bs_t)
\\    
    \epe(\predoxi)
&=
    H(\ba_t \mid \bs_t, \bs_{t+k})
.
\end{align} 
Hence, the gap under the cross-entropy loss becomes the conditional mutual information between $\ba_t$ and $\bs_{t+k}$ given $\bs_t$:
\begin{align}
    \Delta_{\rm{CE}}
&=
    H(\ba_t \mid \bs_t ) 
    - 
    H(\ba_t \mid \bs_t, \bs_{t+k} )
\notag\\
&=
    H(\ba_t \mid \bs_t )
    +
    H(\bs_{t+k} \mid \bs_t )
    -
    \left(
        H(\ba_t \mid \bs_t, \bs_{t+k} )
        +
        H(\bs_{t+k} \mid \bs_t)
    \right)
\notag\\
&=
    H(\ba_t \mid \bs_t )
    +
    H(\bs_{t+k} \mid \bs_t )
    -
    H(\ba_t, \bs_{t+k} \mid \bs_t )
\notag\\
&=
    I(\ba_t; \bs_{t+k} \mid \bs_t )
\notag\\
&=
    \E_{d, \oplanner} 
    \left[
        \kl
        \left(
            \pi_\xi(\ba_t \mid \bs_t, \bs_{t+k} )
            \|
            \pi_\mu (\ba_t \mid \bs_t)
        \right)
    \right]
\notag\\
&\ge
    0
.    
\end{align}
\end{IEEEproof}

Similarly, we can extend Corollary 1 by using a bias-variance decomposition of the predicted error in terms of the KL divergence (see, e.g., \citep{heskes2025bias}).
Since the fundamental uncertainty reduction is similar, we hypothesize the insights on sample efficiency will also transfer to discrete action spaces.

More generally, since the bias-variance decomposition has been extended to Bregman divergence losses \citep{adlam2022understanding,heskes2025bias,pfau2025generalized}, we hypothesize the same mechanism holds for distributional policies more generally, consistent with empirical evidence showing performance gains of PIDM over BC for transformers \citep{tian2025predictive} and diffusion models \citep{xie2025latent}.

\subsection{Discussion on Non-i.i.d. Sampling}
\label{app:non-iid}

The uncertainty reduction result in \Cref{theorem:mse-gap} extends directly to samples drawn from a stationary ergodic Markov process. 
It is a population identity for the stationary triple $\left( s_t,a_t,s_{t+k} \right)$ and its proof relies only on conditional expectation and the law of total variance. Therefore, it does not require temporal independence.

\Cref{corollary:gap-general-estimator} requires only the estimator distribution to be well defined.
Its proof uses the standard bias-variance decomposition of the MSE, which does not rely on temporal independence of the data.
Temporal dependence may change the magnitude of the variance terms, but it does not affect the validity of the bias–variance decomposition itself.

For the sample-efficiency analysis, a fully rigorous extension to Markov trajectories requires replacing the i.i.d. information term $nF$ with the Fisher information of the full trajectory, defined through the the joint likelihood of the full trajectory under the Markov factorization. 
However, under standard ergodicity and mixing assumptions, the Fisher information of the full trajectory still increases approximately linearly with trajectory length for sufficiently long trajectories \citep{radaelli2023fisher}. 
Moreover, since both $F_\mu$ and $\fisherxi$ are multiplied by the same leading-order factor, their ratio is expected to remain approximately unchanged.

In summary, the uncertainty-reduction effect persists under correlated samples. While temporal dependence alters the effective sample sizes of both estimators, it does so in a comparable manner, so the relative efficiency advantage of PIDM over BC is expected to persist.

\section{Additional Results for 2D Navigation Environment}
\label{app:toy_additional_analysis}

\subsection{Expected Prediction Error Gaps for all Tasks}

\begin{figure*}[h]
    \centering
    \includegraphics[width=\linewidth]{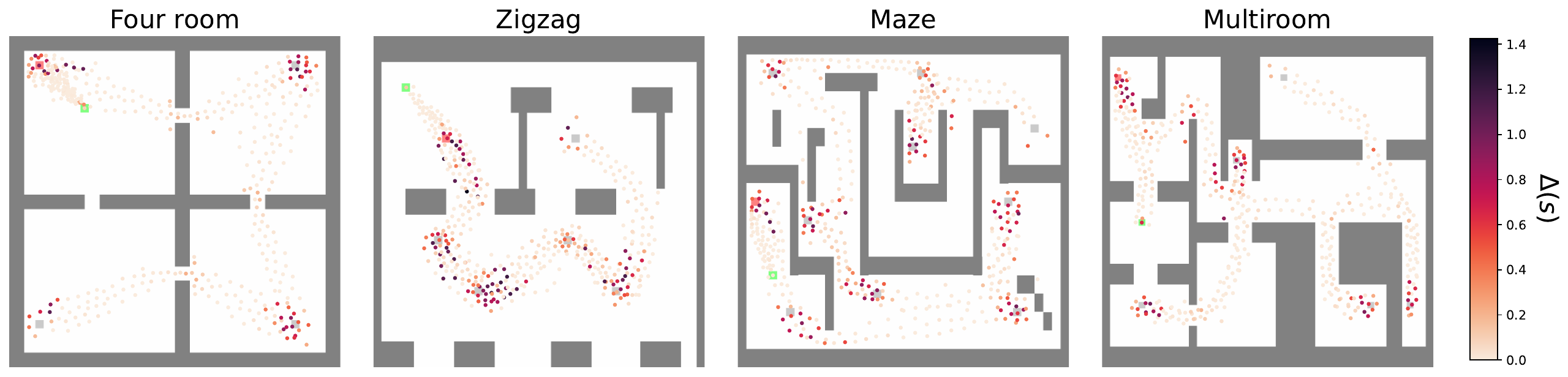}
    \caption{Visualized state-wise EPE gaps $\Delta(s)$ from \eqref{eq:delta_per_state} computed for each dataset. We observe large gaps in states surrounding the goals where human actions are more diverse.}
    \label{fig:toy_datasets_variance_delta_per_state}
\end{figure*}

\subsection{IDM Policy Visualizations in all Tasks}

\begin{figure*}[h]
    \begin{subfigure}{0.225\linewidth}
        \centering
        \includegraphics[width=\linewidth]{images/toy_env/idm_policy_visualization/idm_policy_vector_field_four_room_stochastic_transitions_std_0_2_50_samples_seed_0_idm.yaml_kmeans_50.pdf}
        \caption{Four room}
        \label{fig:four_room_idm_policy_visualization}
    \end{subfigure}
    \begin{subfigure}{0.225\linewidth}
        \centering
        \includegraphics[width=\linewidth]{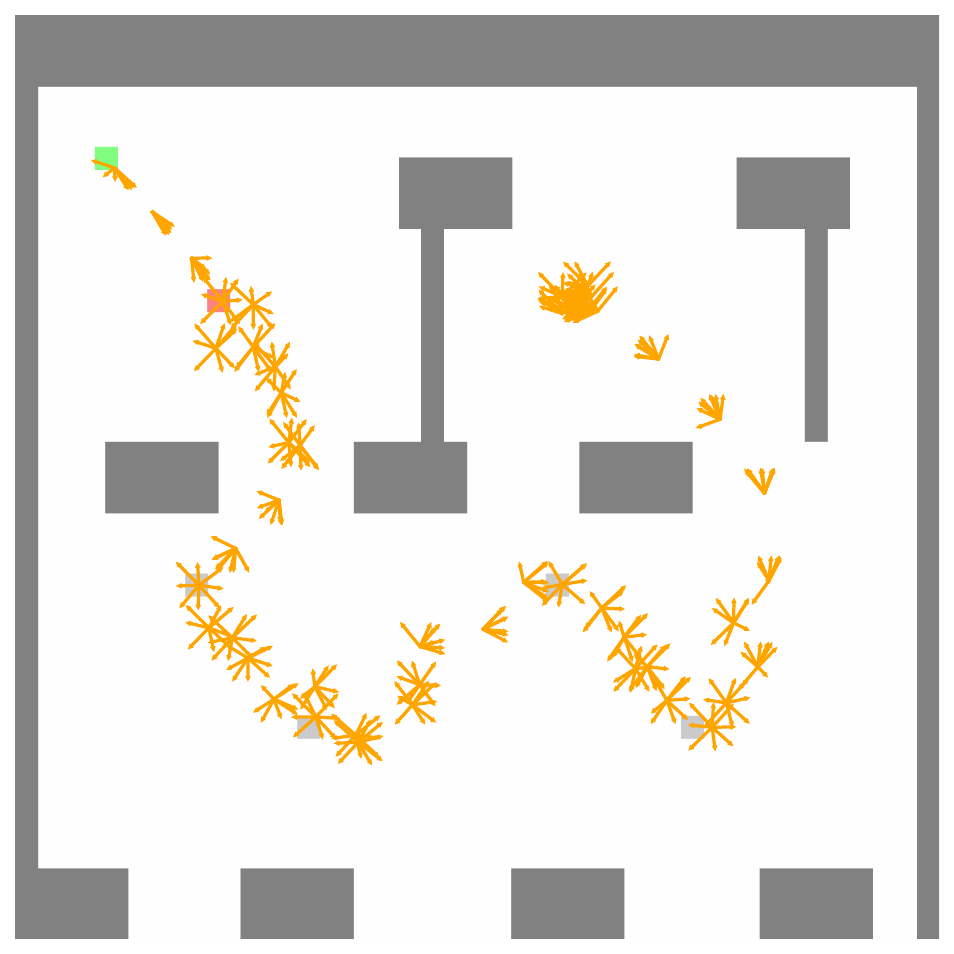}
        \caption{Zigzag}
        \label{fig:zigzag_idm_policy_visualization}
    \end{subfigure}
    \begin{subfigure}{0.225\linewidth}
        \centering
        \includegraphics[width=\linewidth]{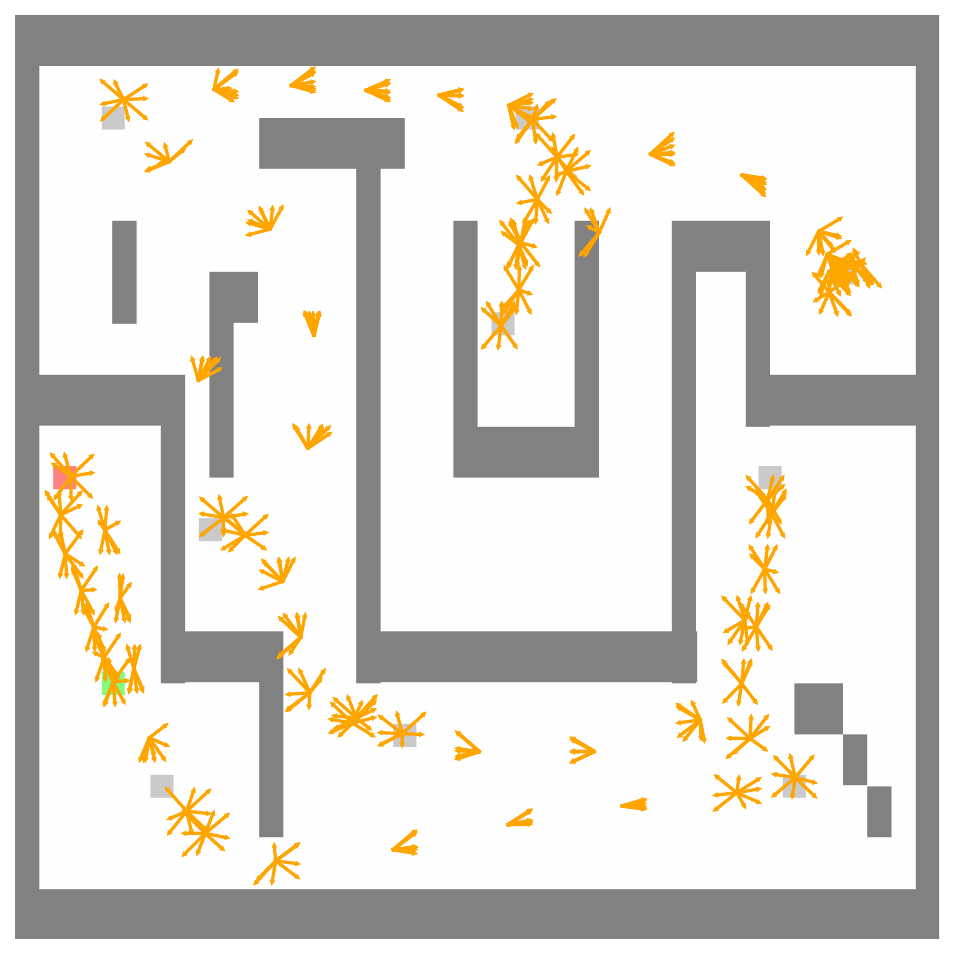}
        \caption{Maze}
        \label{fig:maze_idm_policy_visualization}
    \end{subfigure}
    \begin{subfigure}{0.225\linewidth}
        \centering
        \includegraphics[width=\linewidth]{images/toy_env/idm_policy_visualization/idm_policy_vector_field_multiroom_stochastic_transitions_std_0_2_50_samples_seed_0_idm.yaml_kmeans_75.pdf}
        \caption{Multiroom}
        \label{fig:multiroom_idm_policy_visualization}
    \end{subfigure}
    \hfill
    \caption{Visualization of IDM policies when queried for representative states and possible future states in each cardinal and diagonal direction for all four tasks. Predicted actions spread out in states where the dataset exhibits large $\Delta(s)$.}
    \label{fig:idm_policy_visualization}
\end{figure*}

\subsection{Evaluation Results for Best vs Last Checkpoint}
\label{app:best_last_checkpoint}
\begin{figure}[t]
    \centering
    \begin{subfigure}{.245\linewidth}
        \centering
        \includegraphics[width=\linewidth]{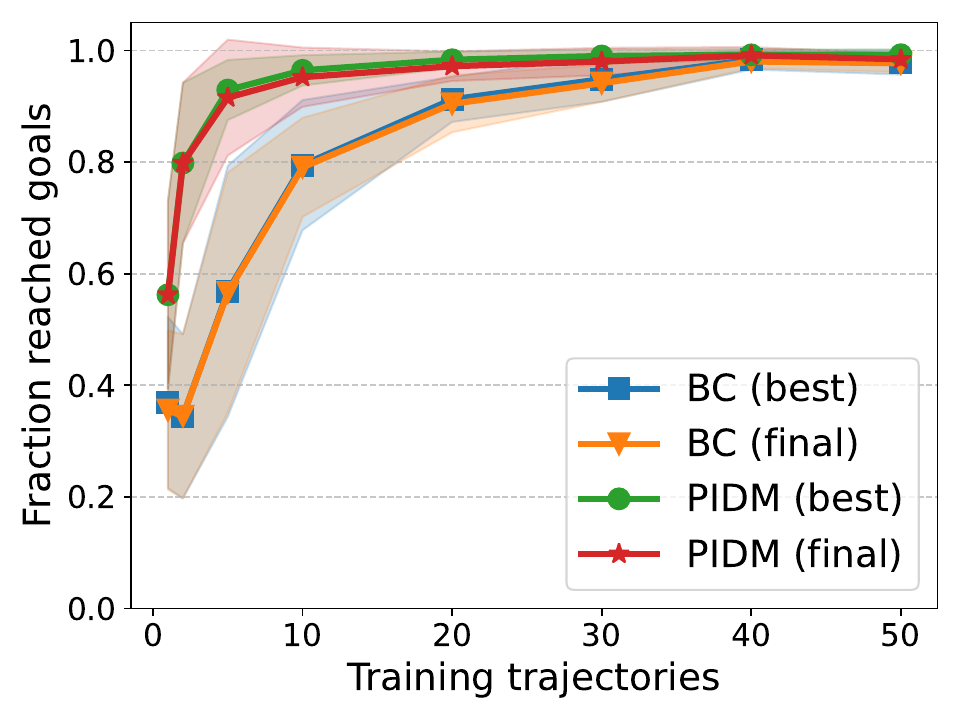}
        \caption{Four room}
        \label{fig:toy_four_room_pidm_best_last_checkpoint}
    \end{subfigure}
    \hfill
    \begin{subfigure}{.245\linewidth}
        \centering
        \includegraphics[width=\linewidth]{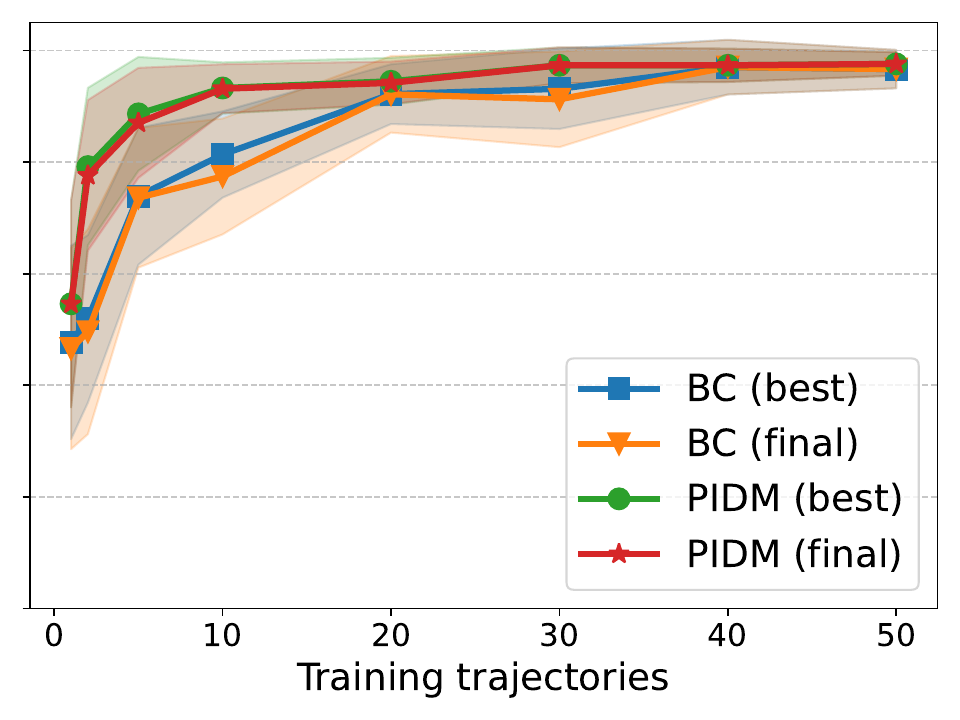}
        \caption{Zigzag}
        \label{fig:toy_zigzag_pidm_best_last_checkpoint}
    \end{subfigure}
    \hfill
    \begin{subfigure}{.245\linewidth}
        \centering
        \includegraphics[width=\linewidth]{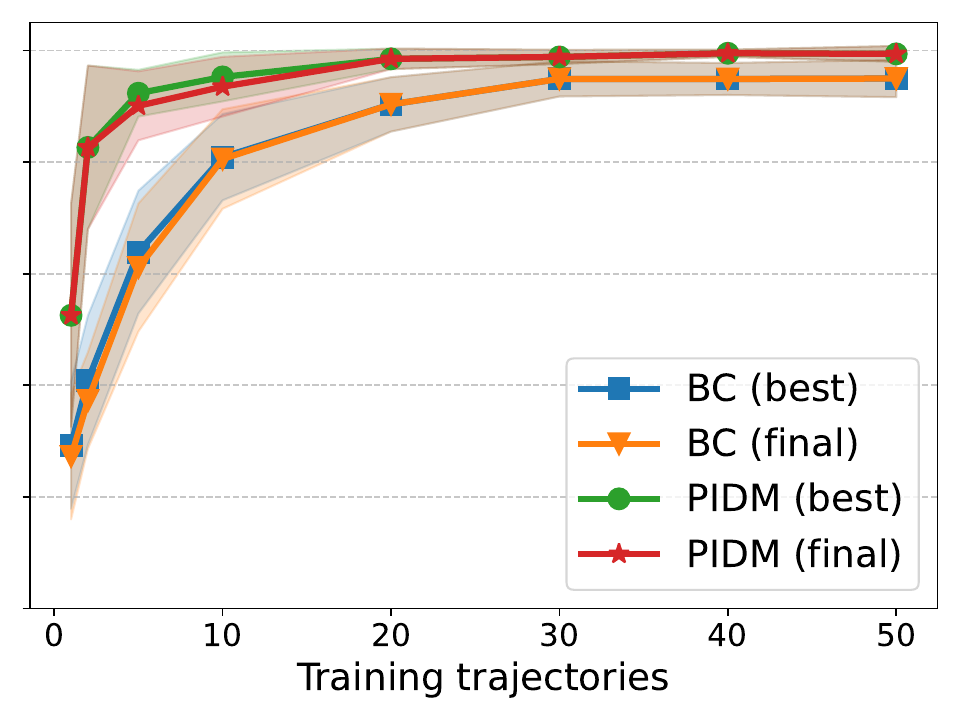}
        \caption{Maze}
        \label{fig:toy_maze_pidm_best_last_checkpoint}
    \end{subfigure}
    \hfill
    \begin{subfigure}{.245\linewidth}
        \centering
        \includegraphics[width=\linewidth]{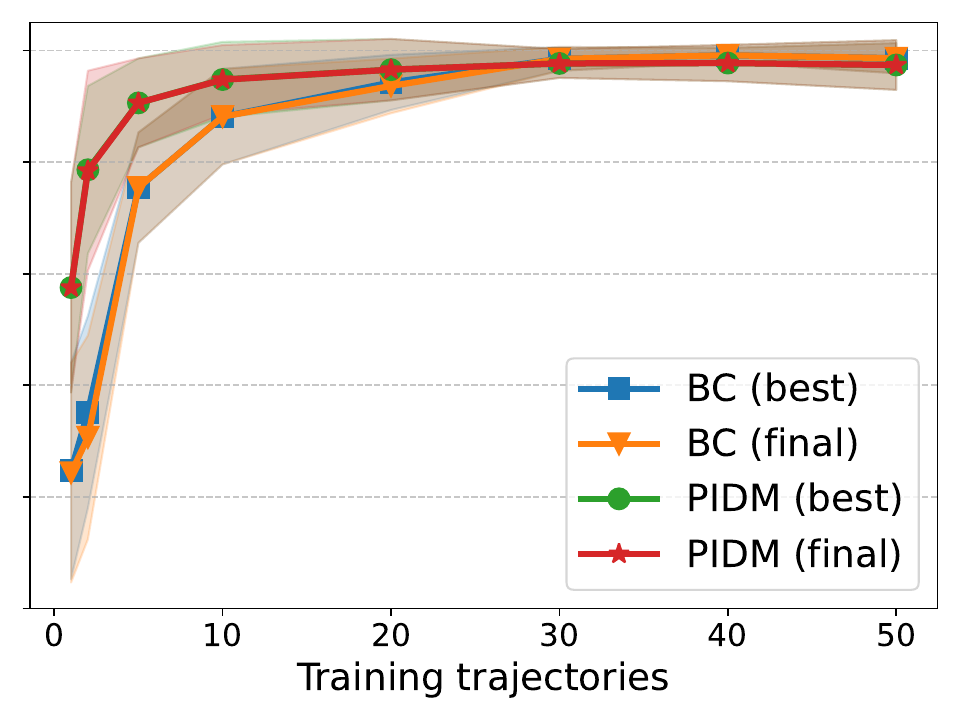}
        \caption{Multiroom}
        \label{fig:toy_multiroom_pidm_best_last_checkpoint}
    \end{subfigure}
    \caption{Performance per number of training demonstrations for BC and PIDM in four tasks trained on human demonstrations and evaluated at the last or best checkpoint. Lines and shading correspond to the average and standard deviation across 20 random seeds, respectively.}
    \label{fig:best_last_checkpoint}
\end{figure}

For our sample efficiency results in 2D navigation tasks, we evaluate BC and PIDM at four checkpoints throughout training (after \num{5000}, \num{10000}, \num{50000}, and \num{100000} gradient steps) and report the best checkpoint for each method to ensure a fair and robust comparison. However, it is common practice to report results just from the last checkpoint at the end of training, which might be more sensitive to noise and other sources of variability. To ensure that our results also hold when evaluating the last checkpoint after \num{100000} gradient steps, we compare these for both BC and PIDM in \Cref{fig:best_last_checkpoint}. We observe that for most tasks and number of training demonstrations, the last checkpoint is the highest-performing checkpoint such that the trends are consistent across all tasks and methods.

\section{Experiments under Deterministic Target Policy}
\label{app:evaluation-deterministic-target-policy}
Our main experiments in the 2D navigation environment use human demonstrations for all tasks. Human demonstrations are naturally stochastic which might add further complexity to learning a policy from these demonstrations in addition to the stochastic transitions of the environment. In this section, we conduct additional experiments within the same four tasks but with policies being trained on demonstrations collected from a deterministic \astar planner.

\subsection{Data Collection and Dataset Details}

\begin{figure}[t]
  \centering
  \begin{subfigure}{0.24\textwidth}
    \centering
    \includegraphics[width=\textwidth]{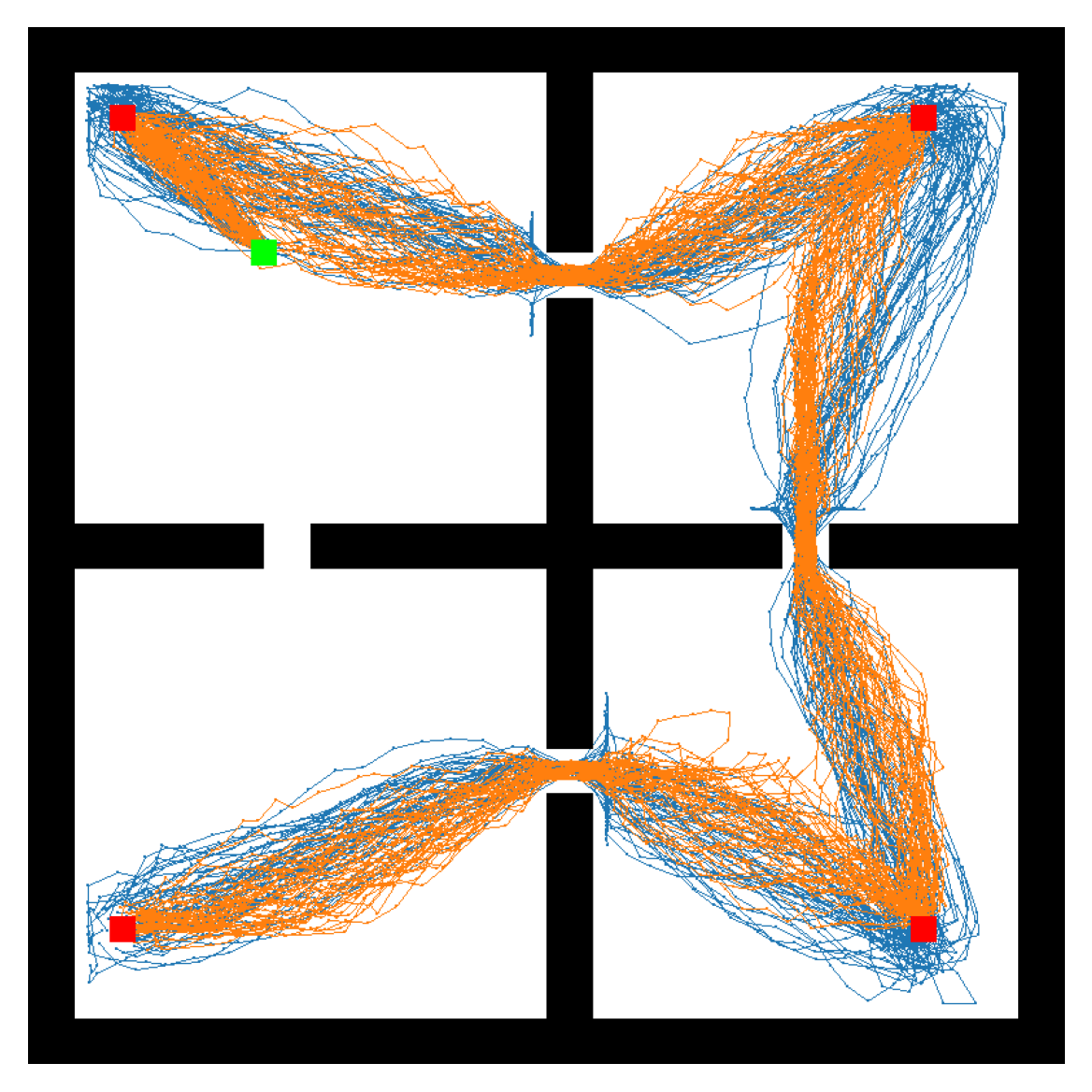}
    \caption{Four room}
    \label{fig:toy_four_room_human_planner_dataset}
  \end{subfigure}
  \hfill
  \begin{subfigure}{0.24\textwidth}
    \centering
    \includegraphics[width=\textwidth]{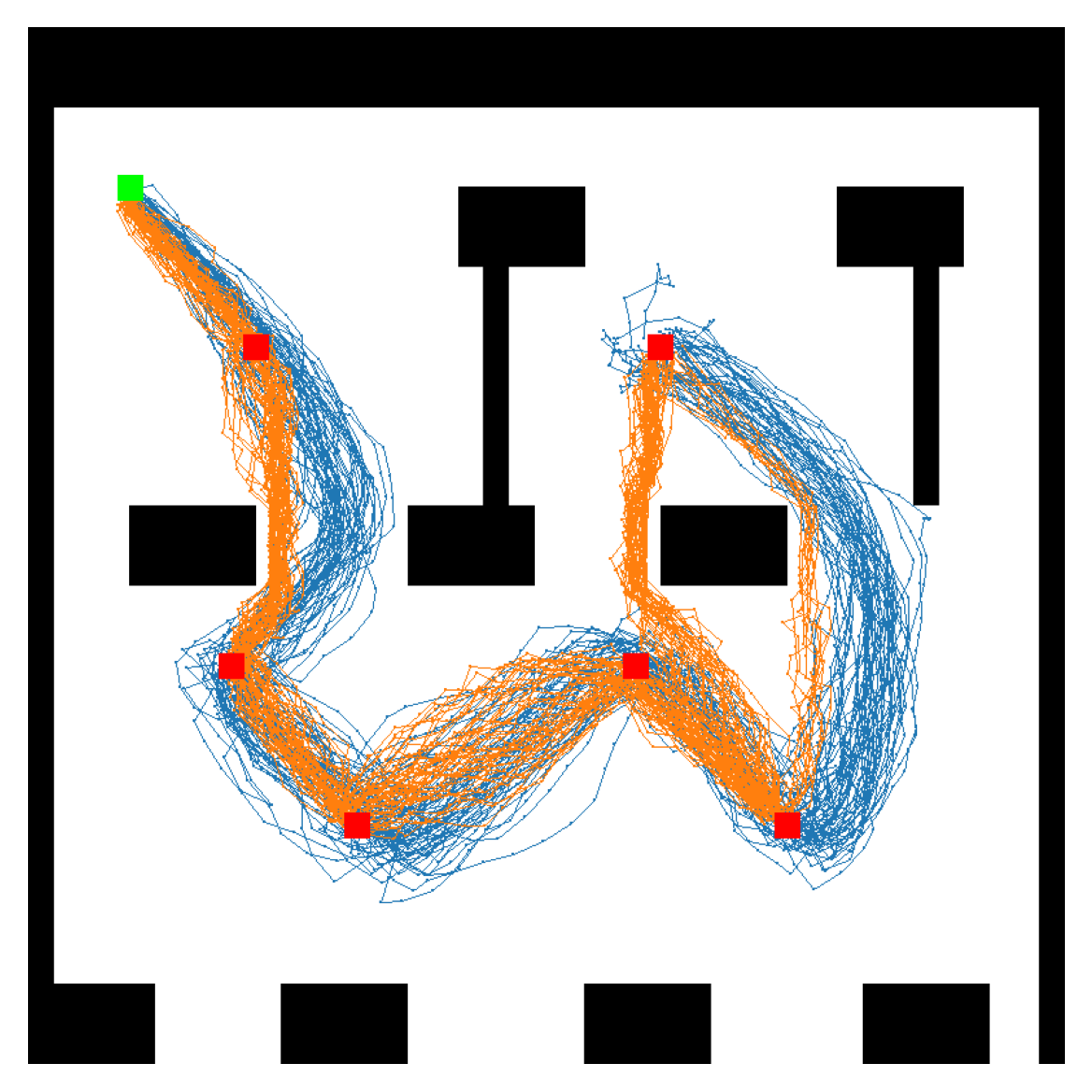}
    \caption{Zigzag}
    \label{fig:toy_zigzag_human_planner_dataset}
  \end{subfigure}
  \hfill
  \begin{subfigure}{0.24\textwidth}
    \centering
    \includegraphics[width=\textwidth]{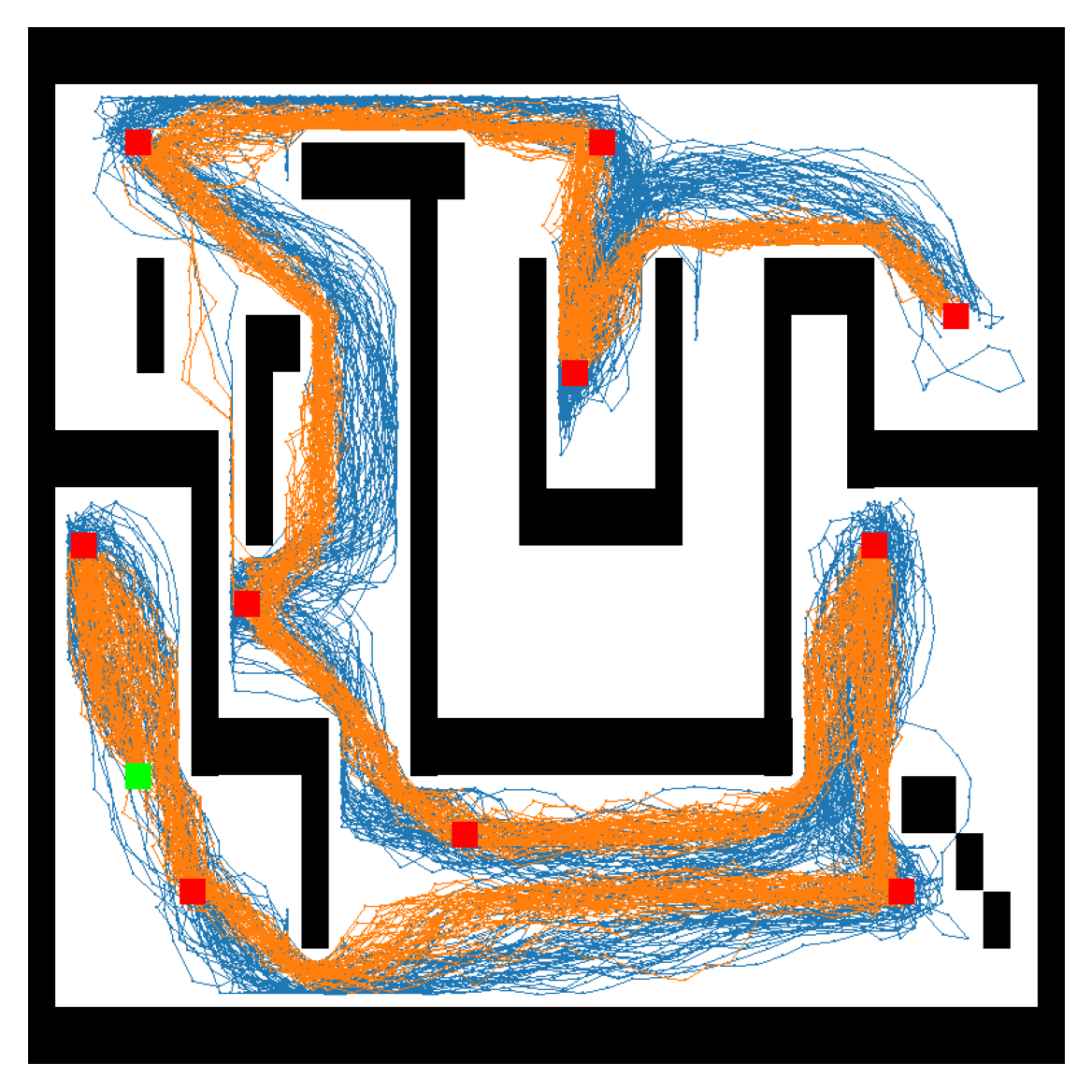}
    \caption{Maze}
    \label{fig:toy_maze_human_planner_dataset}
  \end{subfigure}
  \hfill
  \begin{subfigure}{0.24\textwidth}
    \centering
    \includegraphics[width=\textwidth]{images/toy_env/datasets/multiroom_human_planner_data.png}
    \caption{Multiroom}
    \label{fig:toy_multiroom_human_planner_dataset_new}
  \end{subfigure}
  \caption{\revised{Traces of the 50 human (blue) and \astar planner (orange) trajectories within all four 2D navigation tasks.}}
  \label{fig:toy_human_planner_datasets}
\end{figure}

\begin{table}[!htbp]
  \centering
  \caption{\revised{Statistics of all four 2D navigation tasks and the \astar datasets. The first four columns correspond to properties of the tasks, given by the number of goals, maximum number of time steps to complete the task, and the state dimensionality, while the last four columns correspond to the total number of trajectories/ time steps within the collected dataset (across all 50 trajectories) and statistics over the trajectory length.}}
  \label{tab:toy_astar_dataset_stats}
  \begin{tabular}{l c c c S[table-format=5.0] S[table-format=3.0] S[table-format=3.2] S[table-format=3.0]}
    \toprule
    \multirow{2}{*}{Task} & \multirow{2}{*}{Num goals} & \multirow{2}{*}{Max time steps} & \multirow{2}{*}{$|\st|$} & {\multirow{2}{*}{Total steps}} & \multicolumn{3}{c}{{Trajectory length}} \\
                         & & & & & {Min} & {Avg} & {Max}\\
    \midrule
    Four room & 4 & 200 & 14 & 6148 & 115 & 122.96 & 131\\
    Zigzag & 6 & 150 & 20 & 3820 & 71 & 76.4 & 85\\
    Maze & 10 & 300 & 32 & 9777 & 186 & 195.54 & 207\\
    Multiroom & 6 & 500 & 20 & 13146 & 245 & 262.92 & 277\\
    \bottomrule
  \end{tabular}
\end{table}

\textbf{\astar planner.} Given a state, the \astar planner computes an optimal plan to the next unreached goal and executes the first action along this plan. We note that this planning process is executed under average transitions which are noise-free since the Gaussian noise added within the transition function of the environment has zero mean (see \Cref{app:toy_env_details} for more details). To ensure that the planner is able to react to noise, we re-compute the plan to the next goal at every step, as in Receding Horizon Control \citep{kwon2005receding}.

\textbf{\astar datasets.} \Cref{tab:toy_astar_dataset_stats} shows statistics for each 2D navigation task and the collected \astar datasets. We also visualize the 50 collected demonstrations of the human and \astar planner for each of the tasks in \Cref{fig:toy_human_planner_datasets}. From this visualization, we can see that the \astar demonstrations tend to exhibit significantly lower variance in their trajectories, a trend that is particularly apparent in the more complex Maze and Multiroom tasks, leading to a more narrow state visitation distribution.

\subsection{Hyperparameter Search for \astar Data}
Similar to the hyperparameter tuning on human datasets for 2D navigation tasks (\Cref{app:toy_hyperparam_tuning}), we observe that BC is less stable and more sensitive to learning rate variations when trained on \astar planner demonstrations. To improve stability and evaluation robustness, we performed a search over 10 learning rate configurations in the Multiroom task using a training dataset of 50 demonstrations. These configurations were chosen based on those that yielded the highest evaluation performance in our original tuning for human datasets.

The best result for BC in Multiroom was achieved with a linear decay from $1e^{-3}$ to $1e^{-6}$ over \num{100000} steps without gradient clipping. However, BC remained less robust on \astar demonstrations. In contrast, PIDM showed stable performance across learning rates, so we used the same constant rate of $1e^{-5}$ as in the experiments with the human dataset.

To ensure reliable results despite BC’s instability, we trained and evaluated each algorithm with 50 random seeds, compared to 20 seeds for the human data evaluation.

\subsection{Evaluation Results for 2D Navigation with Deterministic Target Policy}
\label{app:evaluation-deterministic-target-policy-results}

\textbf{PIDM with human vs \astar demonstrations.} To assess the impact of the narrower data distribution of the \astar datasets on PIDM, we follow the methodology described in \Cref{sec:experiments}. 
For each task, PIDM is trained using 1, 2, 5, 10, 20, 30, 40 and 50 randomly sampled demonstrations for 50 random seeds.
We evaluate four checkpoints throughout training (after \num{5000}, \num{10000}, \num{50000}, and \num{100000} optimization steps) per seed using 50 rollouts and report aggregate performance for the best checkpoint of each task and number of training demonstrations. 

\begin{figure}[t]
    \centering
    \begin{subfigure}{.245\linewidth}
        \centering
        \includegraphics[width=\linewidth]{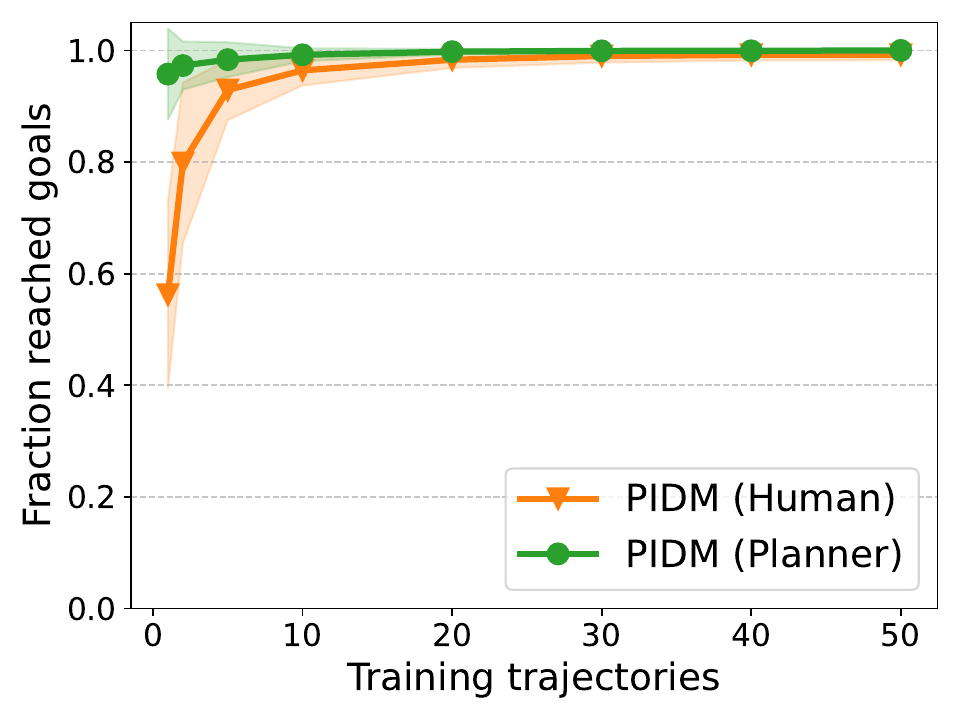}
        \caption{Four room}
        \label{fig:toy_four_room_pidm_human_planner_sample_efficiency}
    \end{subfigure}
    \hfill
    \begin{subfigure}{.245\linewidth}
        \centering
        \includegraphics[width=\linewidth]{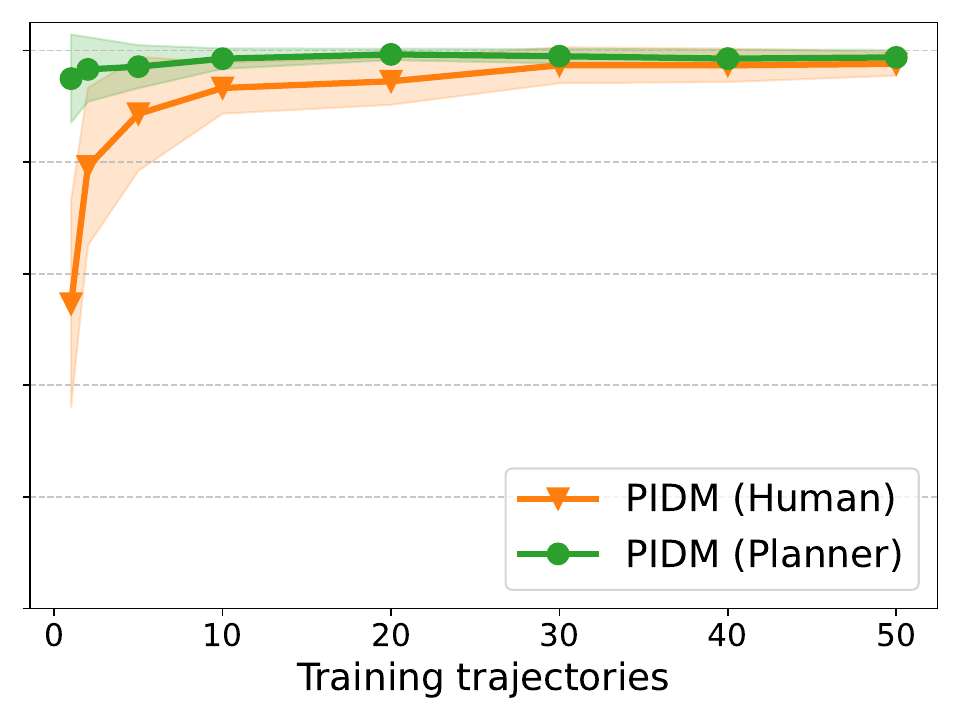}
        \caption{Zigzag}
        \label{fig:toy_zigzag_pidm_human_planner_sample_efficiency}
    \end{subfigure}
    \hfill
    \begin{subfigure}{.245\linewidth}
        \centering
        \includegraphics[width=\linewidth]{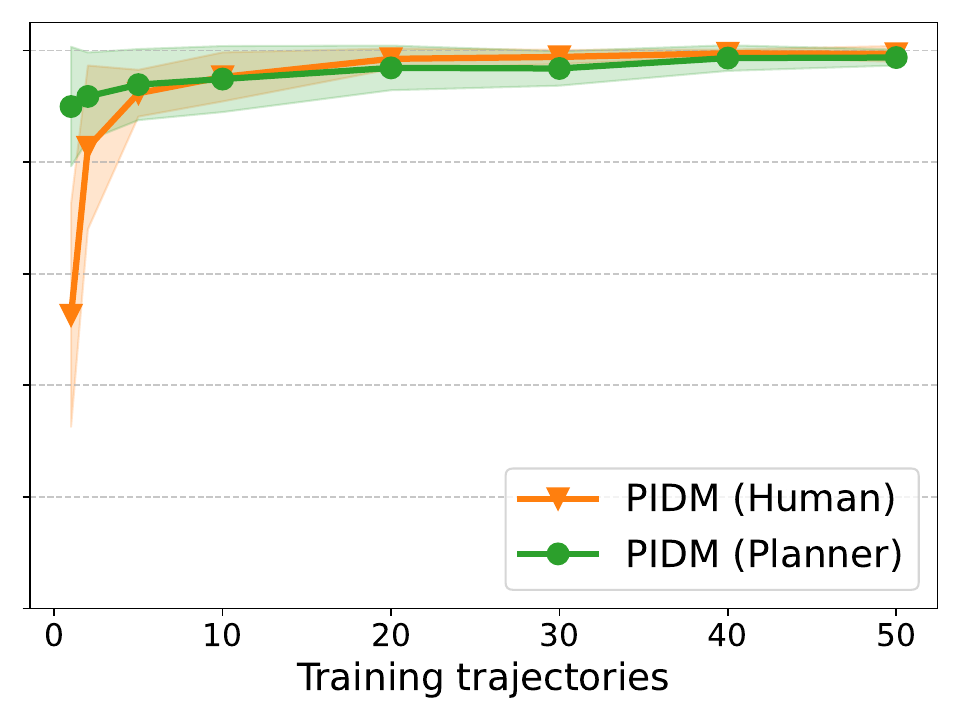}
        \caption{Maze}
        \label{fig:toy_maze_pidm_human_planner_sample_efficiency}
    \end{subfigure}
    \hfill
    \begin{subfigure}{.245\linewidth}
        \centering
        \includegraphics[width=\linewidth]{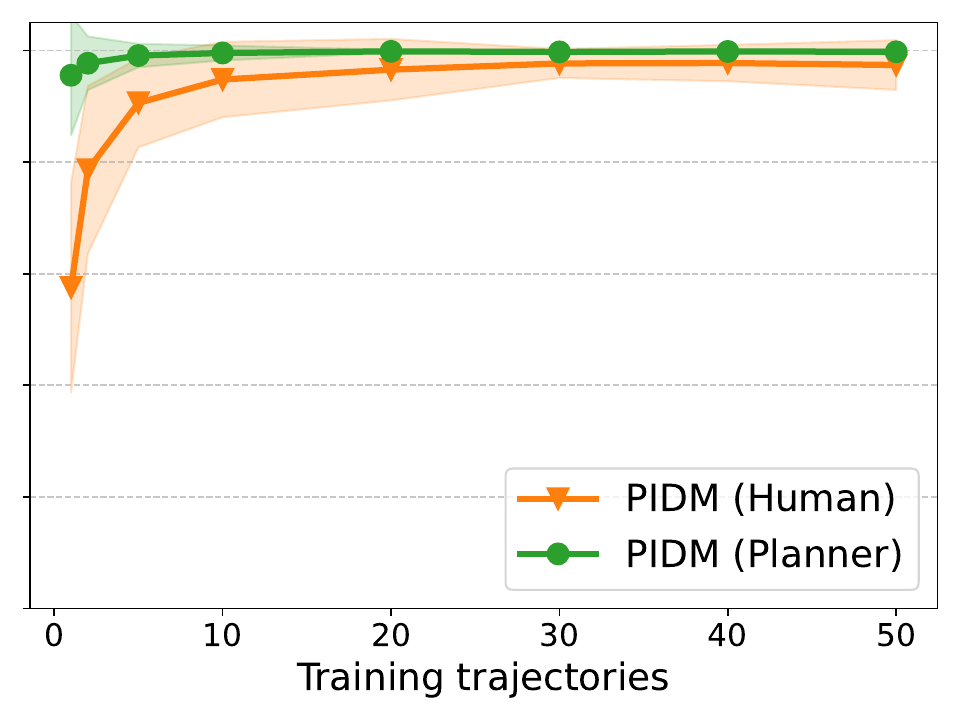}
        \caption{Multiroom}
        \label{fig:toy_multiroom_pidm_human_planner_sample_efficiency_new}
    \end{subfigure}
    \caption{\revised{Performance per number of training demonstrations for PIDM in four tasks trained on human and \astar planner demonstrations. Lines and shading correspond to the average and standard deviation across 20 and 50 seeds, respectively.}}
    \label{fig:toy_pidm_human_planner_sample_efficiency}
\end{figure}

\Cref{fig:toy_pidm_human_planner_sample_efficiency} compares PIDM's evaluation performance in the four 2D navigation tasks when trained on human vs \astar datasets. 
PIDM models trained on \astar demonstrations are notably more sample efficient, achieving high performance with as few as just one training demonstration. 
\Cref{tab:toy_pidm_human_planner_efficiency} further highlights this trend by showing the sample efficiency ratios between PIDM trained on the \astar planner and human datasets.

\begin{table}[b]
    \centering
    \caption{\revised{Maximum reached goal ratio and sample efficiency ratios of PIDM trained on \astar planner demonstrations over PIDM trained on human demonstrations for 2D navigation tasks and average across tasks.}}
    \label{tab:toy_pidm_human_planner_efficiency}
    \begin{tabular}{l S[table-format=2.2] S[table-format=2.2] S[table-format=2.2] S[table-format=2.2] S[table-format=2.3]}
        \toprule
        Task & {Four room} & {Zigzag} & {Maze} & {Multiroom} & {Average} \\
        \midrule
        $\max$ PIDM (Planner) $\uparrow$ & 1.00 & 0.99 & 0.99 & 1.00 & {--}\\
        $\max$ PIDM (Human) $\uparrow$ & 0.99 & 0.98 & 0.99 & 0.98 & {--}\\
        \midrule
        {$\eta_{\rm{PIDM (Planner)}}(80\%)$ $\uparrow$} & 5.0 & 5.0 & 2.0 & 5.0 & 4.25\\
        {$\eta_{\rm{PIDM (Planner)}}(90\%)$ $\uparrow$} & 5.0 & 10.0 & 2.5 & 5.0 & 5.625 \\
        {$\eta_{\rm{PIDM (Planner)}}(95\%)$ $\uparrow$} & 10.0 & 15.0 & 0.5 & 20.0 & 11.375 \\
        \bottomrule
    \end{tabular}
\end{table}

\textbf{State predictor error.} 
Why is PIDM notably more efficient when trained on the narrower data distribution of \astar planner demonstrations compared to human demonstrations? We hypothesize that the instance-based state predictor, as defined in \eqref{eq:instance_state_predictor}, provides more accurate future-state predictions when trained on \astar data. For states in held-out \astar trajectories, the predictor is more likely to find similar states in the training set. Additionally, the \astar collection policy is deterministic, resulting in lower action variability than the human policy.
To investigate this hypothesis, we compute the state predictor error compared to ground-truth future states in held-out trajectories for each dataset and varying number of demonstrations. 
We consider multiple sizes $n \in \lbrace 1, 2, 5, 10, 20, 30, 40, 45 \rbrace$. For each $n$, we randomly sample 50 different subsets of the 50 available trajectories that are used to learn the state predictor (i.e.\ the instance-based state predictor defined in \eqref{eq:instance_state_predictor} will lookup closest states and predicted future states within these trajectories), and use the remaining $(50 - n)$ demonstrations as held-out. 
From these held-out trajectories, we take all states and predict a future state with the state predictor and compare the predicted future state with the ground-truth.

\begin{figure}[t]
    \centering
    \begin{subfigure}{.245\linewidth}
        \centering
        \includegraphics[width=\linewidth]{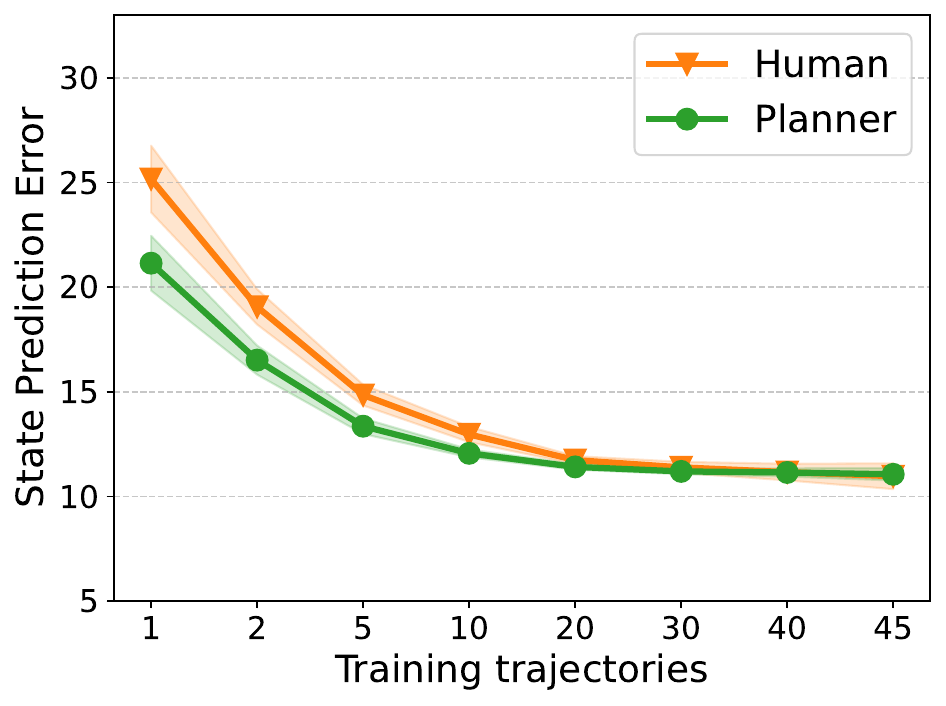}
        \caption{Four room}
        \label{fig:toy_pidm_human_planner_state_prediction_error_four_room}
    \end{subfigure}
    \hfill
    \begin{subfigure}{.245\linewidth}
        \centering
        \includegraphics[width=\linewidth]{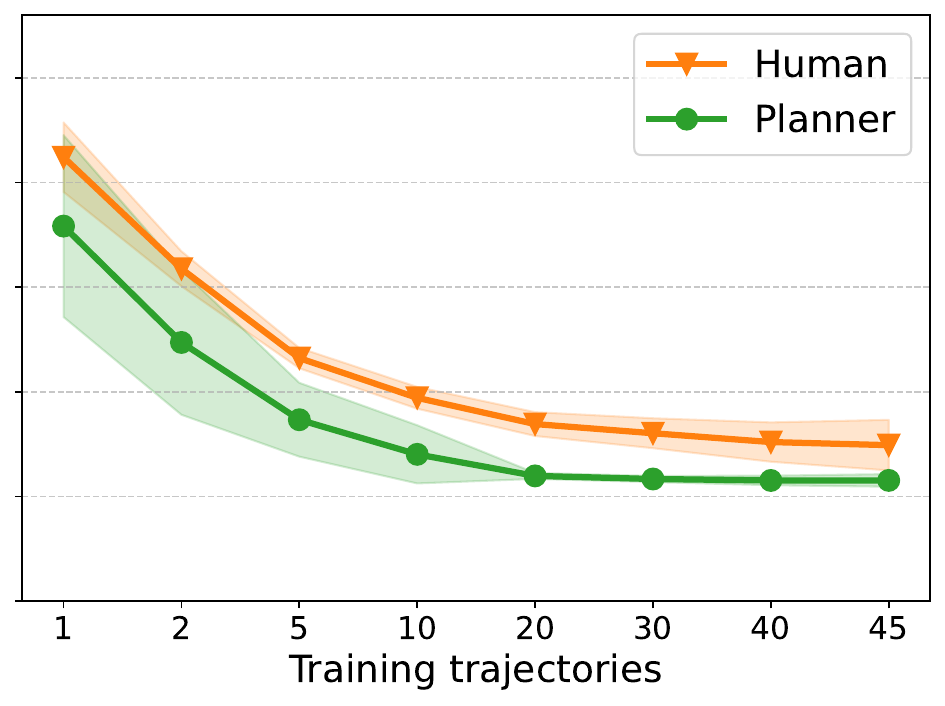}
        \caption{Zigzag}
        \label{fig:toy_pidm_human_planner_state_prediction_error_zigzag}
    \end{subfigure}
    \hfill
    \begin{subfigure}{.245\linewidth}
        \centering
        \includegraphics[width=\linewidth]{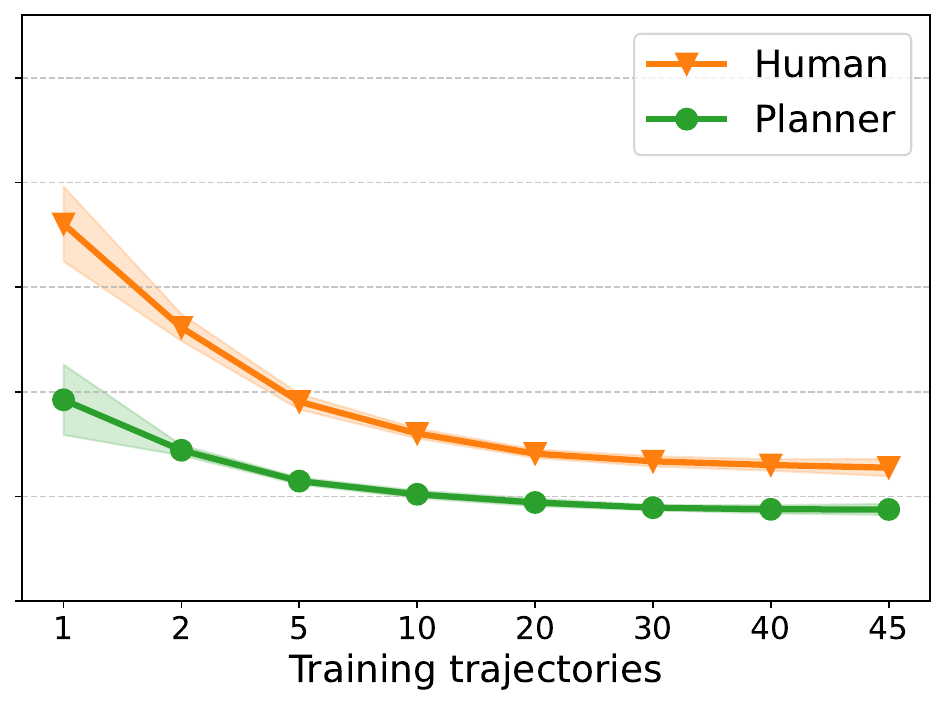}
        \caption{Maze}
        \label{fig:toy_pidm_human_planner_state_prediction_error_maze}
    \end{subfigure}
    \hfill
    \begin{subfigure}{.245\linewidth}
        \centering
        \includegraphics[width=\linewidth]{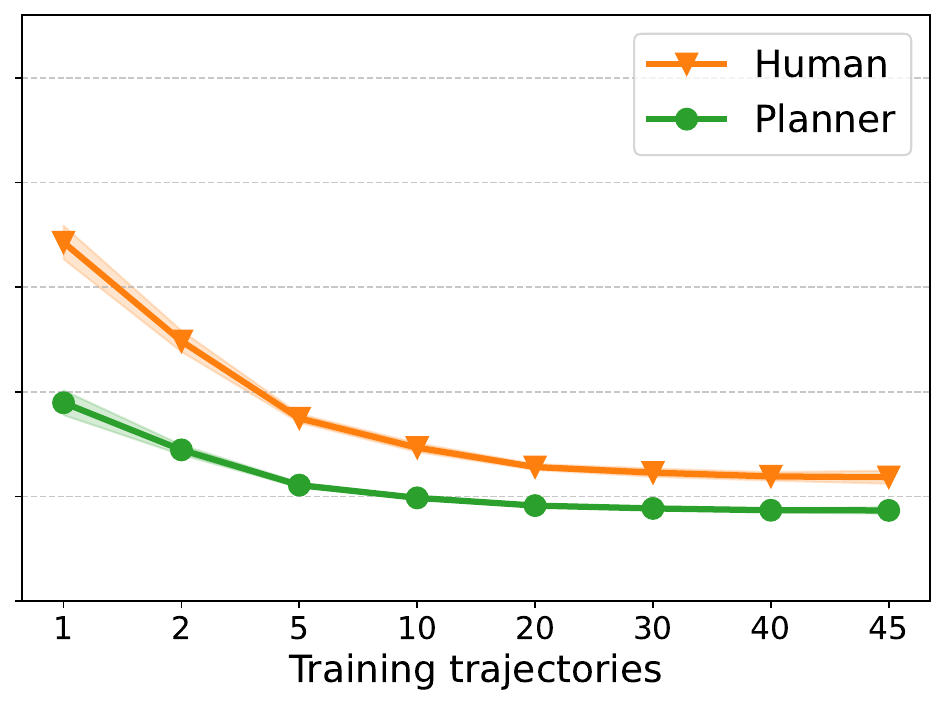}
        \caption{Multiroom}
        \label{fig:toy_pidm_human_planner_state_prediction_error_multiroom_new}
    \end{subfigure}
    \caption{\revised{Error of the 2D navigation state predictor (as defined in \eqref{eq:instance_state_predictor}) on held-out trajectories when trained on the human and \astar planner demonstrations. Lines and shading correspond to the average and standard deviation across 50 seeds that determine the sampling of demonstrations used for training, respectively.}}
    \label{fig:toy_pidm_human_planner_state_prediction_error}
\end{figure}

\Cref{fig:toy_pidm_human_planner_state_prediction_error} visualizes the state prediction error vs number of training demonstrations for each task and human and \astar datasets. 
As expected, the state prediction error decreases as the number of demonstrations grows, but plateaus before $45$ demonstrations. 
Crucially, the state state predictor error is notably lower for  \astar data, especially with few demonstrations, where the gap is largest.
This reduced error correlates with the higher sample efficiency observed on the \astar dataset (see  \Cref{fig:toy_pidm_human_planner_sample_efficiency}), providing empirical evidence for the effect of state predictor bias predicted by theoretical analysis, even in the small-data regime.

\begin{figure}[!h]
    \centering
    \begin{subfigure}{.35\linewidth}
        \centering
        \includegraphics[width=\linewidth]{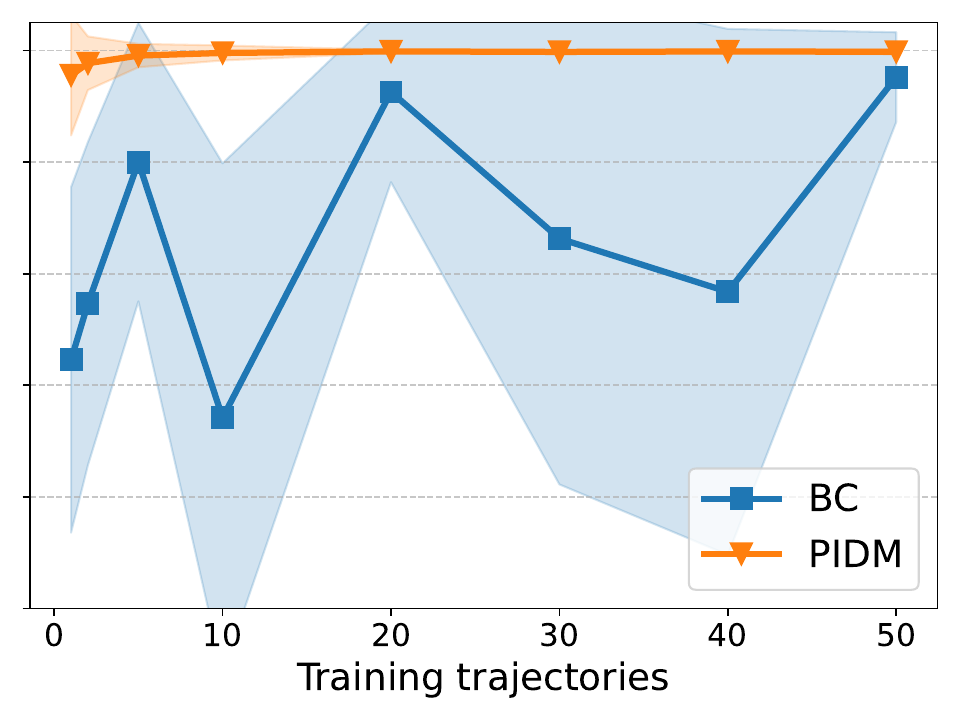}
    \end{subfigure}
    \caption{Sample efficiency of PIDM and BC in Multiroom, trained on \astar data.}
\label{fig:toy_pidm_planner_multiroom_sample_efficiency}
\end{figure}

\textbf{PIDM vs BC trained on \astar demonstrations.} 
In \Cref{sec:toy_efficiency} we showed that PIDM is notably more efficient than BC when trained on human demonstrations. \Cref{fig:toy_pidm_human_planner_sample_efficiency} further demonstrates that PIDM achieves even greater sample efficiency when trained on the \astar dataset, and this is correlated with the lower bias of the state predictor in this setting. Does BC similarly benefit similarly from the narrower distribution of \astar data? 

\Cref{fig:toy_pidm_planner_multiroom_sample_efficiency} compares the sample efficiency for PIDM and BC in the Multiroom task when trained on \astar data, aggregated across 50 random seeds. 
While PIDM clearly benefits, BC is negatively affected by the narrow data distribution, resulting in less stable training and overall lower sample efficiency compared to BC trained on the human data. We hypothesize that BC struggles to generalize from the narrow data distribution of the \astar dataset to states visited during evaluation rollouts, leading to poor evaluation performance.

\begin{figure}[t]
    \centering
    \begin{subfigure}{.245\linewidth}
        \centering
        \includegraphics[width=\linewidth]{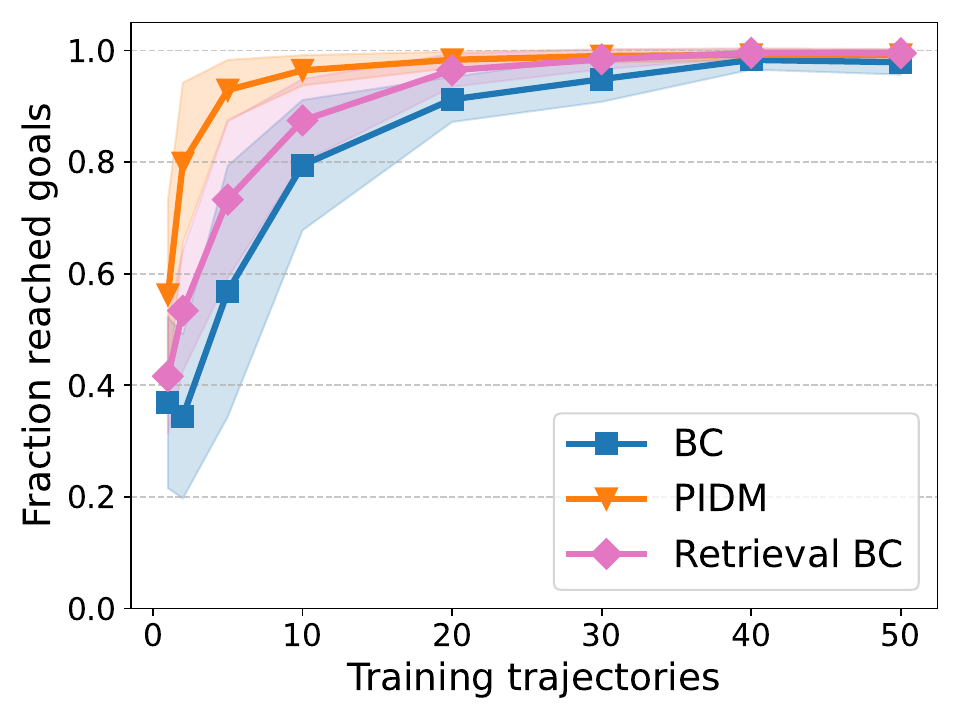}
        \caption{Four room}
        \label{fig:four_room_retrieval_bc_comparison}
    \end{subfigure}
    \hfill
    \begin{subfigure}{.245\linewidth}
        \centering
        \includegraphics[width=\linewidth]{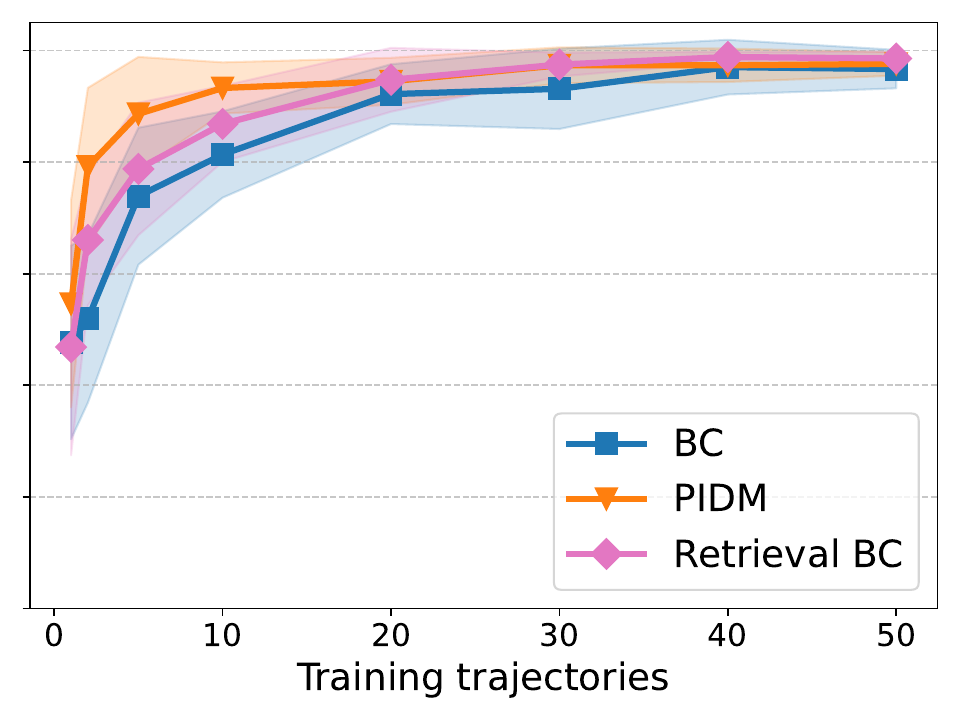}
        \caption{Zigzag}
        \label{fig:zigzag_retrieval_bc_comparison}
    \end{subfigure}
    \hfill
    \begin{subfigure}{.245\linewidth}
        \centering
        \includegraphics[width=\linewidth]{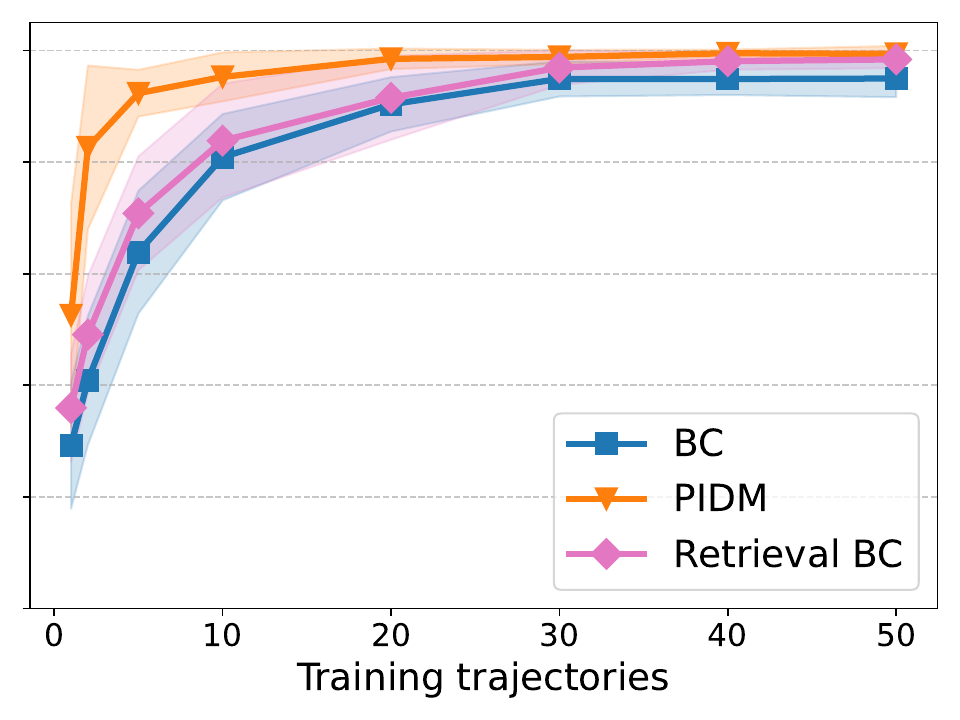}
        \caption{Maze}
        \label{fig:maze_retrieval_bc_comparison}
    \end{subfigure}
    \hfill
    \begin{subfigure}{.245\linewidth}
        \centering
        \includegraphics[width=\linewidth]{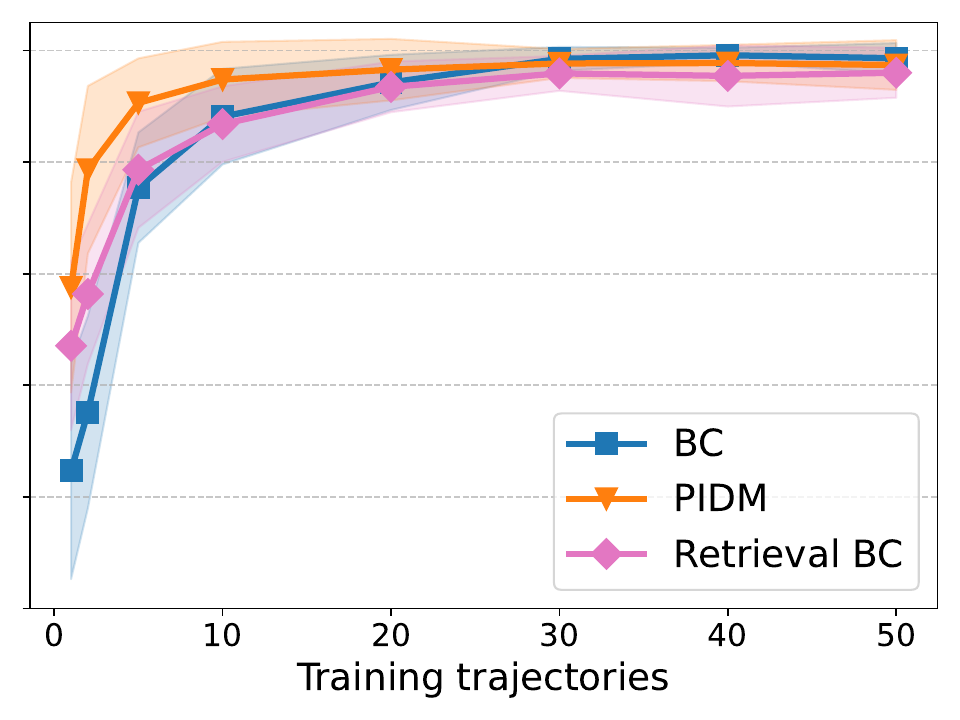}
        \caption{Multiroom}
        \label{fig:multiroom_retrieval_bc_comparison}
    \end{subfigure}
    \caption{Performance per number of training demonstrations for neural network BC, retrieval-based BC, and PIDM in four tasks trained on human demonstrations. Lines and shading correspond to the average and standard deviation across 20 random seeds, respectively.}
    \label{fig:retrieval_bc_comparison}
\end{figure}

\section{Retrieval-based Behavior Cloning}
\label{app:retrieval_bc_comparison}
In our experiments, we trained neural network policies for both BC and the IDM of PIDM, and use an instance-based state predictor that retrieves the closest state in the training set to predict a future state. To validate that the retrieval-based state predictor does not represent an unfair advantage for PIDM, we also implemented a retrieval-based BC policy. For this policy, we follow the same methodology as for the state predictor in 2D experiments, described in \Cref{sec:env_and_algorithm_details}, but instead of predicting a state $k$ steps in the future, we predict the action taken in the retrieved state:
\begin{equation}
    \label{eq:instance_bc_policy}
    \pi_\text{R-BC}(s_t) 
= 
    \ac_{\tau^\star}^{i^\star} 
\; 
    \text{with} 
\; 
    \left(
        \tau^\star, i^\star
    \right)
\defeq
    \arg\min_{\tau, i} \lvert\lvert s_t - \st_\tau^i \rvert\rvert^2
,
\end{equation}
with $\st_\tau^i$ referring to the state at time step $\tau$ in demonstration $i$ of the training dataset. We apply the same constraint for the BC policy: the nearest identified state, as measured by Euclidean distance, must match the same goal as the current state.

\begin{table}[!htbp]
    \centering
    \caption{Sample efficiency ratios of PIDM over BC and retrieval-based BC for 2D navigation tasks and average across tasks.}
    \label{tab:toy_efficiency_retrieval_bc_comparison}
    \begin{tabular}{l S[table-format=1.2] S[table-format=1.2] S[table-format=1.2] S[table-format=1.2] S[table-format=1.2]}
        \toprule
        Task & {Four room} & {Zigzag} & {Maze} & {Multiroom} & {Average} \\
        \midrule
        {$\eta_{\rm{PIDM} > \rm{BC}}(80\%)$ $\uparrow$} & 4.0 & 2.0 & 5.0 & 2.0 & 3.25\\
        {$\eta_{\rm{PIDM} > \rm{BC}}(90\%)$ $\uparrow$} & 4.0 & 2.0 & 4.0 & 4.0 & 3.5\\
        {$\eta_{\rm{PIDM} > \rm{BC}}(95\%)$ $\uparrow$} & 4.0 & 1.33 & 5.0 & 2 & 3.08 \\
        \midrule
        {$\eta_{\rm{PIDM} > \rm{R-BC}}(80\%)$ $\uparrow$} & 2.0 & 2.0 & 5.0 & 2.0 & 2.75\\
        {$\eta_{\rm{PIDM} > \rm{R-BC}}(90\%)$ $\uparrow$} & 4.0 & 2.0 & 4.0 & 4.0 & 3.5\\
        {$\eta_{\rm{PIDM} > \rm{R-BC}}(95\%)$ $\uparrow$} & 2.0 & 1.0 & 3.0 & 2.0 & 2.0 \\
        \bottomrule
    \end{tabular}
\end{table}

\Cref{fig:retrieval_bc_comparison,tab:toy_efficiency_retrieval_bc_comparison} visualize the sample efficiency and efficiency ratios comparing PIDM to both the neural network BC policy and the retrieval-based BC policy. Interestingly, these results show that the retrieval-based BC policy matches or slightly exceeds the neural network BC policy across all four 2D navigation tasks. Nevertheless, PIDM significantly outperforms both BC variants, confirming that the sample efficiency gains of PIDM hold regardless of whether BC uses a neural network or retrieval-based policy, indicating that the instance-based learning approach is not sufficient to explain PIDM's efficiency gains.

\section{Discussion on how the Lookahead Horizon $k$ affects PIDM}
\label{app:toy_sensitivity_k}

In contrast to BC, PIDM introduces a new hyperparameter, the lookahead horizon $k$, which determines how far into the future the state predictor predicts a future state, and how far into the future the IDM conditions its actions on. To understand the impact of this hyperparameter, we note that $k$ affects the PIDM algorithm differently at training and rollout times.

\begin{figure}[t]
    \centering
    \includegraphics[width=\textwidth]{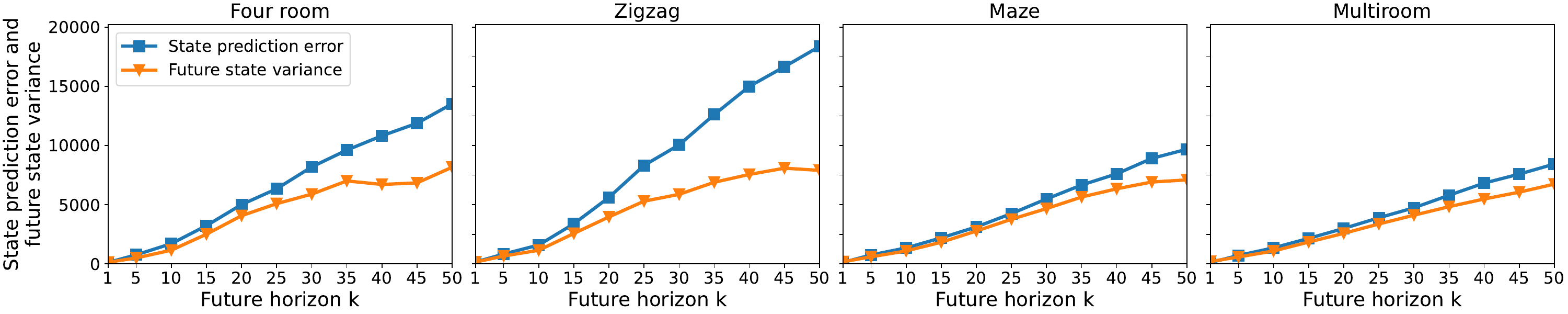}
    \caption{Variance of future states and state prediction error for varying $k$ in the human datasets of all four 2D navigation tasks. The state predictor has been trained on 40 demonstrations and evaluated on the remaining 10 held-out trajectories.}
    \label{fig:toy_future_std_and_state_pred_error}
\end{figure}

Training IDM models with sufficiently large $k$ and multiple values of $k$ can benefit the state representations learned by the network, e.g. by reducing the impact of exogeneous noise  under some assumptions~\citep{mhammedi2023musik, efroni2022ppe, lamb2023guaranteed}, as also discussed in \Cref{sec:related_work}. These benefits grow as representation learning becomes more relevant for the task in question, e.g.\ they can be expected to be more pronounced in our 3D video game experiments where PIDM learns from high-dimensional vision inputs. In contrast, in tasks where informative and compressed state representations are already available, e.g.\ in our 2D navigation experiments where agents receive compact state feature vectors as input, training on multiple or large values of $k$ might not be critical. Additionally, training PIDM with very large values of $k$ can make the state prediction task more difficult, potentially leading to higher state prediction error and thus reduced performance. Motivated by these considerations, we train PIDM models with varying values of $k$ ranging from $k=1$ to $k=26$ in our 3D video game experiments but keep $k=1$ fixed in our 2D navigation experiments.

At test time, we observe that using a smaller value of $k$ typically results in improved rollout performance as long as it was included in the training set. This trend can be explained by the distribution $p(s_{t+k} \mid s_t)$ for states that are further in the future exhibiting higher variance for larger $k$ (caused by stochasticity of transitions accumulating over multiple steps). The increase in variance further leads to higher state prediction error and, thus, reduced performance, as discussed in \Cref{sec:theoretical-analysis}. To empirically validate this accumulation of variance and its impact on the state prediction error, we compute the expected variance of future states given each current state, as well as the state prediction error, on the human datasets for all 2D navigation tasks and for varying $k\in\{1,5,10,15,20,25,30,35,40,45,50\}$. \Cref{fig:toy_future_std_and_state_pred_error} clearly confirms this trend by visualizing the squared $L2$ distance for the state prediction error to obtain quantities of comparable scale
\begin{equation}
    \text{State prediction error}: 
        \E \left[ 
            \sum_{i=1}^{|\st|} 
            \left( 
                \widehat{\bs}_{t+k}^{(i)} - \bs_{t+k}^{(i)} 
            \right)^2
        \right] 
    ,\;\; 
    \text{and} 
    \;\;
    \E
    \left[ 
        \mathrm{Var}(\bs_{t+k} \mid \bs_t) 
    \right] 
    = 
    \E 
    \left[ 
        \sum_{i=1}^{|\st|} 
        \left( 
            \bs_{t+k}^{(i)} - \E[\bs_{t+k}^{(i)} \mid \bs_t] 
        \right)^2 
    \right],
\end{equation}
with $s_t^{(i)}$ and $\widehat{s}_{t+k}^{(i)}$ denoting the $i$-th component of state vector, $s_t$, and the predicted state, $\widehat{s}_{t+k}$, respectively.
To approximate the future-state variance, we follow the same clustering methodology as described in \Cref{sec:toy_future_conditioning_analysis}. To obtain a state predictor that is as accurate for a given $k$ as it can be but still retain some held-out demonstrations for evaluation, we learn our instance-based state predictor on 40 randomly sampled demonstrations and evaluate its error on the remaining 10 held-out demonstrations. To ensure our estimation is not biased by the random sampling of demonstrations, we repeat this computation for 100 randomly sampled training-test splits and report the averaged state prediction error across all runs. Across all tasks, we see that the lowest state prediction error is achieved for $k=1$, motivating this value for our experiments.

\section{Details for 2D Navigation Environment and Experiments}
\label{app:toy_details}

\subsection{Additional Environment Details}
\label{app:toy_env_details}
Tasks within the 2D navigation environment specify a layout of the environment and differ in the number of goals. The general setting stays the same with each task specifying an order to its goals and the agent needs to reach a goal before being able to reach any subsequent goals. This setup makes these tasks punishing since missing any goal will mean that subsequent goals cannot be reached anymore unless the agent returns back to the currently required goal. An episode within any task finishes after all goals have been reached, or after a maximum number of time steps has been reached. The state dimensionality, number of goals, and maximum number of time steps for each task is listed in \Cref{tab:toy_dataset_stats}.

In all tasks, we introduce stochasticity in the transition function through Gaussian noise. Instead of displacing the agent based on its selected action $a \in [-1, 1]^2$ alone, we displace the agent based on clipped noise-added actions:
\begin{equation}
    \text{clip}(a + \epsilon, -1, 1) \quad \text{with} \quad \epsilon \sim \mathcal{N}(0, 0.2 \cdot \mathds{1})
\end{equation}
We emphasize that the sampled noise is \emph{not} modifying the actions but rather modeled as part of the environment, meaning that, from the perspective of the agent, the environment transitions are stochastic given a state and action. The agent will bounce off any walls that it collides with with walls being visualized as black bars in all figures.

\subsection{Dataset Details}
\Cref{tab:toy_dataset_stats} shows statistics for each 2D navigation task and the collected human dataset. During data collection, the human player was instructed to collect high-quality trajectories that reach all goals as fast as possible. The player controlled the movement of the controllable agent using the joystick of a gamepad controller. We note that the player was unaware of the data analyses that we conducted to avoid any risk of introducing bias.

\begin{table}
  \centering
  \caption{Statistics of all four 2D navigation tasks and the human datasets. The first four columns correspond to properties of the tasks, given by the number of goals, maximum number of time steps to complete the task, and the state dimensionality, while the last four columns correspond to the total number of trajectories/ time steps within the collected dataset (across all 50 trajectories) and statistics over the trajectory length.}
  \label{tab:toy_dataset_stats}
  \begin{tabular}{l c c c S[table-format=5.0] S[table-format=3.0] S[table-format=3.2] S[table-format=3.0]}
    \toprule
    \multirow{2}{*}{Task} & \multirow{2}{*}{Num goals} & \multirow{2}{*}{Max time steps} & \multirow{2}{*}{$|\st|$} & {\multirow{2}{*}{Total steps}} & \multicolumn{3}{c}{{Trajectory length}} \\
                         & & & & & {Min} & {Avg} & {Max}\\
    \midrule
    Four room & 4 & 200 & 14 & 5821 & 103 & 116.42 & 154\\
    Zigzag & 6 & 150 & 20 & 4009 & 66 & 80.18 & 106\\
    Maze & 10 & 300 & 32 & 9785 & 176 & 195.70 & 227\\
    Multiroom & 6 & 500 & 20 & 12961 & 241 & 259.22 & 314\\
    \bottomrule
  \end{tabular}
\end{table}

\subsection{Hyperparameter Search}
\label{app:toy_hyperparam_tuning}
To ensure fair comparison, we conducted a comparable hyperparameter search for both BC and PIDM in the multiroom task using 50 training demonstrations. First, we conducted a hyperparameter search over the model architecture considering sixteen different sizes of the MLP network architecture, the use of normalization in the network (either batch normalization, layer normalization, or no normalization), and learning rate with three constant candidate learning rates ($1e^{-6}, 1e^{-5}, 1e^{-4}$. The considered architectures consisted of any of five MLP blocks before any potential normalization layer and any of the five MLP blocks after the normalization. The considered network blocks were: MLP(256), MLP(256, 128), MLP(512, 256), MLP(512, 1024, 256), and MLP(1024, 2048, 512).

From this search, we identified a single network architecture that performed best for BC and among the best for PIDM to keep for consistent comparisons thereafter. The architecture consists of network block MLP(512, 1024, 256) followed by batch normalization before MLP(256, 2) with the last 2D layer outputting the action logits. We apply ReLU activation in between all layers and $tanh$ activation to the output logits.

\begin{table}[h]
    \centering
    \caption{Learning rate configuration for each task and algorithm}
        \begin{tabular}{c c c}
            \toprule
            Task & BC configuration & IDM configuration \\
            \midrule
            Four room & Linear decay $1e^{-3} \rightarrow 1e^{-6}$ over \num{50000} steps + grad norm clipping & constant $1e^{-5}$ \\
            Zigzag & Linear decay $1e^{-4} \rightarrow 1e^{-6}$ over \num{50000} steps + grad norm clipping & constant $1e^{-5}$ \\
            Maze & Linear decay $1e^{-4} \rightarrow 1e^{-6}$ over \num{50000} steps + grad norm clipping & constant $1e^{-5}$ \\
            Multiroom & Linear decay $1e^{-4} \rightarrow 1e^{-6}$ over \num{50000} steps & constant $1e^{-5}$ \\
            \bottomrule
        \end{tabular}
\end{table}

After fixing the network architecture, we still found some training instability for BC and IDM so we decided to further tune the learning rate for BC and IDM by searching over 14 learning rate configurations defined by their initial learning rate, and potential learning rate scheduling, and considered each configuration with and without gradient norm clipping. We first tuned the learning rate configuration for BC and IDM in multiroom after which we found IDM training to be stable across tasks. For BC, we further tuned the learning rate for each individual task to obtain stable training results. The identified learning rates are shown in the table below.

For the choice of the lookahead horizon $k$ during training, we used a fixed $k=1$ for all 2D navigation tasks during training and evaluation. For further discussion, we refer to \Cref{app:toy_sensitivity_k}.

\section{Details for Complex Task in 3D-World}
\label{app:xbox_details}

\subsection{Dataset Details}
The dataset consists of 30 demonstrations collected by a human playing the game.
\Cref{tab:tour_dataset_stats} shows the number of steps and length (seconds) of the demonstrations in the dataset.
\begin{table}[h]
  \centering
  \caption{Statistics of demonstrations of "Tour" task.}
  \label{tab:tour_dataset_stats}
  \begin{tabular}{c
                  S[table-format=5.0] S[table-format=5.0] S[table-format=5.0]
                  S[table-format=3.2] S[table-format=3.2] S[table-format=3.2]}
    \toprule
    & \multicolumn{3}{c}{Total steps} & \multicolumn{3}{c}{Trajectory length (in seconds)} \\
    \cmidrule(lr){2-4} \cmidrule(lr){5-7}
    Task & {Min} & {Avg} & {Max} & {Min} & {Avg} & {Max} \\
    \midrule
    Tour & 1006 & {$1067.2 \pm 29.4$} & 1139 & 33.83 & {$35.91 \pm 0.99$} & 38.29 \\
    \bottomrule
  \end{tabular}
\end{table}

\subsection{Additional Environment Details}
\label{app:tour-env-details}
\Cref{tab:dojo_tour_milestones} contains the 11 milestones required to complete the "Tour" task.
\begin{table}
\centering
\caption{Milestones of "Tour" task in Bleeding Edge with corresponding thumbnails}
\label{tab:dojo_tour_milestones}
\begin{tabular}{c l c}
\toprule
\# & \textbf{Milestone} & \textbf{Thumbnail} \\
\midrule
1 & Start off with a sharp left 180$^\circ$ turn & \includegraphics[width=3cm]{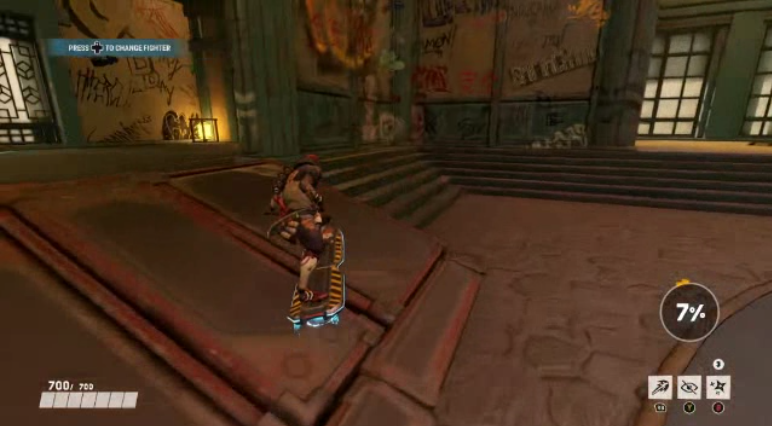} \\
2 & Navigate towards the first health marker and grab it & \includegraphics[width=3cm]{images/bleeding_edge/milestones/tour_2.png} \\
3 & Cross the main floor of the Dojo & \includegraphics[width=3cm]{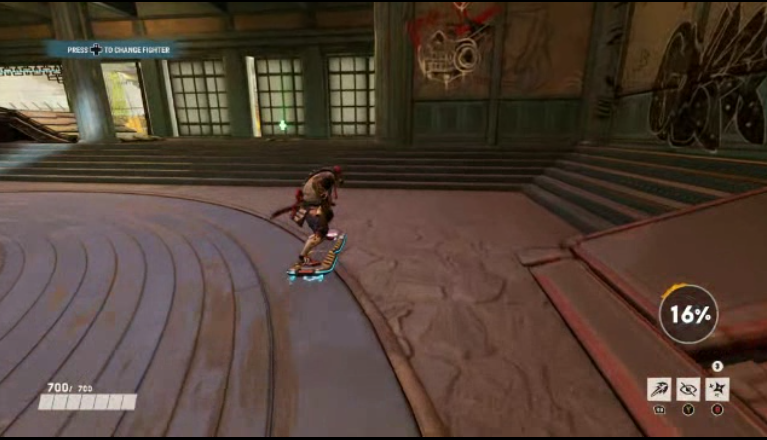} \\
4 & Take a left onto the ramp & \includegraphics[width=3cm]{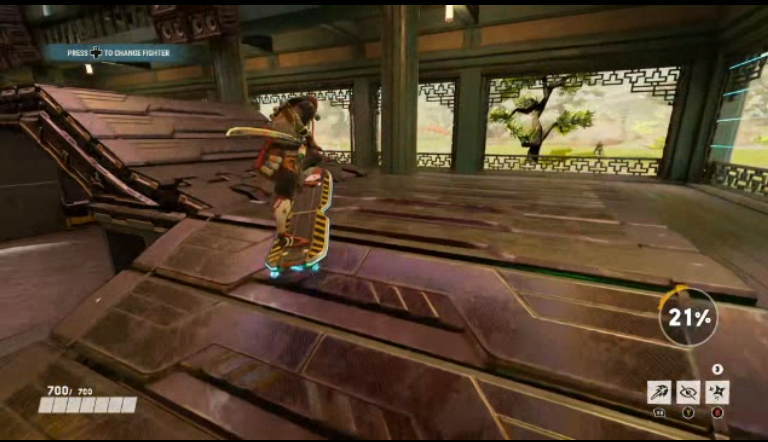} \\
5 & Turn while going up and stay on the ramp for 6-7 secs & \includegraphics[width=3cm]{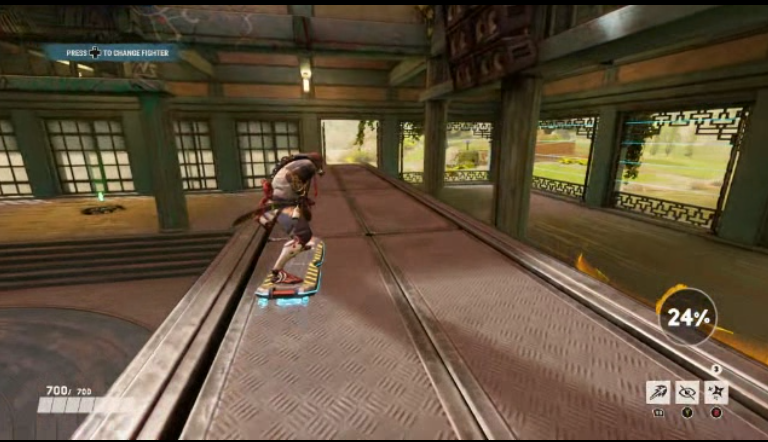} \\
6 & Right turn and navigate the corridor & \includegraphics[width=3cm]{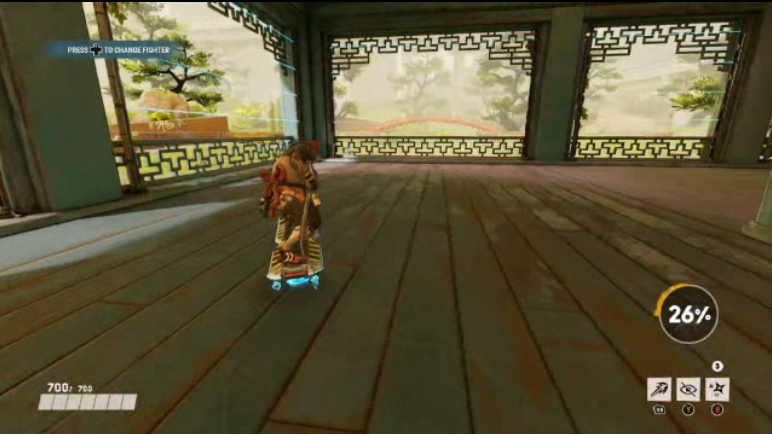} \\
7 & Circumvent the box by steering left & \includegraphics[width=3cm]{images/bleeding_edge/milestones/tour_7.png} \\
8 & Navigate towards the second health marker and grab it & \includegraphics[width=3cm]{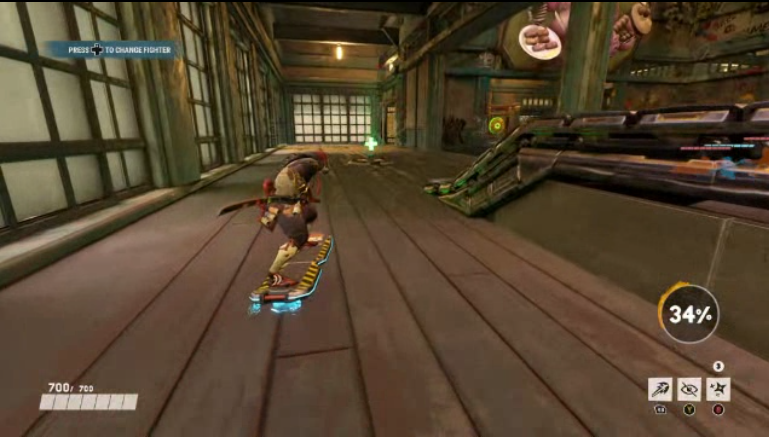} \\
9 & Pass through the final corridor & \includegraphics[width=3cm]{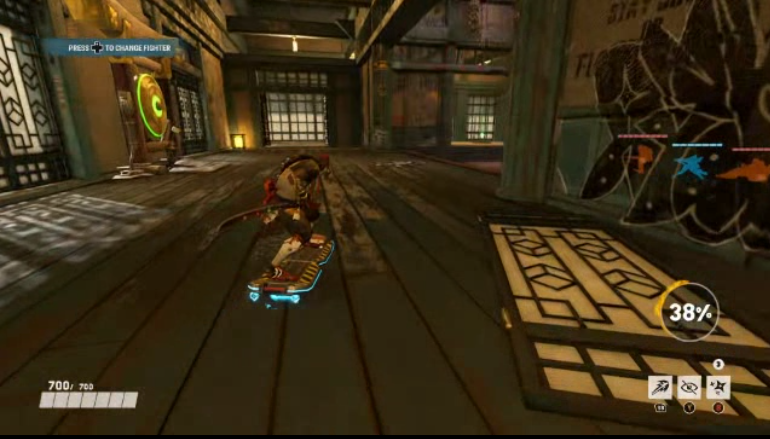} \\
10 & Hit the Gong & \includegraphics[width=3cm]{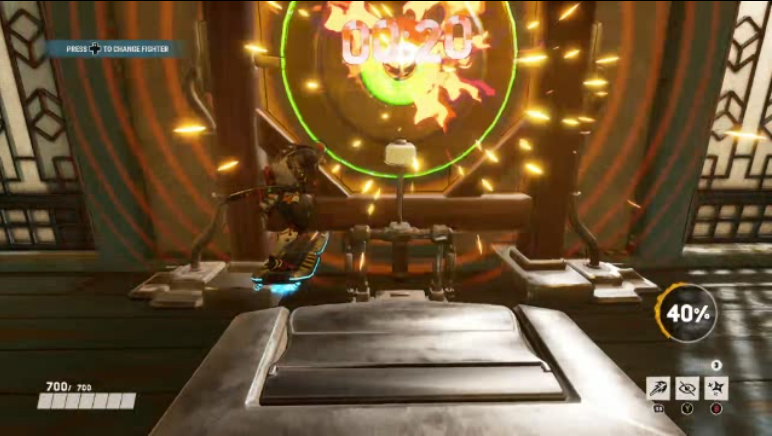} \\
11 & Stop and don't move anymore & \includegraphics[width=3cm]{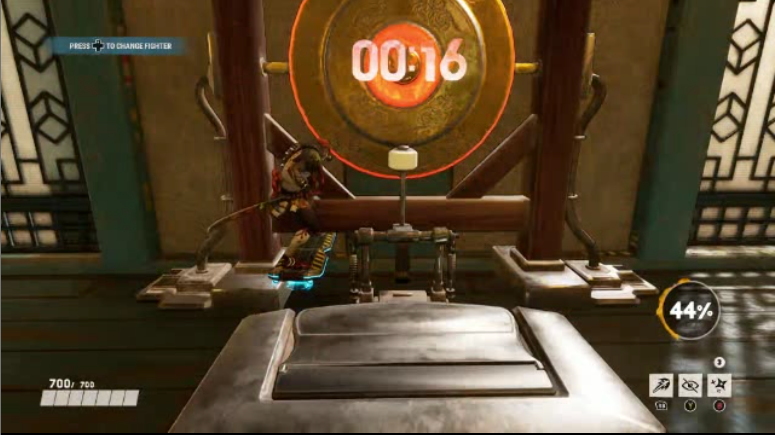} \\
\bottomrule
\end{tabular}
\end{table}

\subsection{Evaluation Protocol}
Two human experts that were familiar with the task evaluated all the rollouts. The evaluation was blind to avoid cognitive bias, since the evaluators did not know whether the rollout they were evaluating corresponded to BC or PIDM.
For each rollout, they checked if the agent achieved every milestone of the task, scoring with value $1$ if the milestone was achieved and $0$ otherwise, so the maximum score per rollout is $11$ (the number of milestones).
However, we report performance in terms of \% of this maximum score.

\subsection{Additional Algorithmic Details}
\subsubsection{Vision encoder}
We use "theia-base-patch16-224-cddsv" from Huggingface as pretrained vision encoder. 
The vision encoder remains frozen during training (and evaluation).
Each video frame is passed to this encoder, which generates an embedding vector of length 768.
This embedding vector of the current frame is the input to the BC policy.
While the embedding of the current and future frames are the input to the state encoder of the PIDM.

\subsubsection{Hyperparameter Search}
To ensure fair comparison and some degree of generalization, we conducted a comparable hyperparameter search for both BC and PIDM in a different more complex task, with more milestones, for which none of the algorithms could achieve 100\% performance after being trained with a dataset of 30 demonstrations.
We used the results from the hyperparameter search in the 2D environment as a basis, with ReLu activations in between all layers and batch normalization at the output of the state encoder. The output was a \textit{tanh} activation.
We evaluated two different sizes of the MLP network architecture, under two learning rates. The considered MLP network blocks were:
\begin{enumerate}
    \item State encoder: MLP(1024, 512, 512), Policy: MLP(512, 256)
    \item State encoder: MLP(1024, 2048, 1024, 512, 512), Policy: MLP(512, 512, 256)
\end{enumerate}
We also tried two learning rates per algorithm, namely linear decay $1e$-3 $\rightarrow 1e$-6 and $5e$-5 for PIDM, and linear decay $1e$-4 $\rightarrow 1e$-6 and $1e$-4 for BC, with decay for 60,000 steps.
Other hyperparameters that remained constant where:
training lasted 60,000 steps, optimization algorithm was Adam with standard parameters $(\beta_1=0.9$, $\beta_2=0.999$, $\epsilon=1e$-8), and batch size was 4096. 

We observed the small network blocks with linear decay was the best combination, and BC (88\%) achieved slightly higher average performance than PIDM (86\%) for that task, but not statistically significant. 
For training in the "Tour" task, we used this configuration and used the rest of the parameters used for the hyperparameter search, with the only exception of the number of training steps, which we increased to 100,000 and we could see the loss had converged and remained stable after 60,000 (which is when the linear decay stops).

For the choice of the lookahead horizon $k$, we train PIDM with varying $k\in\{1,6,11,16,21,26\}$ and use $k=1$ in rollouts for evaluation. For further discussion of these choices, we refer to \Cref{app:toy_sensitivity_k}.

\end{document}